\documentclass{class/thesis-arxiv}

\usepackage{blindtext} 
\usepackage{import}    

\begin{document}


\begin{titlepage}

	\centering

	\includegraphics[width=0.4\linewidth, valign=c]{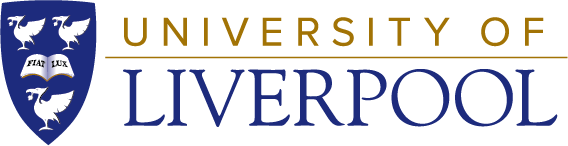}
	\vfill

	\rule{\linewidth}{0.5mm}
	\vspace{0.01cm}

	\raggedright
	{\fontsize{26pt}{26pt}\selectfont\bfseries{Towards Autonomous Navigation in Endovascular Interventions}}

	\vspace{0.01cm}
	\rule{\linewidth}{0.5mm}

	\vspace{0.5cm}

	\small
	\renewcommand{\arraystretch}{1.4}
	\begin{originaltable}[!h]
		{
			\begin{tabularx}{\linewidth}{r l p{2.0cm} r l}
				\textbf{Author}    & Tudor Jianu              &  & \textbf{Supervisors} & Dr. Anh Nguyen$^{1}$          \\
				\textbf{Submitted} & May 5$^{\text{th}}$ 2025 &  &                      & Dr. Sebastiano Fichera$^{2}$   \\
				                   &                          &  &                      & Dr. Pierre Berthet-Rayne$^{3}$
			\end{tabularx}
		}
	\end{originaltable}
	\renewcommand{\arraystretch}{0.8}

	\raggedright
	\vspace{0.5cm}

	$^{1}$Department of Computer Science, University of Liverpool, UK.\\
	$^{2}$Department of Mechanical and Aerospace Engineering, University of Liverpool, UK.\\
	$^{3}$Caranx Medical, Research and Development, FR.

	\vfill

	\centering

	A thesis submitted in fulfilment of the requirements for the degree of \textit{Philosophiae Doctor}

	\vspace{.5cm}
	{\large Department of Computer Science,\\\vspace{0.2cm}
		\Large University of Liverpool}

	\vspace{1cm}

\end{titlepage}

\clearpage
\thispagestyle{empty}
\vspace*{\fill}

\begin{center}
	Department of Computer Science,\\
	University of Liverpool,\\
	Brownlow Hill,\\
	Liverpool,\\
	L69 7ZX\\
	\vspace{1cm}
	Tudor Jianu \textcopyright \, 2025\\
\end{center}

\vspace{2cm}
\clearpage

\clearpage
\thispagestyle{empty}

\vspace*{\fill}

\setlength\epigraphwidth{0.70\textwidth}

\epigraph{
	\textit{``We are what we repeatedly do. Excellence, then, is not an act, but a habit.''}
}{
	Will Durant --- The Story of Philosophy
}



\vspace*{\fill}
\clearpage

\thispagestyle{empty}
\vspace*{\fill}

\begin{cabstract}
	\aboveparrule

	I dedicate this thesis to everyone I have ever met, whether for a fleeting moment or a lifetime. Each encounter, no matter how brief, has left a mark on me. I am grateful for the lessons, the memories, and the shared experiences that have shaped who I am today.

	You have all contributed to my journey, and I carry a part of you with me as I continue to grow. Thank you for inspiring me to become the best version of myself. I hope we keep learning from one another, and that our paths cross again in meaningful ways.

	This is for all of you. Thank you for being part of my life.
\end{cabstract}

\vspace*{\fill}
\clearpage

\thispagestyle{empty}
\chapter*{Declaration}

I hereby declare that the research work presented in this Thesis is entirely my own and that it has not been submitted, in whole or in part, for any other academic degree or qualification. I further declare that this Thesis has been submitted solely for the purpose of obtaining the PhD degree.

In each chapter of this Thesis, all published papers and articles that have contributed to the work are properly referenced and linked. Similarly, any illustrations created by me and used in previous versions of this work are appropriately linked and credited.

The formatting of this document was initially based on a template by Rob Robinson \parencite{robinson2020thesis}.

\clearpage

\clearpage
\chapter*{Acknowledgements}
\onehalfspacing

This PhD journey has been one of the most challenging yet rewarding experiences of my life, and I could not have completed it without the support, guidance, and encouragement of many incredible people.

First and foremost, I would like to express my deepest gratitude to my primary supervisor, Dr. Anh Nguyen. Your support, patience, and insightful advice have been instrumental in shaping this research and in helping me grow as a researcher. Your encouragement and belief in my abilities sustained me through both the highs and lows of this journey.

I am also deeply grateful to my second supervisor, Dr. Sebastiano Fichera, and my third supervisor, Dr. Pierre Berthet-Rayne. Sebastiano, your invaluable feedback and technical expertise have greatly enriched this work. Pierre, your guidance and steady support have been both reassuring and motivating. I feel truly fortunate to have had such a brilliant and supportive supervisory team.

My sincere thanks also go to the University of Liverpool for providing an excellent academic environment, and to the Engineering and Physical Sciences Research Council (EPSRC) for funding this research. Without which this work would not have been possible.

To my friends, near and far: thank you for being there --- celebrating milestones, offering encouragement, and providing welcome distractions when they were needed most. Your friendship has been a constant source of strength.

Finally, to my family, especially to my mom. Your love, sacrifices, and steadfast belief in me have been the foundation of everything I’ve achieved. I could not have done this without you.

\chapter*{Abstract}
\glsresetall

\onehalfspacing
\begin{cabstract}
	Cardiovascular Diseases (CVDs) remain the leading cause of global mortality, with minimally invasive treatment options offered through endovascular interventions. However, the precision and adaptability of current robotic systems for endovascular navigation are impeded by constraints such as heuristic control, limited autonomy, and the absence of haptic feedback. In this thesis, an integrated AI-driven framework is presented to enable autonomous guidewire navigation within complex vascular environments, addressing critical limitations in data availability, simulation fidelity, and navigational accuracy.

	A high-fidelity, real-time simulation platform, CathSim, is introduced for Reinforcement Learning (RL)-based catheter navigation, featuring anatomically accurate vascular models and contact dynamics. To leverage the multimodal data generated by CathSim, the Expert Navigation Network is developed—an RL-based policy integrating visual, kinematic, and force feedback for autonomous tool control. To address data scarcity, an open-source, bi-planar fluoroscopic dataset, Guide3D, is proposed, comprising over 8,700 annotated images for 3D guidewire reconstruction. Finally, a transformer-based model, SplineFormer, is proposed, which directly predicts guidewire geometry as continuous B-spline parameters, enabling interpretable, real-time navigation.

	The findings demonstrate that the integration of high-fidelity simulation, multimodal sensory fusion, and geometric modelling significantly enhances autonomous endovascular navigation. CathSim proves effective for training RL algorithms with realistic interactions, while the Expert Navigation Network shows superior performance in complex anatomical structures. Guide3D successfully addresses data scarcity challenges, and SplineFormer achieves high-precision guidewire prediction with improved interpretability. These contributions establish a comprehensive foundation for advancing autonomous endovascular interventions, offering a pathway toward safer and more precise minimally invasive procedures.
\end{cabstract}


\tableofcontents
\listoffigures
\listoftables


\clearpage
\singlespacing
\newacronym[longplural={Expert Navigation Networks}]{expert navigation network}{ENN}{Expert Navigation Network}
\newacronym[longplural={Force Prediction Networks}]{force prediction network}{FPN}{Force Prediction Network}
\newacronym{3d-fgrn}{3D-FGRN}{3D Fluoroscopy Guidewire Reconstruction Network}

\newacronym[longplural={Success Weighted by Normalized Inverse Path Lengths}]{spl}{SPL}{Success Weighted by Normalized Inverse Path Length}
\newacronym[longplural={Mean Squared Errors}]{mean squared error}{MSE}{Mean Squared Error}
\newacronym[longplural={Binary Cross-Entropies}]{binary cross-entropy}{BCE}{Binary Cross-Entropy}
\newacronym[longplural={Maximum Euclidean Distances}]{maxed}{MaxED}{Maximum Euclidean Distance}
\newacronym{mete}{METE}{Mean Error in Tip Tracking}
\newacronym{mers}{MERS}{Mean Error related to the Robot's Shape}
\newacronym[longplural={Dice Coefficients}]{dice coefficient}{DiceM}{Dice Coefficient}
\newacronym{mIoU}{mIoU}{Intersection over Union}

\newacronym{artificial intelligence}{AI}{Artificial Intelligence}
\newacronym{machine learning}{ML}{Machine Learning}
\newacronym{deep learning}{DL}{Deep Learning}
\newacronym{reinforcement learning}{RL}{Reinforcement Learning}
\newacronym{learning from demonstration}{LfD}{Learning from Demonstration}
\newacronym{imitation learning}{IL}{Imitation Learning}
\newacronym{degree of autonomy}{DoA}{Degree of Autonomy}
\newacronym{backpropagation}{BP}{Backpropagation}
\newacronym{natural language processing}{NLP}{Natural Language Processing}
\newacronym{model predictive control}{MPC}{Model Predictive Control}
\newacronym{probabilistic ensembles with trajectory sampling}{PETS}{Probabilistic Ensembles with Trajectory Sampling}
\newacronym{large language models}{LLMs}{Large Language Models}
\newacronym{large vision models}{LVMs}{Large Vision Models}

\newacronym[longplural={Convolutional Neural Networks}]{cnn}{CNN}{Convolutional Neural Network}
\newacronym[longplural={Fully Convolutional Networks}]{fully convolutional network}{FCN}{Fully Convolutional Network}
\newacronym[longplural={Fully Connected Networks}]{fully connected network}{FC}{Fully Connected Network}
\newacronym[longplural={Recurrent Neural Networks}]{recurrent neural network}{RNN}{Recurrent Neural Network}
\newacronym[longplural={Artificial Neural Networks}]{artificial neural network}{ANN}{Artificial Neural Network}
\newacronym[longplural={Long Short-Term Memories}]{long short-term memory}{LSTM}{Long Short-Term Memory}
\newacronym[longplural={Vision Transformers}]{vit}{ViT}{Vision Transformer}
\newacronym[longplural={Feed-Forward Networks}]{feed forward network}{FFN}{Feed-Forward Network}
\newacronym[longplural={Multi-Head Self-Attentions}]{multi-head self-attention}{MHA}{Multi-Head Self-Attention}
\newacronym{mlp}{MLP}{Multi-Layered Perceptron}
\newacronym{adam}{ADAM}{Adaptive Moment Estimation}
\newacronym{gru}{GRU}{Gated Recurrent Unit}
\newacronym[longplural={Deep Neural Networks}]{deep neural network}{DNN}{Deep Neural Network}

\newacronym{resnet}{ResNet}{Residual Network}
\newacronym{bert}{BERT}{Bidirectional Encoder Representations from Transformers}

\newacronym{relu}{ReLU}{Rectified Linear Unit}
\newacronym{lrelu}{Leaky ReLU}{Leaky Rectified Linear Unit}
\newacronym{elu}{ELU}{Exponential Linear Unit}

\newacronym[longplural={Partially Observable Markov Decision Processes}]{partially observable markov decision process}{POMDP}{Partially Observable Markov Decision Process}
\newacronym[longplural={Markov Decision Processes}]{markov decision process}{MDP}{Markov Decision Process}
\newacronym[longplural={Dueling Deep Q-Networks}]{dueling deep Q-networks}{DDQN}{Dueling Deep Q-Network}
\newacronym[longplural={Deep Q-Networks}]{deep q-network}{DQN}{Deep Q-Network}
\newacronym[longplural={Generative Adversarial Imitation Learnings}]{generative adversarial imitation learning}{GAIL}{Generative Adversarial Imitation Learning}
\newacronym{deep reinforcement learning}{DRL}{Deep Reinforcement Learning}
\newacronym{ppo}{PPO}{Proximal Policy Optimization}
\newacronym{sac}{SAC}{Soft Actor Critic}
\newacronym{behavioural cloning}{BC}{Behavioural Cloning}
\newacronym{inverse reinforcement learning}{IRL}{Inverse Reinforcement Learning}
\newacronym{reinforce}{REINFORCE}{REward Increment = Nonnegative Factor $\times$ Offset Reinforcement $\times$ Characteristic Eligibility}

\newacronym[longplural={Degrees of Freedom}]{degrees of freedom}{DoF}{Degree of Freedom}
\newacronym[longplural={Electromagnetics}]{electromagnetic}{EM}{Electromagnetic}

\newacronym{virtual reality}{VR}{Virtual Reality}
\newacronym{artificial reality}{AR}{Artificial Reality}

\newacronym[longplural={Minimally Invasive Surgeries}]{minimally invasive surgery}{MIS}{Minimally Invasive Surgery}
\newacronym{cardiovascular}{CV}{Cardiovascular}
\newacronym{pulmonary vein isolation}{PVI}{Pulmonary Vein Isolation}
\newacronym[longplural={Mechanical Thrombectomies}]{mechanical thrombectomy}{MT}{Mechanical Thrombectomy}
\newacronym[longplural={Percutaneous Coronary Interventions}]{percutaneous coronary intervention}{PCI}{Percutaneous Coronary Intervention}
\newacronym{cardiovascular diseases}{CVDs}{Cardiovascular Diseases}
\newacronym{bca}{BCA}{Brachiocephalic Artery}
\newacronym{lcca}{LCCA}{Left Common Carotid Artery}

\printglossary[type=\acronymtype, style=index]


\onehalfspacing
\chapter{Introduction}
\chaptermark{Introduction}
\glsresetall


\gls{cardiovascular diseases} remain one of the foremost global health challenges, accounting for a substantial proportion of morbidity and mortality worldwide. Over the past few decades, significant advancements in medical technology and minimally invasive interventions have transformed the clinical landscape, offering improved patient outcomes through endovascular procedures. Nevertheless, despite their clinical efficacy, these interventions continue to present substantial procedural risks, technical challenges, and operational burdens on healthcare providers.

In this thesis, the integration of \gls{artificial intelligence} into robotic systems is investigated to enhance the precision and autonomy of endovascular interventions. The focus is placed on the development of advanced learning frameworks, high-fidelity simulation environments, and robust modelling techniques that collectively aim to address the persistent challenges of autonomous endovascular tool navigation. By employing cutting-edge methods such as \gls{reinforcement learning}, multimodal sensory integration, and geometric modelling, efforts are directed toward advancing the frontiers of autonomous endovascular robotics, with implications extending beyond navigation to broader domains within surgical robotics and interventional medicine.

The remainder of this chapter is structured as follows. In Section~\ref{ch1sec:motivation}, the motivation behind pursuing autonomy in endovascular procedures is discussed, highlighting both the clinical significance and current limitations of existing robotic systems. In Section~\ref{ch1sec:thesis_scope_objectives}, the scope and objectives of this thesis are outlined, framing the research questions that guide its contributions. Section~\ref{ch1sec:thesis_contributions} provides a summary of the key technical innovations and datasets developed throughout this work. Finally, the chapter is concluded with a detailed outline of the thesis organization in Section~\ref{ch1sec:outline}. 

\section{Motivation}\label{ch1sec:motivation}

\acrfull{cardiovascular diseases} are the leading cause of mortality worldwide, impacting millions each year. Conditions like coronary heart disease and cerebrovascular disease are among the most significant contributors to CV-related deaths~\autocite{townsend2016cardiovascular}.
Treatment options for these diseases have evolved significantly, with minimally invasive endovascular interventions, such as \gls{percutaneous coronary intervention}, \gls{pulmonary vein isolation}, and \gls{mechanical thrombectomy}, now forming critical components of care~\autocite{thukkani2015endovascular, goyal2016endovascular, giacoppo2017percutaneous}.
These procedures allow vascular obstructions or arrhythmias to be treated by navigating \textit{catheters} and \textit{guidewires} from an access point, such as the \textit{femoral artery} or \textit{radial artery}, to specific targets within the vascular system, typically under fluoroscopic guidance~\autocite{brilakis2020manual}.
Once the target is reached, interventions like thrombus removal, stent deployment, or tissue ablation can be performed.

However, despite their efficacy, these procedures are associated with risks and complications.
Fluoroscopy use, for instance, results in exposure of both patients and healthcare practitioners to ionizing radiation, with cumulative exposure increasing the risk of adverse health effects, including cancer~\autocite{klein2009occupational}.
Furthermore, repeated use of contrast dye during imaging has been associated with nephrotoxicity, particularly in vulnerable patients~\autocite{rudnick1995nephrotoxicity}.
From a procedural standpoint, complications such as vessel perforation, thrombosis, or distal embolization continue to present challenges, especially in acute cases where timely and precise intervention is critical~\autocite{hausegger2001complications}.
Time remains particularly crucial in stroke interventions, as the benefits of procedures like \gls{mechanical thrombectomy} diminish significantly if not performed promptly~\autocite{saver2016time}.

In recent years, robotic assistance has been introduced in endovascular procedures to address some of these challenges. The potential of robotics to improve precision, reduce operator fatigue, and lower radiation exposure has been demonstrated by enabling procedures to be controlled from a safer distance~\autocite{ho2007ionizing, madder2017impact}.
Systems such as the Magellan~\textsuperscript{TM} and CorPath GRX~\textsuperscript{\textregistered} have shown promise in procedures like \gls{percutaneous coronary intervention} and \gls{pulmonary vein isolation}, offering enhanced control and stability for catheter navigation~\autocite{duran2014randomized, nogueira2020robotic}.
By eliminating the need for operators to be positioned directly next to patients, robotic systems also reduce reliance on heavy protective lead aprons, which are known to contribute to orthopaedic issues over time for medical staff~\autocite{monaco2020anti}.

Nevertheless, limitations are present in current robotic systems.
A \textit{leader-follower} setup is typically employed, in which complete control over the robot’s movements is retained by the physician, resulting in \textit{high cognitive demands} and the potential for human error~\autocite{mofatteh2021neurosurgery}.
Additionally, most robotic systems lack haptic feedback, which is critical for enabling the interaction between the catheter and vessel walls to be felt, thereby making navigation more challenging~\autocite{crinnion2022robotics}.
The interfaces of these systems often rely on input devices (\ie, joysticks or buttons), requiring a skill set distinct from traditional clinical practices and potentially slowing the adoption of robotic technology in endovascular settings.

To address these limitations, there is growing interest in integrating \gls{artificial intelligence} with robotic systems to achieve greater autonomy in endovascular tool navigation.
\gls{artificial intelligence}, particularly through \gls{machine learning} and \gls{reinforcement learning}, has shown promise in enhancing \textit{decision-making}, \textit{data analysis}, and \textit{pattern recognition} in various healthcare applications~\autocite{sarker2021machine, fatima2017survey}.
\gls{reinforcement learning}, which enables agents to learn optimal actions through interactions with the environment, is especially relevant for autonomous navigation tasks where the system can adapt through a process similar to human \textit{trial-and-error} learning~\autocite{sutton2018reinforcement, arulkumaran2017deep}.
In the context of endovascular procedures, \gls{artificial intelligence}-driven autonomous systems could reduce operator dependence, enhance navigation precision, and ultimately expand the accessibility and safety of these critical interventions.

While \gls{artificial intelligence}-driven autonomy holds great promise for endovascular tool navigation, the field faces significant challenges due to the limited availability of high-quality data and suitable simulation environments. \gls{machine learning} algorithms, particularly those used for tasks like segmentation and object recognition, depend heavily on large, annotated datasets to capture the variability in vascular anatomy and pathological states. However, in the medical domain, such datasets are scarce, creating a bottleneck for algorithmic development.
In \gls{reinforcement learning}, which requires extensive sampling to learn optimal navigation strategies, the need for realistic and responsive simulation environments is paramount. Simulation platforms such as SOFA~\cite{faure2012sofa}, commonly used in medical contexts, are not fully optimized for \gls{reinforcement learning}, as they lack the flexibility and efficiency necessary for high-throughput sampling and training. This gap underscores the dependency of \gls{reinforcement learning} on specialized simulation tools that can accurately model the dynamics of catheter interactions within complex vascular structures.
These limitations in both data availability for \gls{machine learning} and suitable simulation environments for \gls{reinforcement learning} present critical challenges in developing robust, autonomous navigation systems. Bridging these gaps is essential to advancing algorithmic precision and adaptability in autonomous endovascular interventions.

Given these advancements and challenges, this thesis aims to advance autonomous catheter navigation by exploring and developing \gls{artificial intelligence}-driven methods that improve the precision, adaptability, and overall performance of robotic systems in complex vascular environments.
By addressing existing limitations in navigation accuracy, and procedural safety, this work contributes to the foundation for future innovations in minimally invasive, autonomous endovascular procedures.

\section{Thesis Scope and Objectives}\label{ch1sec:thesis_scope_objectives}

The limitations in autonomous catheter navigation primarily arise from heuristic-driven controls, which restrict adaptability in complex vascular conditions, and from \gls{reinforcement learning}'s data dependency in endovascular settings. The primary goal of this thesis is to establish a robust, simulation-based framework that allows \gls{reinforcement learning} models to train efficiently, achieving real-time, accurate navigation within complex vascular structures.

This thesis proposes a new approach to autonomous endovascular navigation, leveraging high-fidelity simulation, multimodal sensory integration, and advanced geometric modelling. By creating a modular simulation environment, this work improves data accessibility, diversifies sensory inputs, and enables precise navigation. These contributions bridge the gap between simulation and clinical application, promoting safer and more effective minimally invasive procedures.

The following research questions guide this thesis:

\begin{itemize}
	\item How can high-fidelity simulation environments improve \gls{reinforcement learning} training for endovascular navigation and its subproblems?
	\item In what ways can multimodal sensory integration enhance navigation accuracy and reduce reliance on manual control?
	\item How can shape inference assist navigation?
	\item How can advanced geometric modelling improve the robustness and interpretability of guidewire navigation across various imaging conditions?
\end{itemize}

To address these questions, this thesis pursues several key objectives:

\begin{enumerate}
	\item Review and analyse existing methods: Conduct an in-depth literature review on autonomous catheter navigation, \gls{reinforcement learning} in medical applications, and high-fidelity simulation, identifying current limitations and opportunities.
	\item Develop a real-time simulation environment: Create the CathSim simulator, supporting \gls{reinforcement learning} training with realistic vascular models, a follower robot, and accurate catheter dynamics.
	\item Design a multimodal learning framework: Develop the \gls{expert navigation network} within CathSim, integrating additional sensory inputs, such as force feedback and shape representations, to support efficient learning and real-world applicability.
	\item Create a high-resolution endovascular dataset: Develop Guide3D, an open-source dataset for segmentation and 3D reconstruction tasks, addressing data limitations in endovascular imaging and supporting algorithm development.
	\item Implement a transformer-based navigation model: Introduce SplineFormer, a B-spline transformer model for real-time prediction of guidewire geometry, enhancing interpretability and enabling navigation across diverse imaging conditions.
\end{enumerate}

Through these contributions, this thesis aims to advance autonomous catheter navigation via high-fidelity simulation, integrated sensory feedback, and refined geometric modelling. Though focused on endovascular navigation, the methods developed here hold promise for broader applications in surgical robotics and other minimally invasive procedures where precision and adaptability are essential.

\section{Thesis Contributions}\label{ch1sec:thesis_contributions}

Existing approaches to autonomous catheter navigation often rely on heuristic-driven controls, limiting adaptability to complex vascular environments and leading to suboptimal performance. Additionally, \gls{reinforcement learning} methods require extensive data that is challenging to gather in clinical settings. This thesis addresses these gaps by developing an integrated framework to enhance simulation fidelity, multimodal data availability, and navigational accuracy.

This work introduces CathSim, a real-time simulation environment featuring \textit{anatomically accurate} aortic models and precise contact dynamics, supporting efficient \gls{reinforcement learning} training. Building directly on this simulation infrastructure, the \gls{expert navigation network} leverages CathSim-generated multimodal data to enhance navigation precision. To further support algorithm development and validation, Guide3D provides a high-resolution, open-source dataset for segmentation and 3D reconstruction, addressing data scarcity in endovascular imaging. Finally, SplineFormer, a B-spline transformer model that builds upon these foundational contributions, predicts guidewire geometry with high interpretability, enabling robust, real-time navigation across diverse imaging conditions.

Together, these interconnected contributions provide a comprehensive foundation for advancing autonomous endovascular navigation, improving safety, precision, and adaptability in minimally invasive interventions.

\clearpage
\subsection{CathSim}

\begin{figure}[ht]
	\centering
	\includegraphics[width=.9\linewidth]{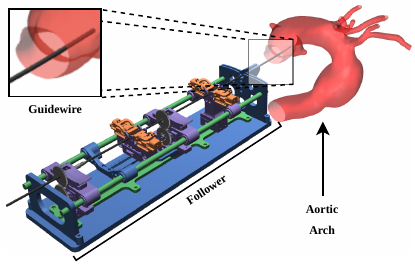}
	\caption[An overview of CathSim]{An overview of CathSim. \copyright 2024 IEEE}\label{fig:cathsim-overview}
\end{figure}

Existing autonomous catheter navigation solutions predominantly use heuristic-driven controls, which lack adaptability in complex vascular environments, resulting in suboptimal cannulation performance. While \gls{reinforcement learning} methods offer potential improvements, they require substantial data for efficacy, and existing simulation environments (\eg, SOFA) prioritize realism over \gls{reinforcement learning} integration.
To address this gap, \textit{CathSim} is introduced as a high-fidelity simulation environment tailored for real-time interaction and optimized for training advanced learning algorithms. \textit{CathSim} comprises three main components: \textit{i)} a follower robot model, \textit{ii)} anatomically accurate aortic arch phantoms, and \textit{iii)} a catheter\/guidewire model with realistic curvature dynamics, as illustrated in Fig.~\ref{fig:cathsim-overview}.

MuJoCo's rigid body dynamics~\cite{todorov2012mujoco} are leveraged to calculate contact interactions between the catheter and aortic walls. Each contact point is computed based on spatial coordinates and contact normals, ensuring realistic frictional forces and efficient resolution. This model enables precise force dynamics without complex deformation calculations, providing a streamlined yet high-fidelity simulation environment for advancing autonomous endovascular navigation systems.

\clearpage
\subsection{Expert Navigation Network}

\begin{sidebysidefigures}[.53]
	\begin{leftfigure}
		\centering
		\includegraphics[width=\textwidth]{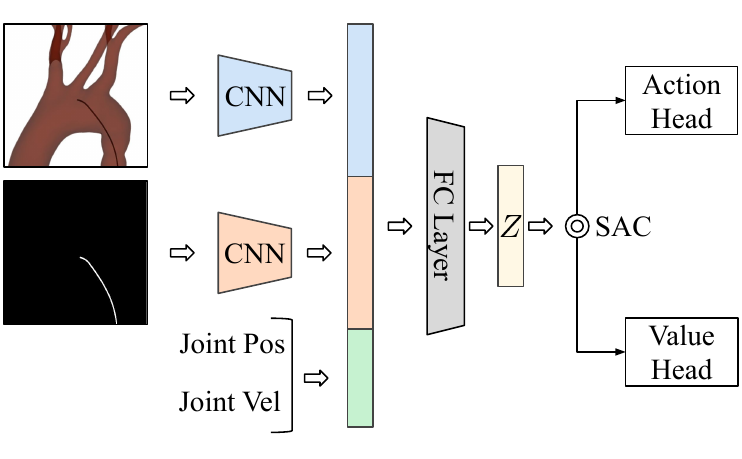}
		\captionof{figure}[Expert Navigation Network]{The Expert Navigation Network processes multimodal inputs (segmented images, joint states, camera views) through dedicated feature extractors (CNN, MLP) into a SAC policy for navigation actions.}
		\label{ch1fig:autocath_expert_model}
	\end{leftfigure}%
	\begin{rightfigure}
		\centering
		\includegraphics[width=\textwidth]{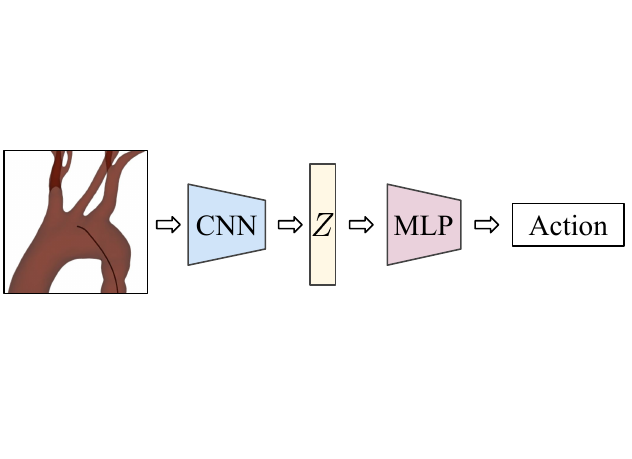}
		\captionof{figure}[Imitation Learning Network]{The downstream Imitation Learning Network distils knowledge from the Expert Navigation Network into a lightweight model for real-time deployment using behavioural cloning.}
		\label{ch1fig:autocath_behavioural_clonning_model}
	\end{rightfigure}
\end{sidebysidefigures}

Current endovascular navigation techniques rely primarily on limited real-world sensory inputs, such as X-ray images, which constrain the ability to gather diverse, multimodal data necessary for effective autonomous navigation.
To overcome these limitations, the \acrfull{expert navigation network} is proposed as a multimodal learning framework trained on extensive labelled data generated within the CathSim simulation environment. By leveraging CathSim, the \gls{expert navigation network} incorporates additional sensor modalities --- such as force feedback and shape representations --- that are generally unavailable in clinical settings, thereby enhancing navigation precision and reducing reliance on manual controls.

The \gls{expert navigation network} is designed to process multiple input modalities, including guidewire segmentation, joint position, joint velocity, and camera images. Dedicated feature extractors (\glspl{cnn} for visual data and an \gls{mlp} for joint data) are used to build a robust navigation representation. The resulting concatenated feature vector is then fed into a \gls{sac} policy, which maps these combined features to precise navigation actions, as illustrated in Fig.~\ref{ch1fig:autocath_expert_model}. To facilitate real-world deployment, the learned policy is distilled into a downstream imitation learning network, shown in Fig.~\ref{ch1fig:autocath_behavioural_clonning_model}, which enables efficient execution with reduced computational requirements. Multimodal integration within the \gls{expert navigation network} substantially improves autonomous catheterization by enabling precise trajectory planning and effective simulation-to-reality transfer, thereby paving the way for real-time surgical support and fully autonomous catheter navigation.

\clearpage
\subsection{3D Reconstruction with stereo X-Ray}

\begin{figure}[ht]
	\centering
	\includegraphics[width=.7\linewidth]{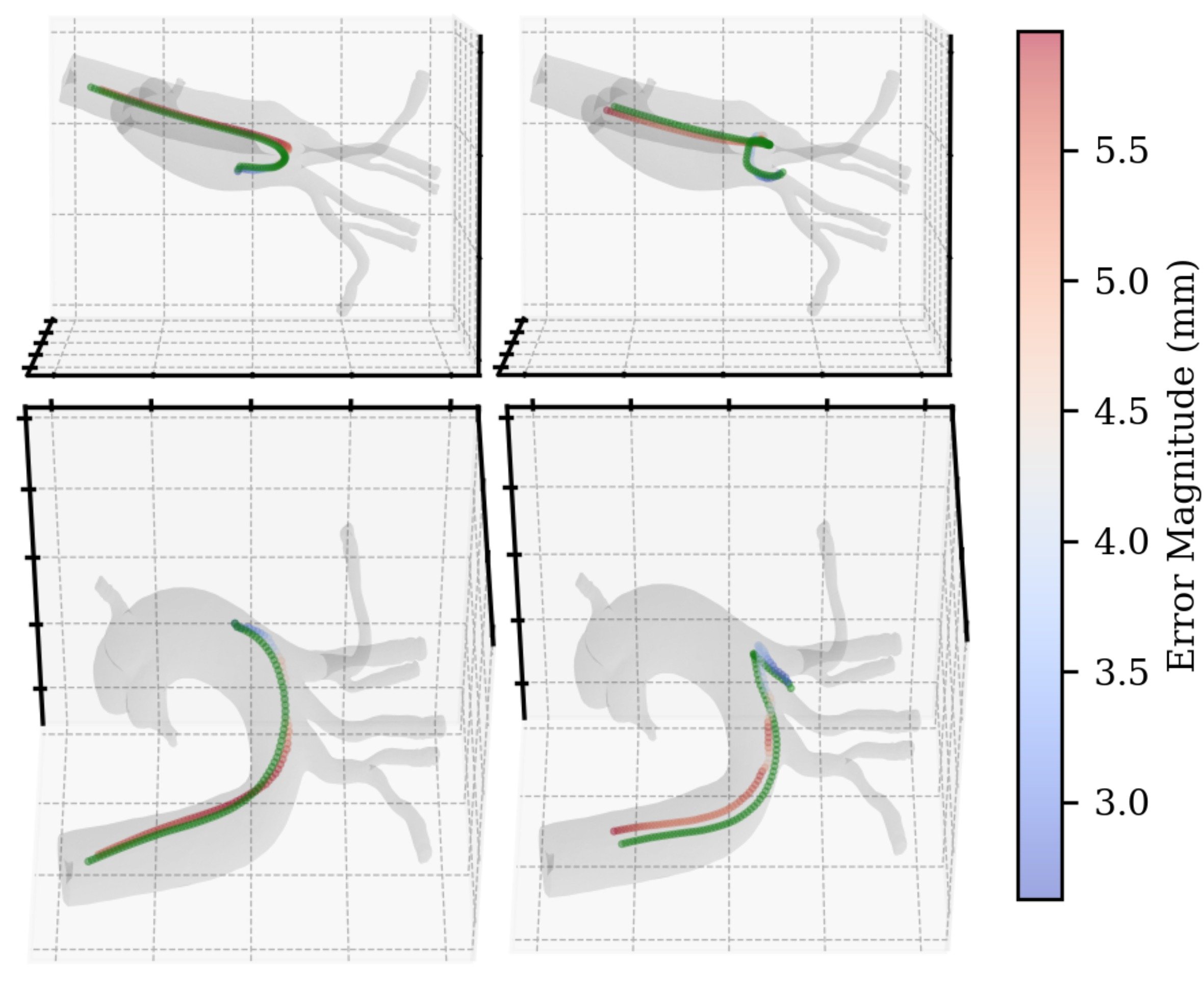}
	\caption{Guidewire Shape Predictions}\label{fig:guidewire_shape_predictions_3D}
\end{figure}

Existing endovascular imaging datasets lack the detailed segmentation and 3D reconstruction data essential for developing advanced algorithms, especially in realistic bi-planar configurations.
To address this limitation, \textit{Guide3D} is introduced as a high-resolution, open-source dataset tailored for segmentation and 3D reconstruction in endovascular imaging. Figure~\ref{fig:guidewire_shape_predictions_3D} illustrates the 3D representation and predictive capabilities. Guide3D contains detailed annotations for two distinct guidewires, offering over \num{8700} annotated images across varied flow conditions. By providing bi-planar images, Guide3D enables accurate 3D guidewire geometry reconstruction, fulfilling a critical need for reliable benchmarks in the field.

Guide3D incorporates a rigorous calibration process with high-fidelity anatomical models and precise X-ray imaging to ensure consistent image quality and anatomically accurate representations. Advanced undistortion and calibration techniques support robust 3D reconstruction via triangulation, enabling thorough algorithm comparisons for both segmentation and dynamic, video-based methods that leverage temporal dimensions for enhanced tracking and analysis. This versatility makes Guide3D a vital resource for advancing and benchmarking endovascular imaging algorithms, offering a foundational tool for the research community.

\clearpage
\subsection{SplineFormer}

\begin{figure}[ht]
	\centering
	\includegraphics[width=.9\linewidth]{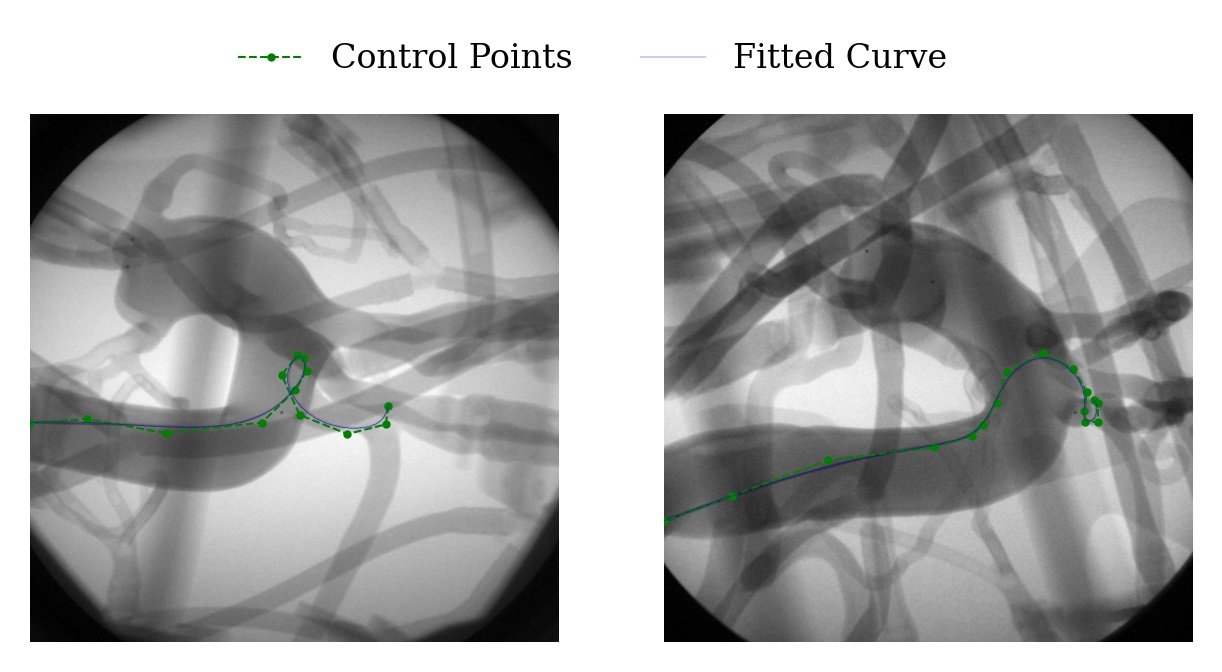}
	\caption[B-Spline Guidewire Representation]{B-spline Representation of a Guidewire}\label{fig:gudiewire_bspline_representation}
\end{figure}

Existing approaches to guidewire navigation often rely on segmentation-based methods, which struggle to accurately model the guidewire's curvilinear structure, leading to suboptimal navigation accuracy.

To address this challenge, \textit{SplineFormer} is introduced as a B-spline transformer model that directly predicts the guidewire’s shape as a set of continuous B-spline parameters, as illustrated in Figure~\ref{fig:gudiewire_bspline_representation}. This model provides a compact, interpretable representation that captures the guidewire's geometry within a constrained parametric space, enhancing both precision and interpretability in endovascular navigation.

Additionally, SplineFormer incorporates a dedicated tip predictor module that identifies the guidewire’s starting position, enabling precise, autonomous navigation in real-time applications. By leveraging the guidewire’s inherent curvature, SplineFormer achieves robust performance, reducing reliance on segmentation pre-processing. In experimental validation using a high-fidelity vascular phantom equipped with a bi-planar X-ray imaging system and robotic manipulation, SplineFormer achieved fully autonomous guidewire control, demonstrating substantial improvements in endovascular navigation tasks. This approach offers a reliable, dimensionally consistent representation that maintains accuracy even in the presence of noise or imaging artefacts, advancing automation and safety in minimally invasive interventions.

\section{List of Publications}

The contributions are summarized as following:

\begin{localbib}
	\nocite{jianu2024guide3d,jianu2024autonomous, jianu2024cathsim, jianu20233d, jianu2024deepwire, jianu2025splineformer, kang2024translating, do2025fedefm,kang2024translating, }
\end{localbib}

\section{Outline}\label{ch1sec:outline}

The primary focus of this thesis is to advance autonomy in endovascular navigation. The structure of this dissertation is as follows:

\begin{itemize}
	\item \emph{Chapter 2} presents the technical background underlying this work, focusing on \textit{deep learning}, \textit{reinforcement learning}, and \textit{autonomous robotic} navigation. It introduces key network architectures and control frameworks, reviews existing methods in endovascular navigation and guidewire segmentation, and establishes the context for the contributions developed in this thesis.

	\item \emph{Chapter 3} presents \textbf{CathSim}, a novel simulation environment designed to expedite the development and testing of navigation algorithms. As an open-source tool, CathSim enables rapid algorithmic prototyping and dataset generation, supporting reproducibility and ease of use. This simulation environment serves as a foundational resource for endovascular navigation research.

	\item \emph{Chapter 4} introduces the \textbf{Expert Navigation Network}, which leverages the extensive dataset generated within CathSim to enable precise navigation within complex aortic environments. This network effectively utilizes the internal data from CathSim to replicate expert-level navigation, outperforming traditional methods in accuracy and robustness within simulated aortic pathways.

	\item \emph{Chapter 5} discusses the development of \textbf{Guide3D}, a pioneering open-source dataset specifically designed for guidewire segmentation and 3D reconstruction. Additionally, this chapter presents a guidewire reconstruction method that surpasses existing techniques by providing higher accuracy and reliability in 3D vascular modelling.

	\item \emph{Chapter 6} introduces \textbf{SplineFormer}, a transformer-based network specifically tailored for predicting complex, arbitrary-shaped splines essential for guidewire navigation. SplineFormer is implemented and validated in real-world scenarios, demonstrating substantial improvement in spline prediction accuracy and adaptability over traditional spline modelling approaches.

	\item \emph{Chapter 7} synthesizes the contributions of this work, summarizing key findings and insights. This chapter also discusses potential future research directions to further enhance autonomous endovascular navigation.
\end{itemize}

\onehalfspacing
\chapter{Background}
\chaptermark{Background}
\glsresetall

Autonomous endovascular navigation, situated at the intersection of medical robotics, \gls{artificial intelligence}, and interventional radiology, addresses the challenge of navigating complex vasculature with minimal human intervention. Achieving full autonomy requires advances in \gls{deep learning}, \gls{reinforcement learning}, computer vision, and robotics.

This chapter presents the core algorithmic foundations that enable autonomous endovascular navigation. It reviews \gls{deep learning} architectures, including \glspl{cnn}, \glspl{recurrent neural network}, and Transformers, for perception and control, and introduces \acrfull{reinforcement learning} as a framework for sequential decision-making under uncertainty. Learning-based approaches to guidewire control, vessel segmentation, and navigation, as well as their integration with robotic systems, are also examined. These foundations support the developments described in subsequent chapters.

\section{Deep Learning Fundamentals}

\acrlong{deep learning} has emerged as a paradigm-shifting approach within the domain of \gls{artificial intelligence}, driving significant advancements across diverse fields, including computer vision, \gls{natural language processing}, healthcare, and autonomous systems. Unlike conventional \gls{machine learning} methodologies, which often necessitate extensive feature engineering, \gls{deep learning} models utilize hierarchically structured \glspl{artificial neural network} to autonomously extract and learn intricate patterns from high-dimensional data~\cite{lecun2015deep}. This capability has rendered \gls{deep learning} particularly effective for complex tasks where traditional statistical and rule-based approaches often exhibit suboptimal performance.

This chapter provides a systematic exposition of the fundamental principles underlying deep learning, focusing on its theoretical foundations, architectural frameworks, and training methodologies. The discussion begins with an examination of \gls{artificial intelligence}, in which essential components such as perceptrons, activation functions, and the backpropagation algorithm are covered. Subsequently, deep network architectures are analysed, including \gls{cnn}, which have achieved state-of-the-art performance in image analysis; \gls{recurrent neural network}, which excel in sequential data processing; and transformer-based models, which have revolutionized natural language understanding and generation~\cite{schmidhuber2015deep}.

Furthermore, critical aspects of \gls{deep learning} model optimization are explored, encompassing \textit{gradient-based optimization techniques}, regularization strategies, and \textit{transfer learning paradigms}, all of which play a crucial role in improving model convergence, generalization, and robustness~\cite{goodfellow2016deep}. By the conclusion of this chapter, a comprehensive understanding of \gls{deep learning}’s foundational principles and methodologies will have been developed, forming a solid basis for subsequent discussions on its practical implementation and advancements.

\subsection{Neural Networks and Learning Representations}

\acrlongpl{artificial neural network} are computational models inspired by the structure and function of biological neural networks. These models form the foundation of deep learning by enabling hierarchical feature extraction and representation learning. Neural networks consist of interconnected computational units, known as \textit{neurons}, that process and transform input data through weighted connections and non-linear activation functions. The primary objective of a neural network is to approximate a target function $f^*(x)$ using training data $(x, y)$, where $y \approx f^*(x)$. This section provides an overview of fundamental neural network architectures, starting from the perceptron and extending to deep \glspl{mlp}.

\subsection{Perceptron and Multi-Layer Perceptrons (MLPs)}

\paragraph{Perceptron.}
The perceptron, introduced by \textcite{rosenblatt1958perceptron}, is the simplest form of an artificial neuron and functions as binary linear classifier. It maps an input vector $\mathbf{x} = [x_1, x_2, \dots, x_n]$ to an output $y$ using a weighted sum followed by a threshold function:

\begin{equation}
	y = f \left( \sum_{i=1}^{n} w_i x_i + b \right) = f(\mathbf{w}^T \mathbf{x} + b),
\end{equation}

where: $\mathbf{x} \in \mathbb{R}^n$ is the input feature vector, $\mathbf{w} \in \mathbb{R}^n$ represents the weight parameters, $b$ is the bias term, and $f(\cdot)$ is a step function:

\begin{equation}
	f(z) = \begin{cases}
		1, & \text{if } z \geq 0 \\
		0, & \text{otherwise}
	\end{cases}
\end{equation}

Despite its simplicity, the perceptron is limited to solving only linearly separable problems. The XOR problem, as highlighted by \textcite{minsky1988perceptrons}, is a classic example where a single-layer perceptron fails. This limitation motivated the development of \glspl{mlp} with non-linear activation functions.

\paragraph{Multi-Layer Perceptron.}
An \gls{mlp} extends the perceptron by introducing multiple layers of neurons, allowing the network to learn hierarchical and complex representations. An \gls{mlp} consists of at least three layers:

\begin{itemize}
	\item Input layer: Receives raw input data $\mathbf{x}$.
	\item Hidden layers: Intermediate layers where neurons apply weighted transformations and activation functions.
	\item Output layer: Produces the final output $\hat{y}$.
\end{itemize}

The transformation at each neuron in an \gls{mlp} is given by:

\begin{equation}
	\mathbf{a}^{(l)} = \sigma\left(\mathbf{W}^{(l)} \mathbf{a}^{(l-1)} + \mathbf{b}^{(l)}\right),
\end{equation}

where $\mathbf{a}^{(l)}$ is the activation vector of layer $l$, $\mathbf{W}^{(l)}$ is the weight matrix for layer $l$, $\mathbf{b}^{(l)}$ is the bias vector for layer $l$, $\sigma(\cdot)$ is a non-linear activation function, and $l \in \{1, 2, \ldots, L\}$ represents the layer index with $L$ being the total number of layers in the network.

The Universal Approximation Theorem \cite{cybenko1989approximation, hornik1991approximation} states that an \gls{mlp} with at least one hidden layer and a non-linear activation function can approximate any continuous function, making \glspl{mlp} suitable for classification and regression tasks.

\subsubsection{Activation Functions}\label{ch2subsubsec:activation_functions}
Activation functions introduce non-linearity into neural networks, enabling them to model complex decision boundaries. Without non-linearity, an \gls{mlp} with multiple layers would reduce to a single-layer linear model. Common activation functions include:

\paragraph{Sigmoid Activation Function.}
The sigmoid function is defined as:
\begin{equation}
	\sigma(x) = \frac{1}{1 + e^{-x}},
\end{equation}

with derivative:

\begin{equation}
	\sigma'(x) = \sigma(x)(1 - \sigma(x))
\end{equation}

Although sigmoid is useful for probabilistic interpretations, it suffers from the vanishing gradient problem, leading to slow convergence in deep networks.

\paragraph{ReLU Activation Function.}
The \gls{relu} function is widely used in deep learning:
\begin{equation}
	\operatorname{ReLU}(x) = \max(0, x),
\end{equation}
with derivative:
\begin{equation}
	\operatorname{ReLU}'(x) =
	\begin{cases}
		1, & x > 0    \\
		0, & x \leq 0
	\end{cases}
\end{equation}

ReLU mitigates the vanishing gradient problem but introduces the dying ReLU problem, where neurons output zero for negative inputs. Variants like \gls{lrelu} and Parametric Rectified Linear Unit (PReLU) address this issue by allowing small gradients for negative inputs.

\subsubsection{Learning Representations in Deep Networks}

\glspl{deep neural network} build hierarchical representations by stacking multiple layers. Each hidden layer captures a more abstract feature representation of the input data. The depth of the model determines its ability to generalize and approximate complex functions. Given a training dataset $\{(x_i, y_i)\}_{i=1}^{N}$, a \gls{deep neural network} approximates the function $f^*(x)$ such that:

\begin{equation}
	y \approx f^*(x) = f(x; \theta, W) = \phi(x; \theta)W,
\end{equation}
where $\phi(x; \theta)$ represents learned features from input $x$, $W$ is a mapping from features to output space, $\theta$ are parameters of the representation function.

The advantage of \gls{deep learning} over shallow models lies in the ability to learn hierarchical feature representations rather than relying on manually engineered features \cite{lecun2015deep}. This characteristic enables \glspl{deep neural network} to generalize better and achieve state-of-the-art performance across various domains, including computer vision, \gls{natural language processing}, and \gls{reinforcement learning}.

\subsection{Convolutional Neural Networks}

\acrfull{cnn} have become the cornerstone of modern deep learning-based image analysis, demonstrating exceptional performance in tasks such as object detection, facial recognition, and medical image processing~\cite{lecun1998gradient, krizhevsky2012imagenet, he2016deep}. Unlike fully connected neural networks, \glspl{cnn} leverage spatial hierarchies within data by employing convolutional operations, enabling automatic feature extraction without the need for handcrafted descriptors~\cite{lecun2015deep, rawat2017deep}. This architecture consists of three primary components: \textit{convolutional layers}, \textit{non-linearity}, and \textit{pooling}, each of which plays a crucial role in learning hierarchical feature representations~\cite{goodfellow2016deep, dumoulin2016guide}.

\subsubsection{Convolution Operation}

The core operation in \glspl{cnn} is the convolution, which applies a set of learnable filters to the input image to extract spatial features. Given an input image $\mathbf{X} \in \mathbb{R}^{H \times W \times C}$, where $H$, $W$, and $C$ denote the height, width, and number of channels, respectively, the convolutional operation with a filter $\mathbf{K} \in \mathbb{R}^{k \times k \times C}$ produces an output feature map $\mathbf{Y} \in \mathbb{R}^{H' \times W' \times C'}$ given by:

\begin{equation}
	Y_{i,j}^{(m)} = \sum_{c=1}^{C} \sum_{u=1}^{k} \sum_{v=1}^{k} X_{i+u, j+v}^{(c)} K_{u,v}^{(c,m)} + b^{(m)},
\end{equation}

where $i \in \{1, 2, \ldots, H'\}$, $j \in \{1, 2, \ldots, W'\}$ represent the spatial indices of the output feature map, $m \in \{1, 2, \ldots, C'\}$ denotes the output channel index, $H'$ and $W'$ are the height and width of the output feature map, respectively, and $C'$ is the number of output channels (filters).

\subsubsection{Activation Functions}

To introduce non-linearity, \glspl{cnn} employ activation functions such as the \gls{relu} (see Subsec.~\ref{ch2subsubsec:activation_functions}). \gls{relu} prevents vanishing gradient issues observed in sigmoid and tanh activations, facilitating efficient training of deep networks~\cite{nair2010rectified, glorot2011deep}. Alternative activation functions such as \gls{lrelu}~\cite{maas2013rectifier} and Swish~\cite{ramachandran2017searching} have been proposed to further enhance gradient flow and improve feature learning.

\subsubsection{Pooling Layers}

Pooling operations reduce spatial dimensions while preserving critical features, thereby improving computational efficiency and mitigating overfitting~\cite{boureau2010theoretical, scherer2010evaluation}. The most commonly used pooling techniques include max-pooling and average-pooling, defined as:

\begin{equation}
	Y_{i,j}^{(m)} = \max_{(u,v) \in R} X_{i+u, j+v}^{(m)},
\end{equation}

for max-pooling, where $R$ denotes the pooling region, and

\begin{equation}
	Y_{i,j}^{(m)} = \frac{1}{|R|} \sum_{(u,v) \in R} X_{i+u, j+v}^{(m)},
\end{equation}

for average pooling, where $|R|$ is the number of elements in the pooling region. These operations contribute to translational invariance in CNNs~\cite{zeiler2014visualizing}.

\subsubsection{Modern CNNs Architectures}

Several advanced \gls{cnn} architectures have been proposed to enhance feature extraction capabilities. For instance, \gls{resnet} introduces residual connections to mitigate the vanishing gradient problem~\cite{he2016deep}, while Densely Connected Convolutional Network (DenseNet) employs dense connectivity to promote feature reuse~\cite{huang2017densely}. Additionally, EfficientNet utilizes neural architecture search to optimize network depth, width, and resolution~\cite{tan2019efficientnet}. These advancements have significantly improved \gls{cnn} performance in large-scale image recognition benchmarks such as ImageNet~\cite{krizhevsky2012imagenet, russakovsky2015imagenet}.

\subsection{Sequence Modelling in Deep Learning}

Sequence modelling plays a fundamental role in \gls{deep learning}, enabling the processing and prediction of sequential data across various domains, including speech recognition~\cite{graves2013speech}, \gls{natural language processing}~\cite{bahdanau2014neural, vaswani2017attention}, and time-series forecasting~\cite{cho2014properties}. Unlike conventional \glspl{feed forward network}, which assume independence between input samples, sequence models incorporate temporal dependencies, allowing for the capture of dynamic patterns within data. This section explores two major paradigms in sequence modelling: \acrfull{recurrent neural network} and their variants --- \glspl{long short-term memory}, \glspl{gru} --- and Transformer-based architectures, which have revolutionized the field with their self-attention mechanism.

\subsubsection{RNNs, LSTMs, and GRUs}

\acrfull{recurrent neural network} extend traditional neural networks by introducing recurrence, allowing them to maintain a hidden state that evolves over time. Given an input sequence $\mathbf{X} = \{x_1, x_2, \dots, x_T\}$, an \gls{recurrent neural network} computes the hidden state $\mathbf{h}_t$ at time step $t$ using:

\begin{equation}
	\mathbf{h}_t = f\left(\mathbf{W}_h \mathbf{h}_{t-1} + \mathbf{W}_x \mathbf{x}_t + \mathbf{b} \right),
\end{equation}

where $\mathbf{W}_h$ and $\mathbf{W}_x$ are weight matrices, $\mathbf{b}$ is the bias term, and $f(\cdot)$ is an activation function, typically \textit{tanh} or \gls{relu}~\cite{elman1990finding}. The output $\mathbf{y}_t$ at each time step is given by:

\begin{equation}
	\mathbf{y}_t = g(\mathbf{W}_y \mathbf{h}_t + \mathbf{b}_y),
\end{equation}

where $g(\cdot)$ is a task-dependent activation function, such as softmax for classification. While RNNs can model short-term dependencies effectively, they struggle with long-term dependencies due to the vanishing and exploding gradient problems during backpropagation through time~\cite{bengio1994learning}.

\begin{figure}[h]
	\centering
	\includegraphics[width=\linewidth]{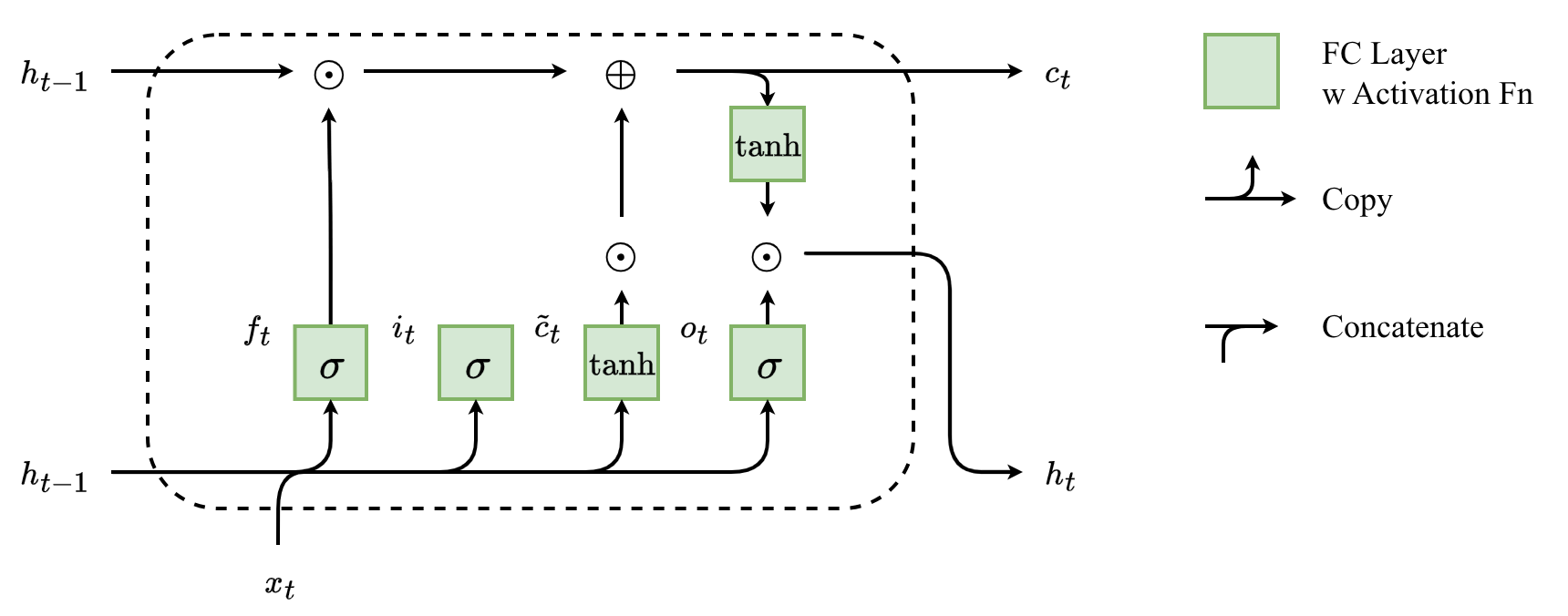}
	\mycaption{LSTM Cell}{LSTM cell architecture showing input, forget, and output gates with cell state and hidden state transitions.}\label{ch2fig:lstm}
\end{figure}

To mitigate these issues, \glspl{long short-term memory} (LSTM) networks introduce a memory cell that allows for more effective gradient flow across long sequences. As shown in Fig.~\ref{ch2fig:lstm}, the LSTM cell includes three gating mechanisms — the forget gate $\mathbf{f}_t$, input gate $\mathbf{i}_t$, and output gate $\mathbf{o}_t$ — that modulate the cell state $\mathbf{c}_t$, enabling the network to retain or discard information as needed. The LSTM dynamics are described by:

\begin{align}
	\mathbf{f}_t         & = \sigma(\mathbf{W}_f [\mathbf{h}_{t-1}, \mathbf{x}_t] + \mathbf{b}_f),          \\
	\mathbf{i}_t         & = \sigma(\mathbf{W}_i [\mathbf{h}_{t-1}, \mathbf{x}_t] + \mathbf{b}_i),          \\
	\mathbf{o}_t         & = \sigma(\mathbf{W}_o [\mathbf{h}_{t-1}, \mathbf{x}_t] + \mathbf{b}_o),          \\
	\tilde{\mathbf{c}}_t & = \tanh(\mathbf{W}_c [\mathbf{h}_{t-1}, \mathbf{x}_t] + \mathbf{b}_c),           \\
	\mathbf{c}_t         & = \mathbf{f}_t \odot \mathbf{c}_{t-1} + \mathbf{i}_t \odot \tilde{\mathbf{c}}_t, \\
	\mathbf{h}_t         & = \mathbf{o}_t \odot \tanh(\mathbf{c}_t),
\end{align}

where $\sigma(\cdot)$ denotes the sigmoid activation, and $\odot$ represents element-wise multiplication. The decoupling of memory and output pathways allows LSTMs to learn long-term dependencies more robustly than standard RNNs.

\begin{figure}[h]
	\centering
	\includegraphics[width=\linewidth]{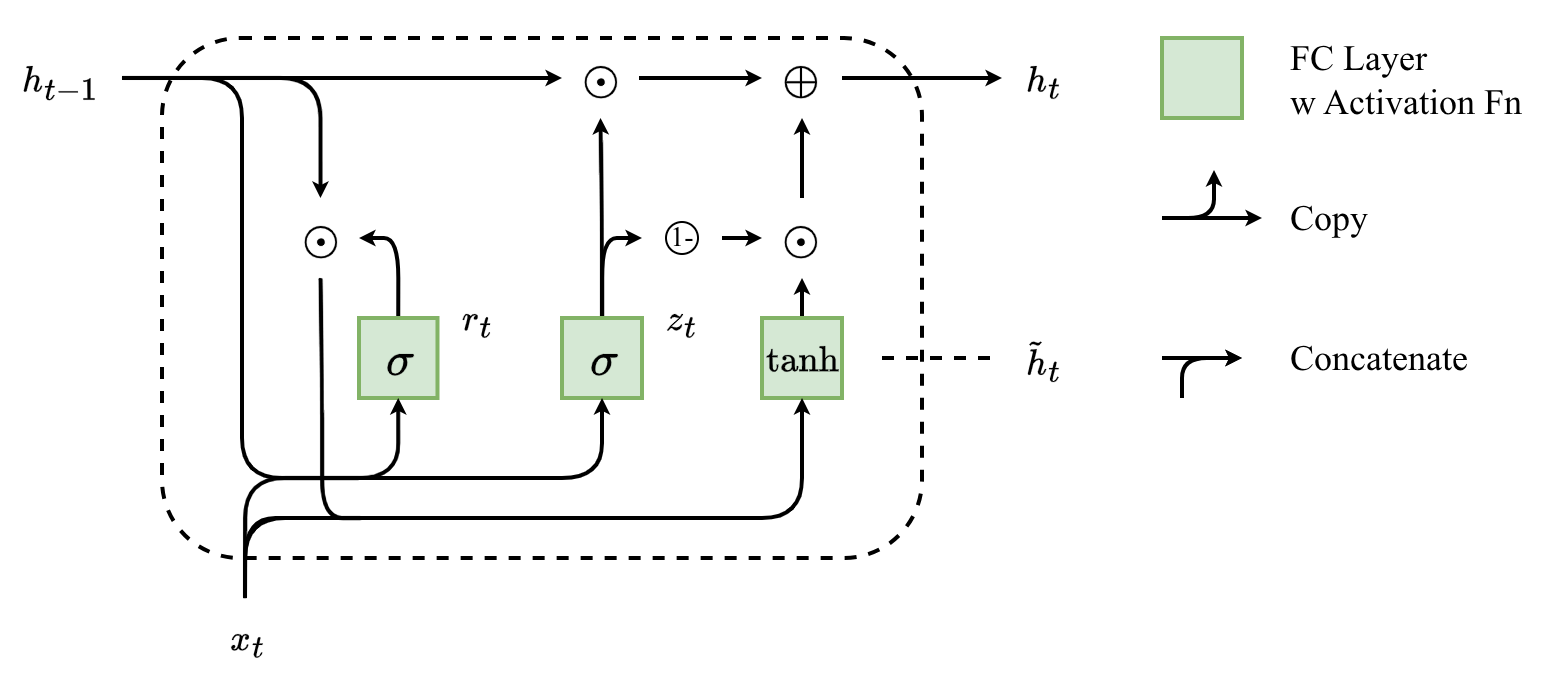}
	\mycaption{GRU Cell}{GRU cell architecture showing update and reset gates with simplified state transitions compared to LSTM.}\label{ch2fig:gru}
\end{figure}

A computationally efficient alternative is the \gls{gru}, which simplifies the gating structure by combining the forget and input gates into a single update gate $\mathbf{z}_t$, and using a reset gate $\mathbf{r}_t$ to control the flow of past information. This design reduces the number of parameters while retaining the model’s ability to capture long-term dependencies. The GRU cell, depicted in Fig.~\ref{ch2fig:gru}, is governed by:

\begin{align}
	\mathbf{z}_t         & = \sigma(\mathbf{W}_z [\mathbf{h}_{t-1}, \mathbf{x}_t] + \mathbf{b}_z),                   \\
	\mathbf{r}_t         & = \sigma(\mathbf{W}_r [\mathbf{h}_{t-1}, \mathbf{x}_t] + \mathbf{b}_r),                   \\
	\tilde{\mathbf{h}}_t & = \tanh(\mathbf{W}_h [\mathbf{r}_t \odot \mathbf{h}_{t-1}, \mathbf{x}_t] + \mathbf{b}_h), \\
	\mathbf{h}_t         & = (1 - \mathbf{z}_t) \odot \mathbf{h}_{t-1} + \mathbf{z}_t \odot \tilde{\mathbf{h}}_t.
\end{align}

By eliminating the separate memory cell, \glspl{gru} offer faster training and inference times, making them suitable for resource-constrained or real-time applications.

Both \glspl{long short-term memory} and \glspl{gru} have demonstrated state-of-the-art performance in sequential tasks such as speech recognition~\cite{graves2013speech}, machine translation~\cite{cho2014properties}, and time-series analysis, providing a foundation for later advances in self-attention-based models.

\subsubsection{Transformer Networks and Their Evolution}

\begin{figure}[h]
	\centering
	\includegraphics[width=.5\linewidth]{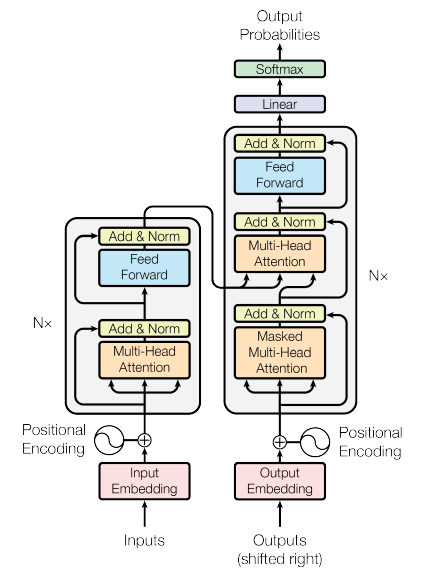}
	\mycaption{Transformer Architecture}{Transformer architecture adapted from~\textcite{vaswani2017attention}.} \label{fig:transformer}
\end{figure}

While \gls{recurrent neural network}-based models have achieved significant success, their sequential nature limits parallelization and scalability. Transformer networks, introduced by \textcite{vaswani2017attention}, overcome these limitations by utilizing self-attention mechanisms instead of recurrence. The core operation in a transformer is the scaled dot-product attention:

\begin{equation}
	\text{Attention}(\mathbf{Q}, \mathbf{K}, \mathbf{V}) = \text{softmax} \left(\frac{\mathbf{Q} \mathbf{K}^T}{\sqrt{d_k}} \right) \mathbf{V},
\end{equation}

where $\mathbf{Q}$, $\mathbf{K}$, and $\mathbf{V}$ represent query, key, and value matrices, respectively, and $d_k$ is the key dimension. This operation enables each token in the input sequence to attend to all other tokens, capturing contextual dependencies regardless of their distance.

The standard transformer architecture (see Figure~\ref{fig:transformer}) is composed of stacked encoder and decoder layers. Each encoder layer contains two main sub-layers: a \gls{multi-head self-attention} mechanism and a position-wise fully connected feedforward network, both surrounded by residual connections and layer normalization. The decoder layers include a third sub-layer that performs multi-head attention over the encoder outputs, enabling effective cross-attention. Positional encodings are added to the input embeddings to inject order information, compensating for the model’s lack of recurrence or convolution.

Transformers have evolved through architectures such as \gls{bert}~\cite{devlin2018bert}, which uses only the encoder for bidirectional language understanding; Generative Pre-trained Transformer (GPT)~\cite{radford2018improving}, which uses only the decoder stack in an autoregressive manner for text generation; and \gls{vit}~\cite{dosovitskiy2020image}, which adapts the encoder for image patches, demonstrating the model’s versatility beyond language tasks.

\subsubsection{Application of Transformers in Medical Image Analysis}

Transformers have recently been applied to medical imaging, particularly in tasks such as tumour segmentation~\cite{hatamizadeh2022unetr}, disease classification~\cite{valanarasu2021medical}, and radiology report generation. \gls{vit} process medical images as patch embeddings, leveraging self-attention to capture spatial relationships without convolutional operations~\cite{dosovitskiy2020image}. Hybrid models, such as Swin Transformers~\cite{liu2021swin}, integrate CNN-like hierarchical structures to enhance feature extraction.

Recent studies show that transformer-based models outperform \glspl{cnn} in various medical imaging benchmarks, highlighting their potential to advance Computer-Aided Diagnosis (CAD)~\cite{gao2021utnet}. However, challenges such as data efficiency, high computational cost, and interpretability remain active areas of research.

\subsection{Backpropagation and Training}

\glspl{mlp} and \glspl{cnn} are trained using Stochastic Gradient Descent (SGD) or its variants, with the loss function typically defined as categorical cross-entropy for classification tasks:

\begin{equation}
	\mathcal{L} = - \sum_{i=1}^{N} y_i \log \hat{y}_i,
\end{equation}

where $y_i$ is the ground truth label, $\hat{y}_i$ is the predicted probability, and $N$ is the number of classes. The gradients of the loss function with respect to the network parameters are computed using backpropagation~\cite{rumelhart1986learning}. Optimization algorithms such as \gls{adam}~\cite{kingma2014adam} and Root Mean Square Propagation (RMSProp)~\cite{tieleman2012rmsprop} are commonly employed to improve convergence.

\subsection{Reinforcement Learning Fundamentals}

\acrfull{reinforcement learning} is a framework for sequential decision-making, where an \textit{agent} learns to interact with an \textit{environment} to maximize cumulative rewards~\cite{sutton2018reinforcement}. Unlike supervised learning, which relies on labelled data, \gls{reinforcement learning} employs a trial-and-error approach, where the agent explores different strategies and refines its policy based on received feedback~\cite{kaelbling1996reinforcement, mnih2015human}. This process is typically formalized using a \gls{markov decision process}, which provides a mathematical foundation for modelling decision-making problems in dynamic environments.

\subsubsection{Markov Decision Processes}

A \acrfull{markov decision process} is a 5-tuple $\mathcal{M} = \langle \mathcal{S}, \mathcal{A}, P, R, \gamma \rangle$, where:

\begin{itemize}
	\item $\mathcal{S}$ is the set of all possible states of the environment.
	\item $\mathcal{A}$ is the set of all possible actions available to the agent.
	\item $P(s' | s, a)$ is the transition probability function, defining the probability of moving to state $s' \in \mathcal{S}$ after taking action $a \in \mathcal{A}$ in state $s$.
	\item $R(s, a)$ is the reward function, specifying the immediate scalar reward received after executing action $a$ in state $s$.
	\item $\gamma \in [0,1]$ is the discount factor, determining the importance of future rewards relative to immediate rewards.
\end{itemize}

The Markov property states that the probability of transitioning to the next state $s_{t+1}$ depends only on the current state $s_t$ and action $a_t$, and not on the history of previous states and actions~\cite{puterman2014markov}:

\begin{equation}
	P(s_{t+1} | s_t, a_t, s_{t-1}, a_{t-1}, \dots, s_0, a_0) = P(s_{t+1} | s_t, a_t).
\end{equation}

Given an \gls{markov decision process}, the agent’s objective is to learn an optimal policy $\pi^*(a | s)$ that maximizes the expected cumulative reward, known as the return:

\begin{equation}
	G_t = \sum_{k=0}^{\infty} \gamma^k r_{t+k+1},
\end{equation}

where $r_{t+k+1}$ is the reward received at time step $t+k+1$ and $\gamma$ ensures that future rewards contribute less significantly than immediate rewards.

\subsubsection{States, Actions, Rewards, and Policies}

\paragraph{States and Observations.}
A state $s_t \in \mathcal{S}$ fully describes the environment at time step $t$. In many real-world scenarios, the agent has access only to a partial observation $o_t$ rather than the full state, leading to a \gls{partially observable markov decision process}~\cite{kaelbling1998planning}. In such cases, the agent maintains a belief distribution ooverstates to make decisions.

\paragraph{Action Space.}
The action space $\mathcal{A}$ defines the set of all possible actions the agent can take. It can be discrete, as in board games where the agent selects a finite set of moves~\cite{silver2016mastering}, or continuous, as in robotic control tasks where actions involve real-valued motor torques~\cite{lillicrap2015continuous}.

\paragraph{Rewards and Return.}
The reward function $R(s, a)$ provides immediate feedback on the desirability of an action in a given state. The agent’s goal is to maximize the expected return, which is either finite-horizon undiscounted:

\begin{equation}
	G_t = \sum_{k=0}^{T} r_{t+k+1},
\end{equation}

or infinite-horizon discounted:

\begin{equation}
	G_t = \sum_{k=0}^{\infty} \gamma^k r_{t+k+1},
\end{equation}

where $\gamma \in (0,1]$ is a discount factor that prioritizes immediate rewards.

\paragraph{Policy Representation.}
A policy $\pi(a | s)$ defines the agent’s strategy by mapping states to actions. Policies can be:

- Deterministic: The agent always selects the same action for a given state:
\begin{equation}
	a_t = \mu(s_t)
\end{equation}

- Stochastic: The agent samples actions from a probability distribution:
\begin{equation}
	a_t \sim \pi(\cdot | s_t)
\end{equation}.

Policy optimization methods in deep reinforcement learning often involve parameterized policies, where a neural network with parameters $\theta$ approximates $\pi_{\theta}(a | s)$~\cite{schulman2017proximal}.

\paragraph{Value Functions.}
Value functions estimate the expected return from a state or state-action pair:

- State-value function:
\begin{equation}
	V^\pi(s) = \mathbb{E}_\pi \left[ G_t \mid s_t = s \right]
\end{equation}

- Action-value function:
\begin{equation}
	Q^\pi(s, a) = \mathbb{E}_\pi \left[ G_t \mid s_t = s, a_t = a \right]
\end{equation}

The Bellman equation expresses the recursive relationship between value functions:

\begin{equation}
	V^\pi(s) = \mathbb{E}_\pi \left[ R(s, a) + \gamma V^\pi(s') \mid s \right]
\end{equation}

The optimal policy $\pi^*$ maximizes the expected return:

\begin{equation}
	\pi^*(s) = \arg\max_{a} Q^*(s, a)
\end{equation}

\subsubsection{Conclusion}

Reinforcement learning provides a mathematical framework for sequential decision-making, where an agent interacts with an environment to maximize cumulative rewards. By formalizing the problem as an \gls{markov decision process}, \gls{reinforcement learning} enables efficient learning through policies, value functions, and reward optimization. Recent advances in deep \gls{reinforcement learning} have extended these principles to complex, high-dimensional problems, such as robotic control~\cite{lillicrap2015continuous}, game playing~\cite{silver2016mastering}, and autonomous decision-making~\cite{mnih2015human}.

\subsection{Policy Learning in Reinforcement Learning}

In \gls{reinforcement learning}, an agent seeks to optimize a \textit{policy}, $\pi(a|s)$, which defines a probability distribution over actions given a state. The primary goal of policy learning is to find an optimal policy $\pi^*$ that maximizes the expected cumulative reward. Policy learning methods can be broadly classified into \textit{model-free} and \textit{model-based} approaches, with further subdivisions into \textit{value-based} and \textit{policy gradient} methods~\cite{sutton2018reinforcement, mnih2015human, silver2014deterministic}. This section provides a formal exposition of these methods and their mathematical underpinnings.

\subsubsection{Model-Free vs. Model-Based RL}

A fundamental distinction in \gls{reinforcement learning} is whether the agent explicitly models the environment dynamics. \textit{Model-free} \gls{reinforcement learning} methods learn a policy or a value function directly from interactions with the environment, without constructing an explicit transition model~\cite{mnih2015human}. These methods typically require large amounts of data but are widely used due to their simplicity and applicability to unknown environments.

Conversely, \textit{model-based} \gls{reinforcement learning} methods attempt to learn a model of the transition dynamics, $P(s' | s, a)$, and the reward function, $R(s, a)$, enabling the agent to simulate and plan ahead~\cite{kaelbling1996reinforcement}. The advantage of model-based methods lies in their sample efficiency, as they leverage the learned model for decision-making without direct environment interaction. However, inaccuracies in the learned model can lead to suboptimal policies due to model bias.

Mathematically, an \gls{reinforcement learning} agent seeks to maximize the expected return:

\begin{equation}
	J(\pi) = \mathbb{E}_{\pi} \left[ \sum_{t=0}^{\infty} \gamma^t R(s_t, a_t) \right],
\end{equation}

where $\gamma \in (0,1]$ is the discount factor ensuring convergence. In model-free \gls{reinforcement learning}, this objective is optimized directly, whereas model-based \gls{reinforcement learning} first estimates the transition model before optimizing $J(\pi)$.

\subsubsection{Value-Based Methods (Q-Learning, Deep Q-Networks)}

Value-based methods focus on estimating the \textit{action-value function}:

\begin{equation}
	Q^\pi(s, a) = \mathbb{E}_\pi \left[ \sum_{t=0}^{\infty} \gamma^t R(s_t, a_t) \mid s_0 = s, a_0 = a \right],
\end{equation}

which represents the expected return from state $s$ after taking action $a$ and following policy $\pi$. The optimal action-value function satisfies the \textit{Bellman optimality equation}~\cite{bellman1957markovian}:

\begin{equation}
	Q^*(s, a) = \mathbb{E} \left[ R(s, a) + \gamma \max_{a'} Q^*(s', a') \mid s, a \right]
\end{equation}

\paragraph{Q-Learning.}
Q-Learning~\cite{watkins1992q} is an off-policy algorithm that approximates $Q^*(s, a)$ using the update rule:

\begin{equation}
	Q(s, a) \leftarrow Q(s, a) + \alpha \left[ r + \gamma \max_{a'} Q(s', a') - Q(s, a) \right],
\end{equation}

where $\alpha$ is the learning rate. This iterative update ensures convergence to the optimal Q-function under mild conditions.

\paragraph{Deep Q-Networks.}
\glspl{deep q-network}~\cite{mnih2015human} extend Q-learning by employing a deep neural network $Q_\theta(s, a)$ to approximate the Q-function, where $\theta$ are the network parameters. The optimization objective minimizes the Temporal Difference (TD) error:

\begin{equation}
	L(\theta) = \mathbb{E} \left[ \left( y_t - Q_\theta(s, a) \right)^2 \right],
\end{equation}

where the target value $y_t$ is given by:

\begin{equation}
	y_t = r + \gamma \max_{a'} Q_{\theta'}(s', a'),
\end{equation}

and $\theta'$ represents the parameters of a slowly updated \textit{target network} to stabilize training. Experience replay is employed to break correlations between consecutive samples, improving convergence.

\subsubsection{Policy Gradient Methods}

Policy gradient methods optimize a parameterized policy $\pi_\theta(a | s)$ by directly maximizing the expected return:

\begin{equation}
	J(\theta) = \mathbb{E}_{\pi_\theta} \left[ \sum_{t=0}^{\infty} \gamma^t R(s_t, a_t) \right]
\end{equation}

The gradient of $J(\theta)$ is computed using the policy gradient theorem~\cite{sutton1999policy}:

\begin{equation}
	\nabla_\theta J(\theta) = \mathbb{E}_{\pi_\theta} \left[ \nabla_\theta \log \pi_\theta(a | s) Q^\pi(s, a) \right]
\end{equation}

\paragraph{REINFORCE Algorithm.}
The \gls{reinforce} algorithm~\cite{williams1992simple} updates the policy parameters using Monte Carlo estimates:

\begin{equation}
	\theta \leftarrow \theta + \alpha \sum_{t=0}^{T} \nabla_\theta \log \pi_\theta(a_t | s_t) G_t,
\end{equation}

where $G_t$ is the empirical return from time $t$ onward. While simple, \gls{reinforce} suffers from high variance in gradient estimates.

\paragraph{Proximal Policy Optimization}
\gls{ppo}~\cite{schulman2017proximal} improves stability by introducing a surrogate objective:

\begin{equation}
	L^{\text{PPO}}(\theta) = \mathbb{E} \left[ \min(r_t(\theta) A_t, \text{clip}(r_t(\theta), 1 - \epsilon, 1 + \epsilon) A_t) \right],
\end{equation}
where $r_t(\theta) = \frac{\pi_\theta(a_t | s_t)}{\pi_{\theta_{\text{old}}}(a_t | s_t)}$ is the probability ratio between new and old policies, $A_t = A(s_t, a_t)$ is the advantage function that measures how much better an action $a_t$ is compared to the average action in state $s_t$, and $\epsilon$ is a clipping parameter to constrain updates.

\paragraph{Soft Actor-Critic.}
\gls{sac}~\cite{haarnoja2018soft} incorporates an entropy term $\mathcal{H}(\pi)$ into the objective:

\begin{equation}
	J(\pi) = \mathbb{E} \left[ \sum_{t=0}^{\infty} \gamma^t (R(s_t, a_t) + \alpha \mathcal{H}(\pi(\cdot | s_t))) \right],
\end{equation}

where $\alpha$ is a temperature parameter controlling the trade-off between exploration and exploitation. \gls{sac} employs an off-policy actor-critic framework, optimizing both the policy and the Q-function.

\subsubsection{Conclusion}

Policy learning in \gls{reinforcement learning} encompasses both model-free and model-based methods, with further distinctions between value-based approaches like Q-learning and \gls{deep q-network}, and policy gradient methods such as \gls{reinforce}, \gls{ppo}, and \gls{sac}. While value-based methods excel in discrete action spaces, policy gradient methods are well-suited for continuous control tasks. Ongoing research continues to refine these approaches, enhancing their stability, sample efficiency, and generalization capabilities~\cite{silver2014deterministic, haarnoja2018soft}.

\subsection{Reinforcement Learning for Robot Control}

\gls{reinforcement learning} has revolutionized control and robotics by enabling autonomous agents to learn complex behaviours through interaction with dynamic environments~\cite{kober2013reinforcement, levine2016end}. Unlike traditional control methods based on predefined models and optimization techniques, \gls{reinforcement learning} provides a data-driven approach to learning policies that map high-dimensional sensory inputs to continuous control actions~\cite{deisenroth2013survey}. This section explores \gls{reinforcement learning} techniques specifically tailored for continuous control tasks and autonomous navigation.

\subsubsection{RL in Continuous Control Tasks}

Control problems in robotics often involve high-dimensional continuous action spaces, where an agent must generate smooth motor commands for tasks such as locomotion and dexterous manipulation~\cite{lillicrap2015continuous}. The challenge lies in efficiently learning policies that generalize well to diverse conditions while ensuring stability and safety.

\paragraph{Optimal Control and RL.}
Classical optimal control relies on solving the Hamilton Jacobi Bellman (HJB) equation:

\begin{equation}
	V^*(s) = \max_{a} \left[ R(s, a) + \gamma \mathbb{E}_{s' \sim P} \left[ V^*(s') \right] \right],
\end{equation}

where $V^*(s)$ is the optimal value function. While Dynamic Programming (DP) methods such as Linear Quadratic Regulator (LQR) solve this equation for linear systems, they do not scale well to high-dimensional robotic tasks. \gls{reinforcement learning} methods overcome this limitation by learning parametric approximations of $V^*(s)$ or directly optimizing policies~\cite{recht2019tour}.

\paragraph{Model-Based RL for Control.}
In robotics, data efficiency is critical, as collecting physical interactions is costly. Model-Based Reinforcement Learning (MBRL) methods improve sample efficiency by learning an approximate transition model $\hat{P}(s' | s, a)$ and using it for planning~\cite{chua2018deep}. Given a learned model, policies can be optimized using \gls{model predictive control}:

\begin{equation}
	a_t = \arg \max_{a} \sum_{k=0}^{T} \gamma^k R(s_{t+k}, a_{t+k}),
\end{equation}

where $T$ is the planning horizon. Hybrid approaches such as Probabilistic Ensembles with Trajectory Sampling (PETS) further improve robustness by modelling uncertainty in $\hat{P}$~\cite{chua2018deep}.

\paragraph{Energy-Based and Safety-Critical RL.}
Safety is paramount in robotic control, requiring constraints on actions and trajectories. Energy-based RL methods incorporate constraints via Control Lyapunov Functions (CLFs) and Control Barrier Functions (CBFs)~\cite{cheng2019end}:

\begin{equation}
	\frac{dV(s)}{dt} + \alpha(V(s)) \leq 0, \quad \forall s \in \mathcal{S},
\end{equation}

where $\alpha(\cdot)$ is a class-$\mathcal{K}$ function ensuring stability. These approaches guarantee that \gls{reinforcement learning} policies respect physical constraints while optimizing for task performance.

\subsubsection{RL for Autonomous Navigation}

Autonomous navigation involves decision-making in dynamic, often partially observable environments. \gls{reinforcement learning}-based approaches leverage spatial reasoning, motion planning, and obstacle avoidance to enable robots to navigate in unstructured environments such as urban streets or indoor spaces~\cite{zhu2017target}.

\paragraph{Policy Learning for Navigation.}
Navigation tasks can be framed as a goal-conditioned \gls{reinforcement learning} problem, where the policy $\pi(a | s, g)$ conditions on a goal state $g$. The reward function is often defined using geodesic distance:

\begin{equation}
	R(s, a) = - d(s, g),
\end{equation}

where $d(s, g)$ is the shortest path distance between the agent's current state and the goal.

\paragraph{Hierarchical RL for Long-Horizon Navigation.}
Hierarchical Reinforcement Learning (HRL) decomposes navigation into high-level planning and low-level control~\cite{nachum2018data}:

\begin{align}
	g_t & = \pi_{\text{high}}(s_t),     \\
	a_t & = \pi_{\text{low}}(s_t, g_t).
\end{align}

This structure enables efficient exploration and generalization across large environments.

\paragraph{Sim-to-Real Transfer in Navigation.}
One challenge in \gls{reinforcement learning}-based navigation is the reality gap—the discrepancy between simulated and real-world environments. Domain adaptation techniques such as Randomized-to-Canonical Adaptation Networks (RCANs)~\cite{james2019sim} minimize this gap by mapping simulated observations $o_{\text{sim}}$ to a canonical space:

\begin{equation}
	o_{\text{real}} = f_{\theta}(o_{\text{sim}}),
\end{equation}

where $f_{\theta}$ is a learned mapping function. This enables \gls{reinforcement learning} policies trained in simulation to generalize effectively to real-world navigation tasks.

\section{Applications}

\subsection{Autonomy in Endovascular Navigation}

The field of autonomous endovascular navigation has advanced significantly due to the integration of robotics and medical imaging technologies. To evaluate a system's autonomy level, \textcite{haidegger2019autonomy} introduces the concept of a \gls{degree of autonomy} metric. This metric provides a structured framework to assess the autonomy levels of Medical Electrical Equipment (MEE) and Medical Electrical System (MES) in surgical settings, highlighting progressive capabilities for enabling autonomous endovascular interventions. The \gls{degree of autonomy} metric, as defined in ISO/IEC Technical Report IEC/TR 60601-4-1, employs a mathematical model to characterize autonomy through four core cognitive-related processes:

\[
	\operatorname{DoA} = F\{G, E, M, S\}
\]

where
\begin{itemize}
	\item \( G \) represents "Generate an option," entailing the formulation of possible actions or strategies based on real-time observations.
	\item \( E \) represents "Execute an option," which involves carrying out a chosen action or strategy, potentially with robotic assistance.
	\item \( M \) represents "Monitor an option," gathering information necessary for assessing the system's status, including internal and external sensory inputs.
	\item \( S \) represents "Select an option," where a decision is made on a specific course of action from the generated options.
\end{itemize}

The \gls{degree of autonomy} is computed as a normalized sum of these functions, each quantified on a linear scale from 0 (fully manual) to 1 (fully autonomous), thus:

\[
	\operatorname{DoA} \in [0, 1]
\]

In this scale, a \gls{degree of autonomy} value of 0 indicates a completely manual system, whereas a \gls{degree of autonomy} of 1 denotes full autonomy. Each functional component \( G, E, M, \) and \( S \) can be managed by either a human operator or a computerized system, thereby influencing the overall autonomy level. This thesis aims to develop a fully autonomous system for navigation in endovascular settings.

Autonomous endovascular navigation, driven by \gls{artificial intelligence}, represents a transformative shift in interventional procedures, with potential benefits including improved procedural precision, minimized radiation exposure, and reduced cognitive and physical strain on clinicians. These benefits are associated with enabling a \(\operatorname{DoA} = 1\) system. Recent research has explored various \gls{machine learning} approaches, including \gls{reinforcement learning} and \gls{learning from demonstration}, for navigating complex vascular structures using catheters and guidewires. This section reviews major methodologies in autonomous endovascular navigation, focusing on reinforcement learning, \gls{imitation learning}, and hybrid techniques that integrate expert-driven data with \gls{reinforcement learning}.

\subsubsection{Reinforcement Learning (RL)}

Reinforcement learning has become a leading approach in autonomous endovascular navigation, with approximately 64\% of studies utilizing \gls{reinforcement learning}-based methods~\autocite{robertshaw2023artificial}. \gls{reinforcement learning} enables systems to autonomously learn optimal navigation strategies by receiving environmental feedback based on actions taken, simulating human learning through iterative trial-and-error~\autocite{arulkumaran2017deep}. Several \gls{reinforcement learning} algorithms, including Deep Deterministic Policy Gradient (DDPG) and \gls{dueling deep Q-networks}, have been applied in endovascular settings. For instance, \glspl{dueling deep Q-networks} has been used to guide endovascular catheters, optimizing navigation paths based on \gls{electromagnetic} tracking sensor feedback~\autocite{behr2019deep}. However, \gls{reinforcement learning} often requires extensive in vitro validation before clinical application due to challenges posed by the dynamic and variable nature of human vasculature~\autocite{schwein2017flexible}.

\subsubsection{Learning from Demonstration (LfD)}

\gls{learning from demonstration}, also known as \gls{imitation learning}, allows models to learn by observing expert operators. In endovascular navigation, \gls{learning from demonstration} is frequently combined with \gls{reinforcement learning} to create a hybrid approach that uses expert-generated trajectories as a baseline, reducing the exploration time needed for effective \gls{reinforcement learning} training~\autocite{chi2020collaborative}. This combined \gls{learning from demonstration}-\gls{reinforcement learning} method has shown promise in improving navigation efficiency, as pre-acquired expert knowledge facilitates a smoother transition from simulation to physical trials~\autocite{cho2022sim}. Furthermore, this approach reduces mechanical wear on robotic systems, a critical advantage for the repeated cycles often required in \gls{reinforcement learning} training~\autocite{karstensen2022learning}.

\subsubsection{Hybrid Approaches: Combining RL and Imitation Learning}

Several studies have implemented a hybrid approach, merging \gls{reinforcement learning} and \gls{learning from demonstration} to leverage the advantages of both techniques. For example, \textcite{chi2020collaborative} used \gls{generative adversarial imitation learning} in conjunction with \gls{reinforcement learning} to enhance the adaptability of robotic catheter navigation across various vascular geometries~\autocite{chi2020collaborative}. This hybrid technique seeks to accelerate the training process, enhance navigation robustness, and ensure that systems generalize beyond specific vascular layouts encountered during initial training~\autocite{kweon2021deep}.

\subsubsection{Experimental Environments and Validation}

Experimental setups for evaluating autonomous navigation systems are primarily in vitro platforms, accounting for 79\% of studies, with a smaller subset employing in silico simulations. These controlled environments help mitigate risk before clinical deployment, though they do not fully capture the complexity of live human anatomy~\autocite{ionita2014challenges}. Studies frequently utilize \gls{electromagnetic}-tracking for real-time 3D visualization, aiding precise catheter orientation, though this technology would require innovation for clinical translation due to high costs and complexity~\autocite{schwein2017flexible}. Additionally, standardized performance metrics, including path length, procedural time, and task success rates, are essential for evaluating and comparing navigation models, though inconsistencies in metrics across studies present challenges for systematic assessment~\autocite{crinnion2022robotics}.

\subsection{Simulation Environments}

Simulation environments for endovascular interventions are categorized based on their fidelity and intended use, including synthetic, \gls{virtual reality}, animal, and cadaveric simulations. Each type offers distinct benefits and limitations depending on the desired outcomes in training, path planning, and system testing~\autocite{nesbitt2016role, dequidt2009towards, talbot2014interactive}. Synthetic environments, such as high-fidelity phantoms, are widely used for skill acquisition and procedural practice. These platforms are particularly advantageous for integrating imitation learning methods, allowing robotic systems to develop foundational skills by observing human techniques~\autocite{chi2020collaborative}. However, synthetic simulators can be costly and require physical maintenance, which poses limitations for continuous, scalable training~\autocite{molinero2019haptic}.

In the realm of \gls{virtual reality}-based simulations, advances have allowed for realistic modelling of endovascular procedures, providing a risk-free environment where trainees and autonomous agents can perform repetitive tasks to reinforce procedural learning and optimize path-planning strategies~\autocite{talbot2014interactive, wei2012near}. \gls{virtual reality} platforms often include haptic feedback systems to mimic tactile sensations, enhancing realism and user engagement~\autocite{molinero2019haptic}. Yet, despite these advances, many \gls{virtual reality} simulators are proprietary and lack open-source accessibility, limiting their integration with customizable training protocols and \gls{reinforcement learning} algorithms.

\subsubsection{Autonomous Catheter Navigation} Simulation environments have played a crucial role in progressing autonomous catheterization research. Initial studies in this field aimed at providing assistive navigation tools, but recent efforts are increasingly focused on higher levels of autonomy, such as semi-autonomous or fully autonomous navigation~\autocite{yang2017medical}. Autonomous catheterization has primarily leveraged \gls{deep reinforcement learning} techniques, which adapt to complex vascular geometries by using fluoroscopic images or other imaging modalities to train models on bi-dimensional synthetic phantoms~\autocite{behr2019deep, kweon2021deep}. However, alternative methods such as Dijkstra's algorithm and breadth-first search have also been applied for path planning in non-\gls{reinforcement learning}-based approaches, demonstrating diverse methodologies for addressing the challenges of autonomous navigation in simulated environments~\autocite{dijkstra1959note, fischer2022using}.

\subsubsection{Imitation Learning in Endovascular Simulation} \gls{imitation learning}, particularly when combined with \gls{reinforcement learning}, is pivotal in endovascular simulations for translating human expertise into robotic actions. It enables surgical agents to replicate complex manoeuvres based on human demonstrations, which is especially beneficial in scenarios requiring precise navigation through dynamic anatomical structures \autocite{ho2016generative}. This technique has been applied to various simulation environments, where agents trained on expert demonstrations were able to improve navigation accuracy and procedural efficiency \autocite{rafii2014hierarchical, chi2018trajectory}. Moreover, advanced models, including \gls{generative adversarial imitation learning}, have been implemented within synthetic and \gls{virtual reality} environments to refine agent decision-making and mimic expert strategies under simulated anatomical conditions \autocite{ho2016generative}. These methods demonstrate significant potential for achieving task autonomy in surgery, where a robotic agent, under supervised conditions, can assume part of the decision-making responsibilities \autocite{dupont2021decade}.

\subsection{Endovascular Tool Segmentation and Tracking}

Endovascular tool segmentation (\ie, guidewires and catheters) in fluoroscopic images is essential for precise navigation during minimally invasive procedures, enabling accurate tracking of the tool position and assisting in real-time decision-making. Techniques for guidewire segmentation have evolved from traditional image processing to advanced deep learning methods, allowing for improved detection accuracy and computational efficiency. This section reviews various approaches for guidewire segmentation, focusing on x-ray-based techniques, machine learning, and deep learning methods.

\subsubsection{Tool Visualization}

X-ray fluoroscopy is the standard imaging modality for visualizing guidewires and catheters during catheter-based interventions, such as \gls{percutaneous coronary intervention} and Transcatheter Aortic Valve Replacement (TAVR). These procedures require real-time visualization to guide the guidewire accurately through complex vascular structures. Traditional x-ray-based segmentation methods rely on the intensity and shape characteristics of the guidewire within the fluoroscopic images. Techniques such as thresholding, edge detection, and filtering are commonly applied to extract the guidewire’s linear structure from the background. For example, methods utilizing Difference-of-Gaussian (DoG) filters and Laplacian-of-Gaussian (LoG) filters have been effective in enhancing the guidewire’s features, enabling accurate differentiation from surrounding anatomical structures \cite{ma2010real, wu2013catheter}.

These intensity-based methods provide a relatively straightforward segmentation approach; however, they are sensitive to variations in image quality, such as noise and occlusions, which are common in fluoroscopic imaging. To mitigate these issues, techniques such as Kalman filtering have been integrated to track the guidewire’s position across frames, compensating for noise and ensuring smoother segmentation~\cite{wu2014fast}. Additionally, multi-view setups, which apply epipolar constraints across paired views, allow for 3D localization of guidewires, enhancing segmentation accuracy and enabling applications like 3D roadmapping~\cite{baur2016automatic}.

\subsubsection{ML Approaches in Endovascular Tool Segmentation}

The integration of machine learning has advanced guidewire segmentation by providing data-driven solutions that can generalize to variations in guidewire shape and appearance. Early machine learning methods applied techniques like dictionary learning, which learns a set of representative guidewire shapes from annotated data, allowing for robust segmentation by fitting these learned shapes to new image frames~\autocite{milletari2013automatic}. This approach enhances segmentation in cases where traditional filtering methods struggle, particularly when dealing with guidewires of different shapes or in varying anatomical locations.

Model-based approaches have also been applied, using predefined models of guidewire appearance to guide segmentation. For instance, B-spline curves are commonly used to model guidewire shapes as they allow for smooth, continuous curves that match the guidewire's natural form. These B-spline models can be initialized in the first frame and updated in subsequent frames to maintain alignment with the guidewire's current position, thus providing a balance between shape accuracy and computational efficiency \autocite{pauly2010machine}. However, model-based approaches still require initial manual labelling or accurate initialization, which can limit their usability in fully automated settings.

\subsubsection{Deep Learning in Endovascular Tool Segmentation}

Recent advances in deep learning have led to the development of highly accurate and automated segmentation frameworks for guidewires. \glspl{cnn} and \glspl{fully convolutional network} have been widely adopted for their ability to learn hierarchical features directly from raw image data, enabling robust guidewire segmentation even in challenging imaging conditions~\autocite{baur2016automatic}. For instance, \textcite{baur2016cathnets} demonstrated the effectiveness of \gls{fully convolutional network} for segmenting catheters and guidewires in fluoroscopic images, leveraging the network’s capability to detect tubular structures across varying levels of contrast and noise \autocite{baur2016cathnets}.

Furthermore, \gls{cnn}-based approaches have facilitated real-time guidewire tracking by using temporal data across sequential frames, providing continuous updates to the guidewire’s segmented path. These methods often outperform traditional algorithms in accuracy and computational speed, making them highly suitable for real-time surgical assistance. However, deep learning approaches are typically data-intensive, requiring large, annotated datasets to achieve high accuracy. In clinical environments where labelled data is limited, this can pose challenges for implementing these methods effectively \autocite{milletari2013automatic}.

\subsection{3D Reconstruction}

Accurate three-dimensional visualization of interventional devices is crucial for successful navigation through complex vascular anatomy, providing clinicians with spatial awareness that cannot be achieved through two-dimensional projections alone. During endovascular procedures, clinicians must navigate guidewires and catheters through tortuous vascular pathways while avoiding perforation or dissection of vessel walls. Traditional 2D fluoroscopy lacks depth perception and can obscure the true spatial relationships between devices and anatomy, potentially leading to navigation errors or procedural complications. The ability to reconstruct accurate 3D representations of guidewires in real-time thus addresses a fundamental clinical need, enabling more precise navigation, reducing procedure times, and improving patient safety.

Among the available techniques, bi-planar X-ray imaging has established itself as a gold standard for 3D reconstruction in clinical settings, offering simultaneous orthogonal views that enable precise triangulation of device positions and shapes. This technology is particularly valuable in complex interventions such as neurovascular procedures, structural heart interventions, and aortic repairs, where millimeter-level accuracy can significantly impact procedural outcomes. Despite its advantages, bi-planar imaging faces challenges including equipment cost, radiation exposure, and workflow integration, prompting the exploration of alternative approaches that attempt to balance reconstruction accuracy with clinical practicality. The evolution of 3D reconstruction methods thus reflects a continuous trade-off between precision, efficiency, and clinical feasibility, with each technique offering distinct advantages and limitations in the context of endovascular interventions.

Advances in guidewire reconstruction and localization have significantly enhanced precision and navigation during minimally invasive interventions. The focus of recent work has been on improving 3D localization accuracy and reducing the clinical workflow disruptions associated with traditional fluoroscopic methods. Various reconstruction techniques have emerged, notably those relying on multi-view fluoroscopy, monoplane approaches supplemented by kinematic data, and, more recently, machine learning-based methods.

\subsubsection{Biplane Fluoroscopic Approaches}

Biplane fluoroscopy has been foundational for accurate guidewire shape estimation, utilizing dual-angle imaging for real-time triangulation of 3D structures. Early works, such as \textcite{burgner2011toward}, employed projective-invariant triangulation methods to derive the guidewire’s centreline, achieving accurate reconstructions but often requiring extensive calibration and image processing. This approach has since been refined by \textcite{wagner20164d} and \textcite{hoffmann2015electrophysiology,hoffmann2012semi,hoffmann2013reconstruction}, who incorporated dynamic 4D modelling to capture continuous guidewire movement, yielding enhanced shape reconstruction. These methods have proved effective in experimental setups; however, the high cost, radiation exposure, and operational constraints of biplane systems during interventions limit their clinical applicability. The dependence on dedicated biplane C-arm configurations also complicates their integration into diverse clinical environments, leading to increased interest in monoplane alternatives.

\subsubsection{Monoplane Fluoroscopic Techniques}

To mitigate the limitations of biplane fluoroscopy, monoplane C-arm systems have been adapted for guidewire shape reconstruction through the integration of supplementary data sources. One approach, introduced by \textcite{lobaton2013continuous}, involved combining fluoroscopic images with kinematic models in bronchoscopy, which provided deformable surface parameterization to improve positional accuracy. This method, while precise, was constrained to offline simulation environments and required meticulous C-arm adjustments. \textcite{vandini20133d} expanded this monoplane approach by reconstructing robotic catheter shapes through 2D centreline extraction and appearance priors, albeit with workflow disruptions due to frequent C-arm repositioning. To address this, \textcite{vandini2015vision} later proposed a kinematic model-integrated monoplane method for transnasal robotic surgery, eliminating the need for C-arm adjustments and allowing more seamless workflow integration.

Further advancements have been made by \textcite{otake2014piecewise} and Papalazarou~\textcite{papalazarou20123d}, who incorporated piecewise-rigid 2D/3D registration and non-rigid structure-from-motion techniques. These methods demonstrated improved intraoperative adaptability, yet multiple monoplane views were often required to achieve accurate 3D reconstructions, increasing computational demands. Although monoplane fluoroscopy reduces radiation exposure compared to biplane setups, it still presents workflow challenges in clinical environments, underscoring the need for real-time adaptable systems.

\subsubsection{Deep Learning and Data-Driven Reconstruction Techniques}

Emerging deep learning methods have shown potential to address limitations of traditional fluoroscopic techniques by interpreting monoplane images with high precision. While these models offer promising avenues for reducing reliance on multi-view systems, their application to full 3D guidewire reconstruction remains largely unexplored. As highlighted by \textcite{ramadani2022survey}, current deep learning efforts primarily target segmentation or localization tasks, with true 3D reconstruction still in its infancy. Moreover, the need for large annotated datasets and high computational demands poses additional challenges for clinical deployment.

\onehalfspacing
\chapter{CathSim}
\chaptermark{CathSim}
\glsresetall 

\begin{cabstract}
	Autonomous robots in endovascular operations have the potential to navigate circulatory systems safely and reliably while decreasing susceptibility to human errors. However, numerous challenges are involved in the training process of such robots, including prolonged training times and safety concerns related to catheter-aorta interactions. Recently, endovascular simulators have been utilized in medical training, but they are typically unsuitable for autonomous catheterization due to a lack of standardization and integration with reinforcement learning frameworks. Furthermore, most current simulators are closed-source, which hinders the collaborative development of safe and reliable autonomous systems through shared learning and community-driven enhancements. To address these limitations, \textit{CathSim} is introduced as an open-source simulation environment designed to accelerate the development of machine learning algorithms for autonomous endovascular navigation. A high-fidelity catheter and aorta are simulated within a state-of-the-art endovascular robot. Real-time force sensing between the catheter and the aorta is provided within the simulation. The simulator is validated through two distinct catheterization tasks --- navigation towards the Left Common Carotid Artery (LCCA) and the Brachiocephalic Artery (BCA) --- using two widely adopted reinforcement learning algorithms: Soft Actor Critic (SAC) and Proximal Policy Optimization (PPO). The experimental results demonstrate that the open-source simulator effectively mimics the behaviour of real-world endovascular robots and facilitates the development of various autonomous catheterization tasks.
\end{cabstract}

This chapter presents the work from the following publication:

\fullcite{jianu2024cathsim}

\section{Introduction}\label{ch3sec:cathsim-introduction}

Endovascular intervention has significantly evolved from traditional open surgery. It is performed through a small incision, typically in the common femoral or radial artery, allowing surgical tools such as catheters and guidewires to be manoeuvred within the vasculature. This form of \gls{minimally invasive surgery} offers numerous benefits, including reduced blood loss, shorter recovery time, less postoperative pain, and a diminished inflammatory response compared to traditional methods~\autocite{wamala2017use}. In clinical practice, navigation of the catheter and guidewire to the diagnostic site is guided by fluoroscopy, a real-time X-ray imaging technique. However, endovascular intervention is not without limitations, such as the absence of sensory feedback, radiation exposure, operator dependency, a steep learning curve, and the need for precise, dexterous manipulation~\autocite{omisore2020review}.

To mitigate the continuous radiation risk to surgeons during fluoroscopic procedures, numerous robotic systems with leader-follower (master-slave) teleoperation architectures have been developed~\autocite{kundrat2021mr, pereira2020first, burgner2015continuum}. In these systems, a leader device is controlled by the surgeon, with commands mapped to a follower robot that executes the corresponding actions. This setup enables remote operation from a safe, radiation-free zone. Recent advancements have focused on the development of Magnetic Resonance (MR)-safe robotic platforms~\autocite{kundrat2021mr}, eliminating ionizing radiation exposure while facilitating soft tissue visualization, including vasculature~\autocite{heidt2019real, abdelaziz2019toward}. In academic research, these robotic systems have been further enhanced with assistive features such as force/torque sensing~\autocite{konstantinova2014implementation}, haptic feedback~\autocite{molinero2019haptic}, and real-time segmentation and tracking~\autocite{nguyen2020end}.

\begin{figure}[t]
	\centering
	\includegraphics[width=.7\linewidth]{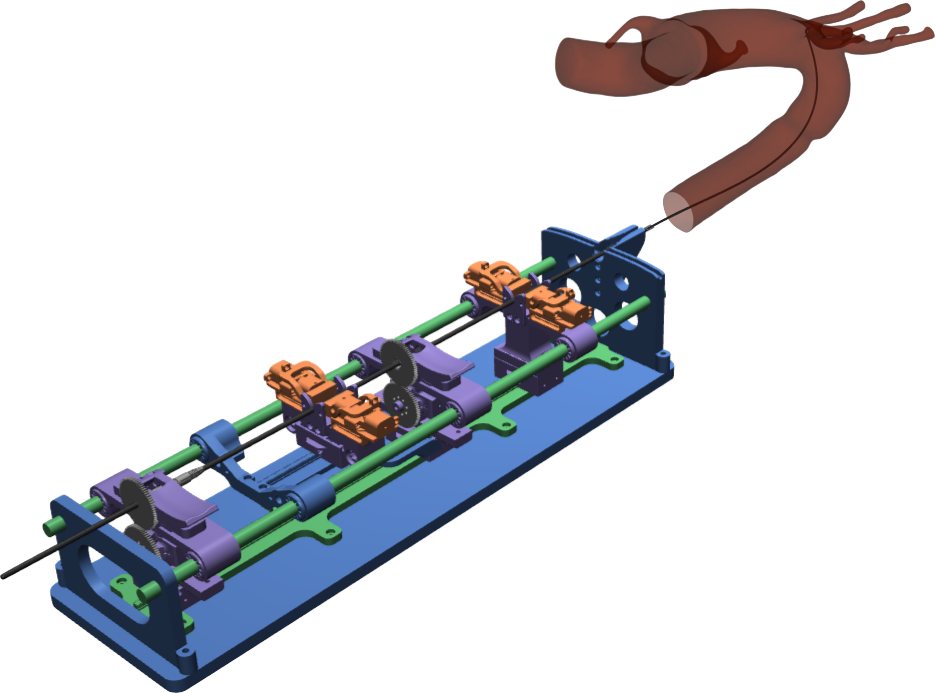}
	\mycaption{CathSim Overview}{An open-sourced simulator for autonomous cannulation. The figure shows the follower robot navigating the catheter through an aortic arch. \copyright 2024 IEEE}
	\label{ch3fig:cathsim_overview}
\end{figure}

Recent robotic platforms for endovascular intervention have demonstrated assistive potential in supporting procedural success. However, two key limitations persist: \textit{i)} the lack of autonomy~\autocite{attanasio2021autonomy}, and \textit{ii)} increased procedural duration compared to manual techniques~\autocite{chi2020collaborative}. In current systems, the surgeon must navigate the three-dimensional vascular space using only two-dimensional fluoroscopic imaging and without the aid of haptic feedback. This setup imposes high cognitive and physical demands, requiring exceptional precision to avoid vascular damage under mentally strenuous conditions.

Procedural automation could alleviate these burdens by improving efficiency and reducing the risk of complications. Yet, achieving safe autonomy remains non-trivial—it necessitates the integration of sophisticated vision, learning, and control systems capable of meeting stringent safety requirements~\autocite{attanasio2021autonomy}.

Endovascular navigation simulations have historically faced significant challenges, limiting their advancement and practical utility~\autocite{simaan2018medical}. One major issue has been the absence of standardized environments. This issue is exemplified by the closed-source nature of many previous simulators, which, despite showcasing algorithmic innovations, are not publicly accessible, thereby hindering reproducibility and broader community contributions. Computational performance, crucial for real-time response and complex algorithmic processing, is often insufficient in existing platforms. Furthermore, integration with frameworks such as the Gymnasium interface~\autocite{towers2024gymnasium}, which is essential for the streamlined development of \gls{reinforcement learning}-based methods, is frequently absent. Additional barriers, including installation complexity and limited interactivity, further impede widespread adoption and iterative development within research contexts. Addressing these challenges is essential to advancing endovascular navigation simulations and supporting training and algorithm development. Related work in Section~\ref{ch3sec:cathsim-introduction} further explores these limitations and introduces \textit{CathSim}, an open-source, efficient, and high-fidelity simulation environment that responds directly to these needs.

In response to these challenges, a new \gls{minimally invasive surgery} simulation environment tailored for endovascular procedures is introduced. The primary objective is to facilitate the development of endovascular navigation autonomy by providing a standardized environment familiar to the \gls{machine learning} community. To this end, \textit{CathSim} is presented as a real-time simulation platform for autonomous cannulation, developed using the MuJoCo physics engine~\autocite{mujoco}. MuJoCo was selected for its real-time performance, accurate physics modelling, and suitability for optimal control applications, making it well-suited to the intended objectives. An overview of the simulator is provided in Fig.~\ref{ch3fig:cathsim_overview}. The key contributions and potential applications of the simulator are summarized as follows:

\begin{enumerate}
	\item \emph{CathSim} is introduced as a novel open-source simulation environment for endovascular procedures.
	\item Baselines and benchmarks are provided for autonomous cannulation tasks in \emph{CathSim} using two widely adopted \gls{reinforcement learning} algorithms.
\end{enumerate}

\section{Related Work}

\begin{table}[h]
	\caption{Endovascular Simulator Environments Comparison}\label{ch3tab:endovascular_simulators_comparison}
	\centering
	\begin{threeparttable}

		\begin{tabular}{l l l l l c}
			\toprule
			\thead{Simulator}                   & \thead{Physics Engine}                    & \thead{Catheter}     & \thead{Guidewire}    & \thead{Aorta} & \thead{OS?}          \\
			\midrule
			\textcite{molinero2019haptic}       & Unity Physics \autocite{juliani2018unity} & Discretized          & \makecell[c]{\xmark} & 3D artery     & \xmark               \\
			\textcite{karstensen2020autonomous} & SOFA \autocite{faure:hal-00681539}        & Timoshenko~\tnote{1} & \makecell[c]{\xmark} & 2D vessel     & \checkmark           \\
			\textcite{behr2019deep}             & SOFA \autocite{faure:hal-00681539}        & Timoshenko~\tnote{1} & \makecell[c]{\xmark} & 2D vessel     & \checkmark           \\
			\textcite{omisore2021novel}         & CopelliaSim \autocite{rohmer2013v}        & \makecell[c]{\xmark} & Discretized          & 3D vessel     & \checkmark~\tnote{2} \\
			\textcite{schegg2022automated}      & SOFA \autocite{faure:hal-00681539}        & \makecell[c]{\xmark} & Timoshenko~\tnote{1} & 3D artery     & \checkmark           \\
			\midrule
			\textbf{CathSim} (ours)             & MuJoCo~\cite{mujoco}                      & Discretized          & \makecell[c]{\xmark} & 3D artery     & \checkmark           \\
			\bottomrule
		\end{tabular}
		\begin{tablenotes}
			\footnotesize
			\item[1] Timoshenko Beam theory~\cite{davis1972timoshenko}
			\item[2] Primary engine is open source; however, certain functionalities require commercial licenses.
		\end{tablenotes}
	\end{threeparttable}
\end{table}

\paragraph{Simulation Environments.} Endovascular surgical simulation environments can be categorized into four primary types: synthetic models, animal models, virtual systems, and human cadavers, each offering distinct advantages and inherent limitations~\autocite{nesbitt2016role, wei2012near, dequidt2009towards, talbot2014interactive}. These environments are primarily designed for skill development~\autocite{sinceri2015basic, nesbitt2016role} or for the integration and development of assistive features, such as haptic feedback~\autocite{molinero2019haptic}. However, reliance on physical materials often renders them unsuitable for \gls{reinforcement learning}-based applications. Recent advancements have included high-fidelity synthetic phantoms for \gls{imitation learning}, as exemplified by the SOFA simulation environment~\autocite{faure:hal-00681539}, which has been tested on two-dimensional synthetic phantoms~\autocite{lillicrap2015continuous}. Despite these developments, most environments remain closed-source, thereby limiting accessibility, hindering reproducibility, and restricting broader contributions to the field.

\paragraph{Autonomous Catheterization.} Machine learning has enabled the shift from assistive features to semi-autonomous navigation in the context of autonomous catheterization~\autocite{yang2017medical}. In recent work, a focus has been placed on \gls{deep reinforcement learning} due to its effectiveness in complex decision-making tasks, as demonstrated in domains such as autonomous driving. Numerous studies have utilized \gls{reinforcement learning}, particularly with image data derived from fluoroscopy~\autocite{karstensen2020autonomous,kweon2021deep,molinero2019haptic,behr2019deep,omisore2021novel}. While alternative methods, such as the Dijkstra algorithm or breadth-first search, have been employed~\autocite{cho2021image,schegg2022automated}, model-free \gls{reinforcement learning} has been regarded as well-suited for handling the uncertainty and complexity associated with achieving higher autonomy. Nonetheless, most research remains in the early stages of autonomy~\autocite{yang2017medical}, and comprehensive autonomous navigation of the vascular system continues to represent a challenging yet promising objective for \gls{reinforcement learning}.

In Table~\ref{ch3tab:endovascular_simulators_comparison}, a detailed comparison is provided of current learning-based works that utilize environments built on a variety of physics engines. A limitation of such environments is their lack of public availability, which hinders reproducibility. \textit{CathSim}, unlike other simulators, offers an open-source platform specifically designed for training autonomous agents, allowing for rapid development of various \gls{machine learning} methodologies. Based on the MuJoCo framework~\autocite{mujoco}, the simulator provides an advanced simulation environment suitable for real-time applications. In addition, real-time force sensing capabilities and high-fidelity, realistic visualization of the aorta, catheter, and endovascular robots are supported. In practice, \emph{CathSim} can be used to train \gls{reinforcement learning} agents or serve as a practice platform for healthcare professionals.

Recent advancements in gym-based simulation frameworks have incorporated the SOFA platform, as demonstrated by SofaGym, a general framework for SOFA environments~\autocite{schegg2023sofagym}, and LapGym, designed specifically for laparoscopic surgery~\autocite{scheikl2023lapgym}. Both frameworks utilize the Finite Element Method (FEM) to simulate deformable objects and integrate a Gym API~\autocite{gymnasium2023}, representing significant progress in medical simulation. However, support for domain randomization—a critical component for training robust \gls{machine learning} and \gls{reinforcement learning} models—is absent in these systems. In contrast, \textit{CathSim} is focused specifically on endovascular navigation, leveraging the MuJoCo physics engine~\autocite{mujoco} and incorporating domain randomization. Additionally, highly realistic aortic models derived from post-mortem anatomies and Computerized Tomography (CT) scans are employed, offering a unique contribution to the medical simulation landscape. Unlike SofaGym’s broader application as a Gym wrapper for diverse SOFA environments and LapGym’s emphasis on laparoscopic surgery, \textit{CathSim} provides targeted advancements in endovascular navigation simulation, enriching the field with application-specific innovations.

\section{The CathSim Simulator}

The \emph{CathSim} environment is composed of three components:
\textit{i)} the follower robotic model for endovascular procedures, as introduced by \textcite{abdelaziz2019toward},
\textit{ii)} the aortic arch phantoms, and
\textit{iii)} the catheter or guidewire.
Real-time simulation is enabled, and support is provided for the training of state-of-the-art learning algorithms.

\subsection{Robot Simulation}

\begin{figure}[ht]
	\centering
	\includegraphics[width=.7\linewidth]{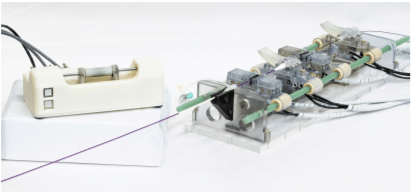}
	\mycaption{CathBot}{The robotic platform for fluoroscopy and MR-guided endovascular interventions consists of leader device \textit{(left)} and follower robot \textit{(right)}. Adapted from~\cite{nguyen2020end}.}
	\label{ch3fig:cathbot_overview}
\end{figure}

\begin{figure}[ht]
	\centering
	\includegraphics[width=.7\linewidth]{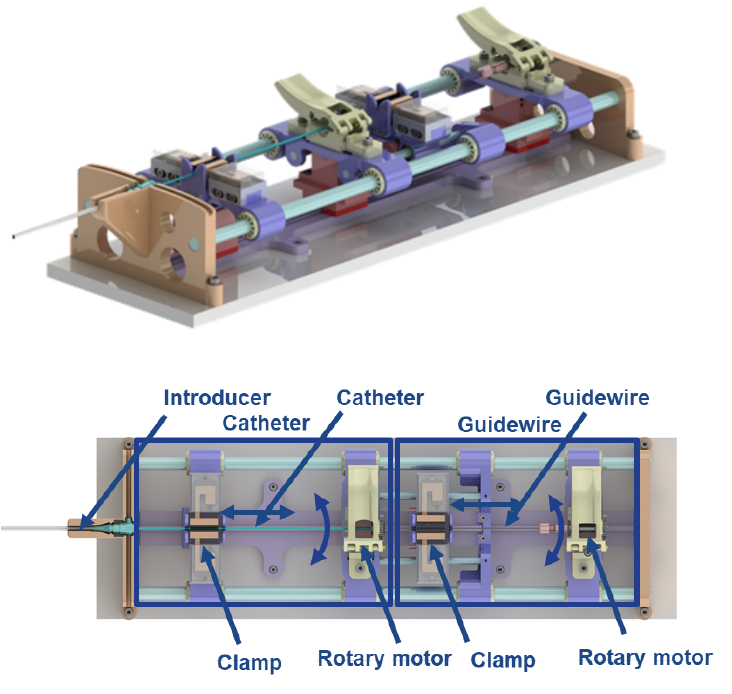}
	\mycaption{CathBot Follower}{The follower robot, which uses a linear leader-follower architecture with pneumatic actuators for translation and rotation, and J-clamps for friction-based motion transfer~\cite{kundrat2021mr}.}
	\label{ch3fig:cathbot_follower}
\end{figure}

In this work, the transfer of CathBot, developed by \textcite{kundrat2021mr}, to simulation is pursued for the purpose of autonomous agent training. CathBot is selected due to its status as a state-of-the-art robot that is not restricted by a commercial licence. The design of CathBot follows the widely adopted leader-follower architecture, in which the leader robot uses haptic feedback generated by the catheter's interaction with the environment through the navigation system~\autocite{molinero2019haptic}, maintaining intuitive control that replicates human motion patterns such as insertion, retraction, and rotation. The follower robot mimics the leader’s motion and is composed of two pneumatic linear motors for translation, one pneumatic rotating stepper motor, and two pneumatic J-clamps for clamping the instrument during translational movements.

Given the linear mapping between the leader and the follower robot in CathBot’s design~\autocite{kundrat2021mr}, the simulation focuses solely on the follower robot for simplicity. The follower robot is simulated by constructing four modular platforms attached to a main rail. On two of these platforms, a pair of clamps is positioned to secure the guidewire during translational motion, while the remaining two platforms are configured with rotary catheter and guidewire components to perform angular motion. The components responsible for translational movement along the main rail, as well as the clamps, are joined using prismatic joints. Revolute joints are used to attach the wheels, thereby enabling rotational motion of the catheter. The rotational behaviour of the catheter involves frictional interaction, in which the clamps lock the catheter and rotate it during actuation. A similar mechanism is applied for linear displacement. However, due to the difficulty of accurately modelling friction-dependent rotation in simulation, a perfect motion assumption is made, and joint actuation is performed directly.

The follower component of the CathBot robot is simulated alongside the aortic arch phantoms and the catheter. These elements are modelled using the MuJoCo physics engine~\autocite{mujoco}, selected for its stability, computational efficiency, and ability to support real-time interactions.

\subsection{Aorta Simulation}

\begin{figure}[ht]
	\centering
	\subfloat{\includegraphics[width=0.39\linewidth]{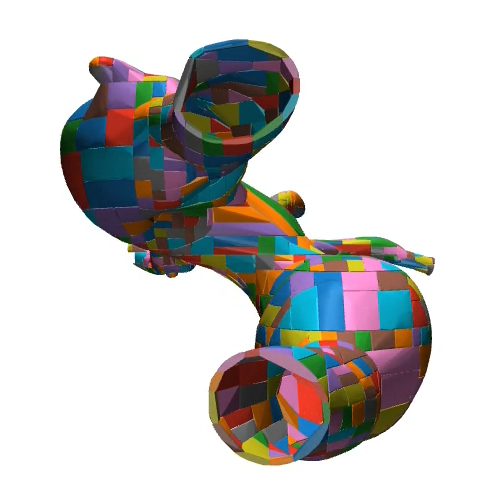}}
	\subfloat{\includegraphics[width=0.59\linewidth]{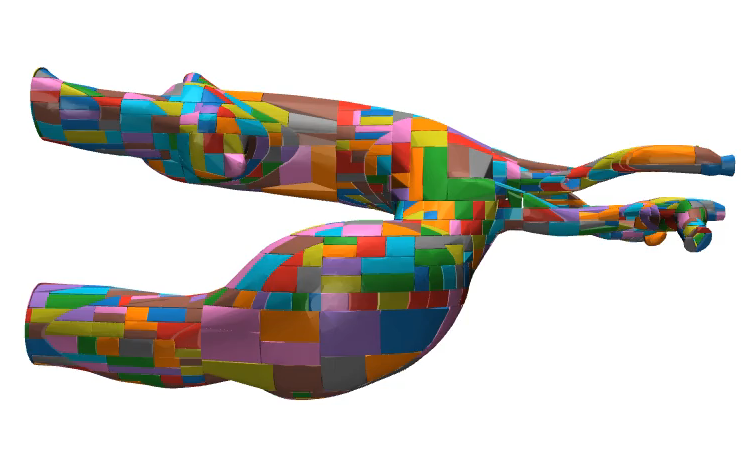}}
	\vspace{1ex}
	\mycaption{Aortic Collision Properties}{The collision property of the aorta is inferred through the decomposition of the aortic arch into a series of convex hulls, whilst the visual properties are given by scanned high fidelity silicone-based, transparent, anthropomorphic phantoms. \copyright 2024 IEEE}
	\label{ch3fig:aorta_convex_decompesition}
\end{figure}

The aortic simulation is generated by scanning silicone-based, transparent, anthropomorphic phantoms of Type-I and Type-II aortic arch models (Elastrat Sarl, Switzerland) to produce high-fidelity 3D mesh models. These models are derived from postmortem vascular casts using methodologies described in~\autocite{gailloud1997vitro,martin1998vitro,gailloud1999vitro}, ensuring anatomical accuracy. The concave mesh is subsequently decomposed into a set of nearly convex surfaces using volumetric hierarchical approximate decomposition~\autocite{silcowitz2010interactive}, resulting in $1,024$ convex hulls for the Type-I aortic arch and $470$ convex hulls for the Type-II aortic arch. The variation in the number of convex hulls corresponds to differences in the concavity of the two meshes~\autocite{mamou2009simple}. Convex hulls are employed for collision modelling, optimizing computational efficiency and enabling the use of soft contacts within the physics engine~\autocite{mujoco}. Figure~\ref{ch3fig:aorta_convex_decompesition} illustrates the aorta simulation.

\subsubsection{Additional Aortic Models in CathSim}

\begin{figure}[ht]
	\centering
	\subfloat[Type-I (Aneurysm)]{\includegraphics[width=0.3\textwidth]{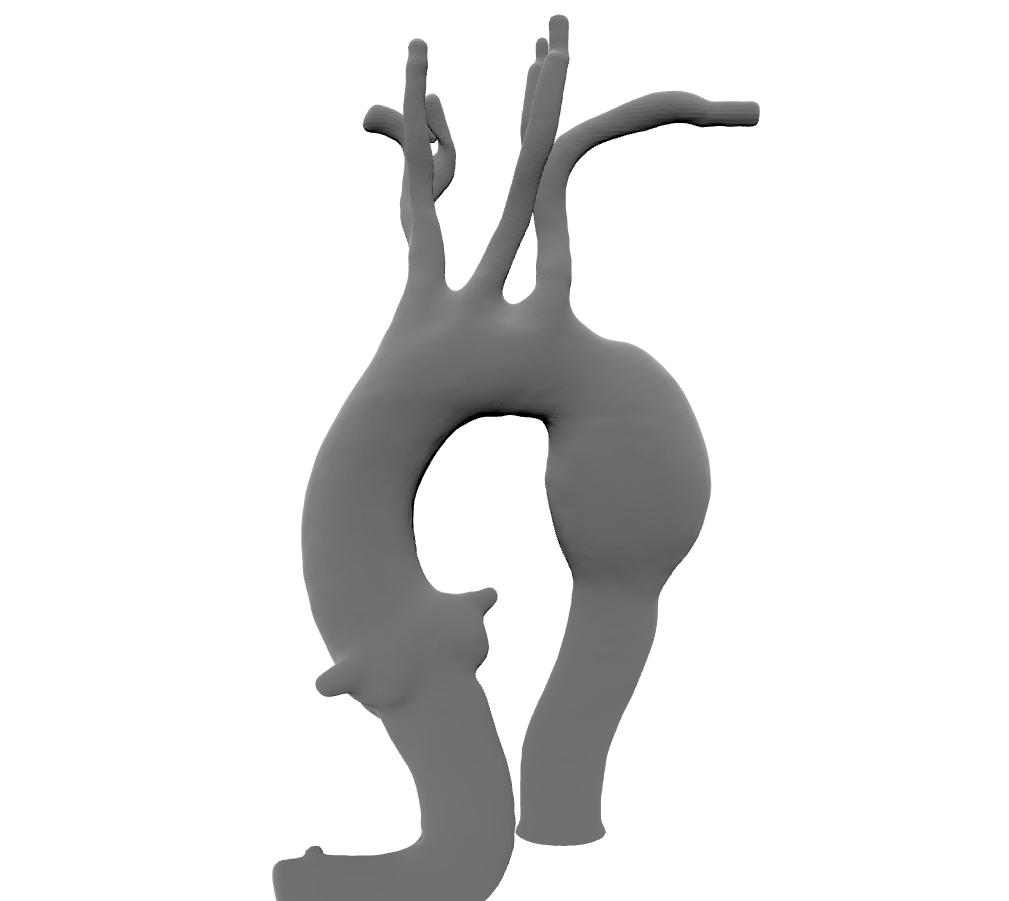}}
	\hfill
	\subfloat[Type-II]{\includegraphics[width=0.3\textwidth]{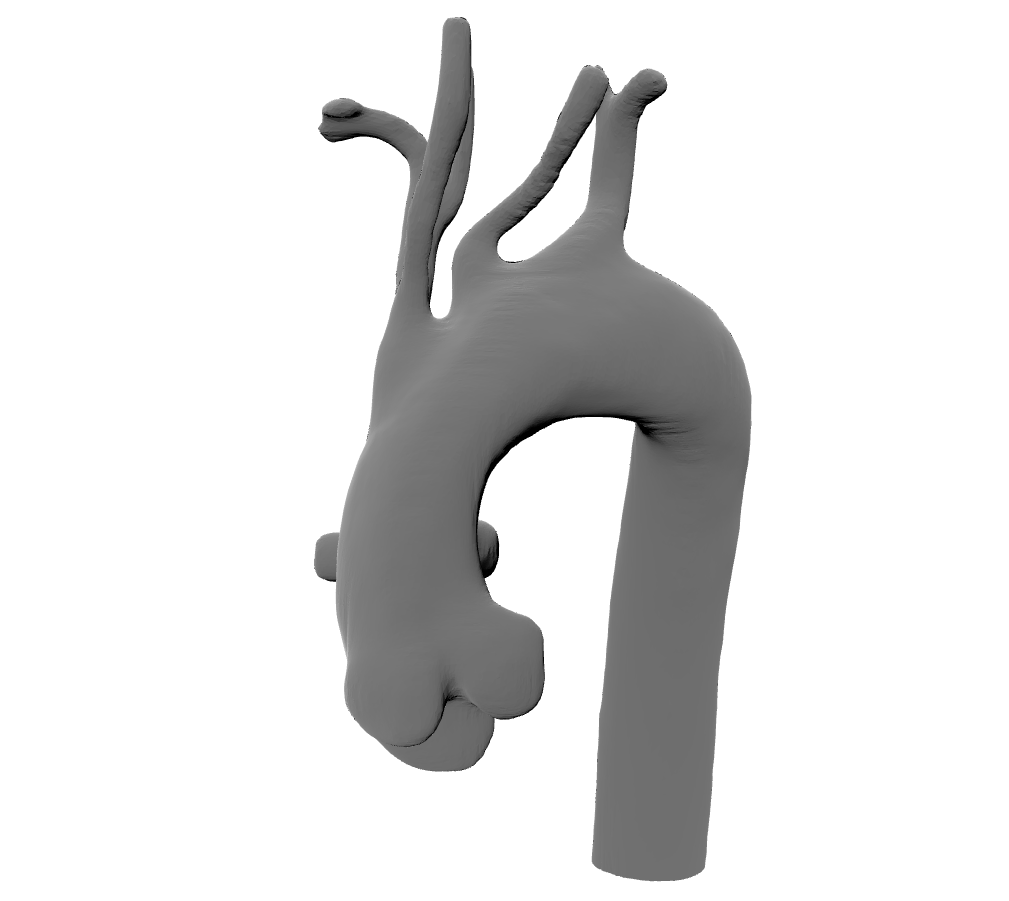}}
	\hfill
	\subfloat[Low Tortuosity]{\includegraphics[width=0.3\textwidth]{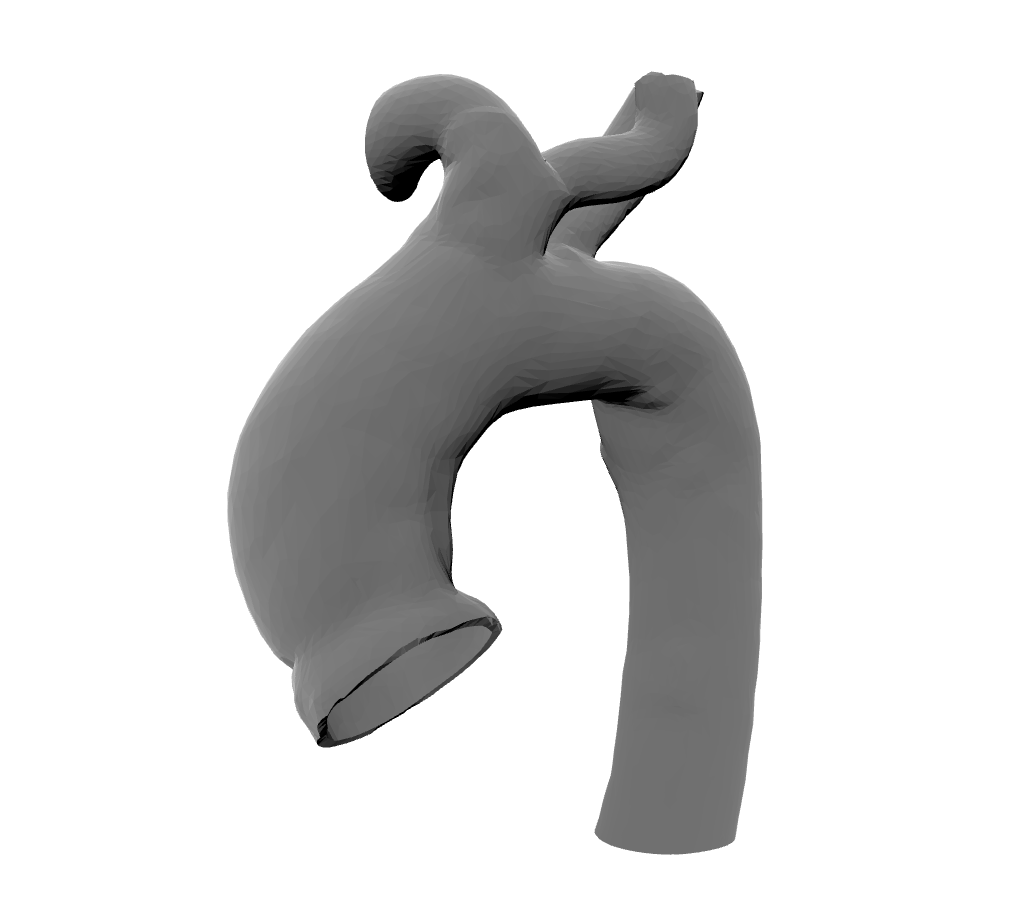}}
	\caption{Aortic Models.}
	\label{fig:three aorta}
\end{figure}

In addition to Type-I Aortic Arch model which is mainly used in our experiments, we incorporate three distinct aortic models to enrich our anatomical dataset. These models include a high-fidelity Type-II aortic arch and a Type-I aortic arch with an aneurysm, both sourced from Elastrat, Switzerland. Furthermore, a low tortuosity aorta model, based on a patient-specific CT scan, is included. With these three additional representations, our simulator contains four distinct aorta models. These models aim to enhance the diversity and accuracy of aortic structures available for research and educational endeavours. These aortas are visualized in Fig.~\ref{fig:three aorta}.

\subsection{Catheter Simulation}

In the context of catheter design, curvature is considered critical for manoeuvrability and precise placement within complex anatomical paths. This curvature is modelled using quaternion differences to represent the relative orientations of catheter segments. For segments with orientations \( \mathbf{q}_i \) and \( \mathbf{q}_{i+1} \), the curvature \( \boldsymbol{\kappa} \) is defined as:

\begin{equation}
	\boldsymbol{\kappa} = \mathbf{q}_i^{-1} \mathbf{q}_{i+1}
\end{equation}

Within this model, bending is prioritized over twisting. The bending energy \( E_{\text{bend}} \) is computed from the bending stiffness \( E \) and the deviation from the initial curvature \( \boldsymbol{\omega}_0 \):

\begin{equation}
	E_{\text{bend}} = \frac{1}{2} E \left( \boldsymbol{\kappa} - \boldsymbol{\omega}_0 \right)^2
\end{equation}

Accordingly, the total energy \( E_{\text{total}} \) of the catheter is expressed as:

\begin{equation}
	E_{\text{total}} = E_{\text{bend}}
\end{equation}

\begin{figure}[ht]
	\centering
	\includegraphics[width=.8\linewidth]{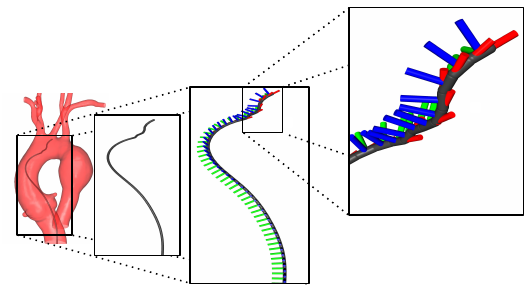}
	\mycaption{Cathether Properties}{The catheter's continuous properties are approximated by a series of rigid bodies linked through revolute joints. The kinematic chain can be visualized through the frames of each individual body, where the last ten links are actuated by internal micro-motors. \copyright 2024 IEEE}\label{ch3fig:guidewire-representation}
\end{figure}

This formulation emphasizes the significance of bending behaviour in determining catheter effectiveness. The behaviour is influenced by the material properties, represented by \( E \) (Young's modulus~\autocite{jastrzebski1976nature}, which defines the stiffness of the material), and the initial configuration \( \boldsymbol{\omega}_0 \). In this context, Young's modulus \( E \) denotes the ratio of stress to strain, indicating the extent of deformation experienced by the catheter under a specific load. The catheter itself can be visualized in Fig.~\ref{ch3fig:guidewire-representation}

\subsection{Contact Simulation}

In the \emph{CathSim} simulation, modelled using MuJoCo~\autocite{mujoco}, the interaction between the aorta and catheter is represented through point contacts within a rigid body framework. Each contact point is defined by a spatial frame in a global coordinate system, where the primary axis is aligned with the contact normal—crucial for calculating normal forces—while the remaining axes define the tangent plane for frictional force computations. The contact distance, which determines whether objects are separated, in contact, or penetrating, influences the resulting contact force calculations.

Within this rigid body model, dynamics are governed by the equation:

\begin{equation}
	M\dot{v} + c = \tau + J^T f ,
\end{equation}
where \( M \) denotes the joint-space inertia, \( \dot{v} \) the acceleration, and \( c \) the bias force, which is computed using the Recursive-Newton-Euler algorithm~\autocite{featherstone2014rigid}. The applied force \( \tau \), which includes components such as fluid dynamics and actuation forces, and the constraint Jacobian \( J \), which relates joint and constraint coordinates, are used to determine force interactions at the contact points. The rigid body assumption streamlines the simulation by focusing on fundamental mechanical principles. Deformations and complex material behaviours are not considered, which ensures both computational efficiency and simulation fidelity~\autocite{todorov2011convex, dreyfus2022simulation}.

\begin{figure}[ht]
	\centering
	\includegraphics[width=.7\linewidth]{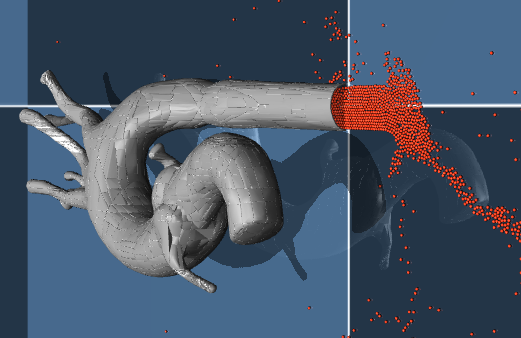}
	\caption[Blood Flow Simulation]{Blood Flow Simulation}
	\label{ch3fig:blood_flow}
\end{figure}

\subsection{Blood Simulation}

Although blood modelling is not the primary focus of the current work, a basic implementation is included for reference purposes. In this model, blood is treated as an incompressible Newtonian fluid, following the real-time methodology described in the study by \textcite{wei2012near} (see Fig.~\ref{ch3fig:blood_flow}). The dynamics of a pulsating vessel are intentionally omitted, resulting in the simplification of assuming rigid vessel walls. This approach is commonly adopted in the field, as demonstrated in studies such as \textcite{yi2018_xray_synthetic,behr2019deep,karstensen2020autonomous}, and serves to minimize computational demands while effectively simulating the forces that oppose guidewire manipulation.

\section{RL for Autonomous Cannulation}

The problem of autonomous cannulation is addressed by formulating it as an episodic \gls{partially observable markov decision process}. The agent, represented by the catheter, interacts with a continuous, partially observable environment \( \mathcal{E} \), modelled on the anatomical structure of an aortic arch. At each time step \( t \), an observation \( s_t \) is received that reflects the current state of the environment, and a continuous action vector \( a_t(s_t) \), corresponding to motor actuation, is selected. A reward \( r_t(s_t, a_t) \) is then obtained based on the agent’s progress toward the goal, followed by a transition to a new state \( s_{t+1} \).

The objective is to optimize a policy \( \pi(a|s) \) to guide the catheter through the aortic environment and reach the target position \( g \in \mathcal{G} \). Each episode terminates either upon reaching the desired position or when predefined termination conditions, such as time limits, are met. This formulation enables adaptability to anatomical variability and accommodates uncertainties inherent in real-world interactions, such as partial observability and noisy sensor feedback.

Subsequent sections describe the components of this framework, including the observation modalities (\ie, Internal, Image, and Sequential), the continuous action space, and the reward structure designed to promote efficient and accurate navigation of the catheter.

\subsection{Observations}

Three observation setups are evaluated: Internal, Image, and Sequential, each designed to provide varying levels of information to the \gls{reinforcement learning} agent.

The Internal setup delivers detailed proprioceptive data regarding the catheter's state, including \textit{position, velocity, centre of mass inertia, centre of mass velocity, and forces generated by both actuators and external interactions}. This rich dataset enables precise decision-making but assumes a complete internal understanding of the system.

The Image setup replicates clinical conditions by utilizing a virtual RGB camera placed above the aortic phantom. The resulting grayscale images, with a resolution of \(128 \times 128\), are analogous to X-ray images commonly used during catheterization procedures. Although higher-resolution images were considered, no performance gains were observed in experiments, indicating that the computational and memory costs outweigh any marginal benefits.

The Sequential setup builds upon the Image modality by incorporating a temporal dimension. It combines three successive images \( \{s_{t-2}, s_{t-1}, s_t\} \) to capture motion and temporal patterns in the catheter’s behaviour. This approach improves the agent’s ability to infer dynamic information, thereby enhancing its capacity to navigate complex environments effectively.

These observation modalities were selected to explore the trade-offs between information richness, temporal awareness, and computational efficiency.

\subsection{Actions and Rewards}

The agent’s actions are represented by a continuous vector \( a_t \in \mathbb{R}^{21} \). Of these, the first 20 elements correspond to the motors that actuate the revolute joints within the catheter tip, while the final element controls the prismatic joint responsible for translational movement. To ensure consistency and prevent extreme outputs, the action values are normalized within the range \( [-1, 1] \), resulting in an action space of \( a_t \in [-1, 1]^{21} \).

To address the sparsity of rewards inherent in the navigation task, reward shaping is applied to provide the agent with additional spatial guidance. Specifically, the reward function incorporates the negative Euclidean distance \(- d(h, g) = \| h - g \| \), where \( h \) denotes the position of the catheter tip and \( g \) represents the goal location. This formulation incentivizes the agent to minimize the distance to the target during its trajectory.

If the catheter tip reaches a goal region within a distance threshold \( \delta = 8 \) mm of the target, the episode is terminated and a terminal reward of \( r = 10 \) is provided. This threshold ensures full insertion of the catheter into the artery at the desired location. The reward function is defined as follows:

\begin{equation}
	r(h_t, g) =
	\begin{cases}
		10       & \text{if } d(h, g) \leq \delta \\
		-d(h, g) & \text{otherwise}
	\end{cases}
\end{equation}

Although this reward structure facilitates convergence by providing continuous feedback, it remains susceptible to local minima. For instance, the catheter may erroneously enter an unintended artery, thereby decreasing the distance to an incorrect but nearby target. In such cases, the agent must increase its distance from the incorrect location before reorienting toward the correct goal, thereby adding complexity to the optimization process.

\subsection{Network Architectures}

To handle the continuous action representation, two state-of-the-art \gls{reinforcement learning} algorithms are employed: \gls{ppo}~\autocite{schulman2017proximal} and \gls{sac}~\autocite{haarnoja2018soft}. The algorithm parameters are adopted from~\textcite{dalal2018safe}.

The network architecture varies depending on the observation type. For the Internal observation, a \gls{mlp}-based policy network is utilized, while for the Image and Sequential observations, a \gls{cnn}-based policy network is applied. Following the methodology described by~\textcite{mnih2013playing}, the same \gls{cnn} architecture is used for both Image and Sequential observations, as these modalities differ only in the number of input channels—one for Image and four for Sequential, with each channel corresponding to a greyscale image.

The \gls{mlp} policy network comprises an input layer, two hidden layers with \( 64 \) units each, and an output layer. All hidden layers use \(\tanh\) activation functions. The final output layer is linear, producing the mean of the stochastic policy in the continuous action space \( \mathbb{R}^{21} \).

The \gls{cnn} architecture consists of three convolutional layers: the first applies \( 32 \) filters of size \(8 \times 8\) with stride \( 4 \), the second uses \(64\) filters of size \(4 \times 4\) with stride \(2\), and the third uses \( 64 \) filters of size \(3 \times 3\) with stride \( 1 \). Each convolutional layer is followed by a \(\mathrm{ReLU}\) activation. The output is flattened and passed through a fully connected layer with \( 512 \) units and \(\mathrm{ReLU}\) activation, which then feeds into the final linear output layer that parameterizes the policy. All networks are trained using the \gls{adam} optimizer with a fixed learning rate of \( 0.0003 \).

\subsection{Traditional Path Planning}

To emphasize our choice of using a neural network policy as the expert, we conducted a defining experiment. We initially used the A* planner~\autocite{hart1968formal} to identify the most straightforward route from a starting position to the goal, ensuring sampling was restricted to the aortic arch. This plan is depicted in Fig.~\ref{appfig:path-planning}. However, the results revealed that the shortest path is not always the best. The cannulation efficiency along this route was suboptimal, and there was no standardized method to determine the exact actuation for the guidewire tip to follow this path. Expanding our research, we utilized a \gls{mlp} to predict the next action from a sequence of observations. Yet, this model also faced difficulties in following the desired route, highlighting the intrinsic complexities of the task. The agent's actual path, showing the guidewire's deviation, can be seen below:

\begin{figure}[h]
	\centering
	\includegraphics{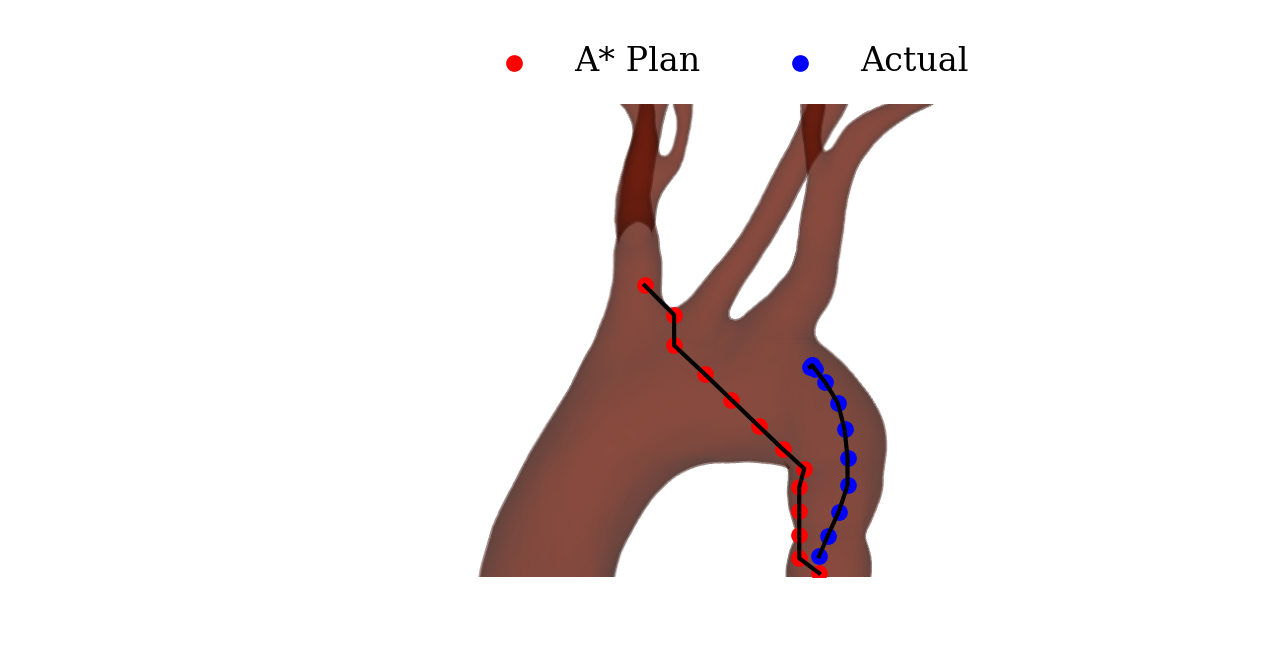}
	\caption{Path Planning}
	\label{appfig:path-planning}
\end{figure}

\section{Experiments}

\begin{figure}[ht]
	\centering
	\subfloat[Type-I Aortic Arch]{\includegraphics[width=0.4\linewidth]{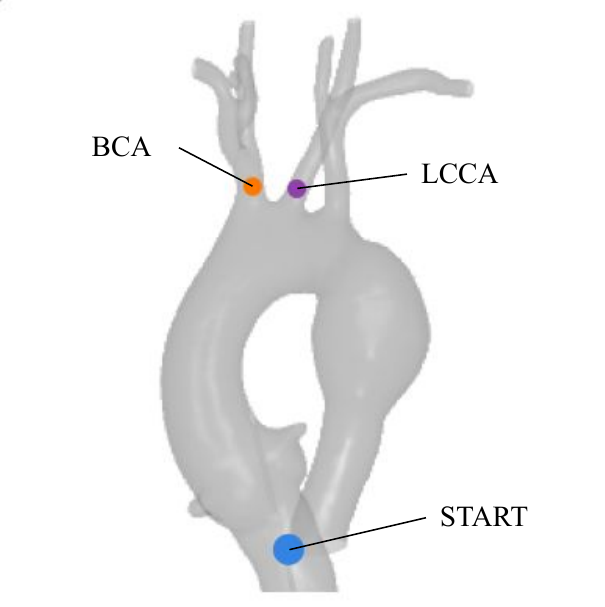}}
	\subfloat[Type-II Aortic Arch]{\includegraphics[width=0.4\linewidth]{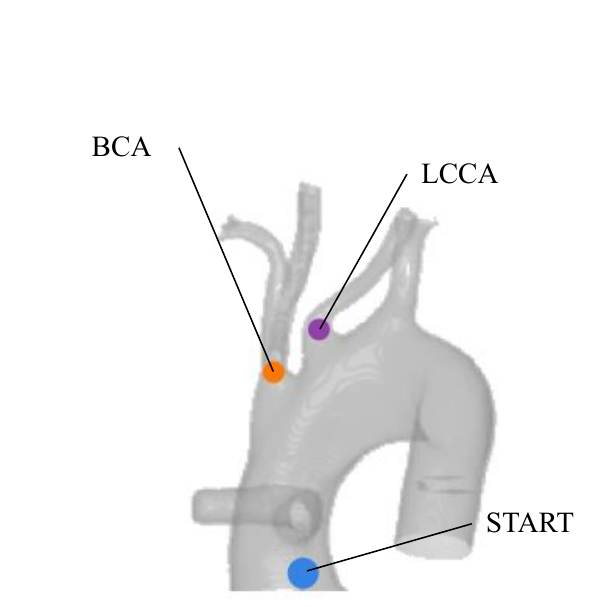}}
	\mycaption{Experiment Starting Configuration}{The figure depicts the navigation task employed in the Type-I and Type-II Aortic Arches. The catheter is initially situated within the ascending aorta with the task of navigating towards the \gls{bca} or the \gls{lcca}. The task finishes when the tip of the catheter is situated within proximity of $8mm$ of the targets. \copyright 2024 IEEE}
	\label{ch3fig:rl_experiment_setup}
\end{figure}

In this section, comprehensive experiments are conducted to evaluate the performance and utility of the simulation framework, \textit{CathSim}. The fidelity of \textit{CathSim} is first verified by comparing its behaviour to that of the physical robotic system, \textit{CathBot}, under similar conditions. This validation ensures that the simulator accurately replicates the dynamics and interactions observed in real-world settings.

Subsequently, the capability of \textit{CathSim} to facilitate autonomous cannulation using state-of-the-art \gls{reinforcement learning} algorithms is demonstrated. Agents are trained within \textit{CathSim} and evaluated based on their ability to navigate the catheter through anatomically realistic environments, such as Type-I and Type-II aortic arches, and achieve precise targeting of key arterial branches.

\paragraph{Training Details.} All experiments were executed on a system equipped with an NVIDIA RTX 2060 GPU, running Ubuntu 22.04 LTS. The system was configured with an AMD Ryzen 7 5800X 8-Core Processor with 16 threads and \SI{64}{\giga\byte} of RAM. All implementations were developed using PyTorch, and the \gls{reinforcement learning} algorithms were implemented using the Stable-Baselines3 library~\cite{stable-baselines3}.

Each episode during training is associated with two terminal conditions:
\begin{itemize}
	\item \textit{Time-bound termination}: The episode is terminated once a predefined maximum number of steps is reached.
	\item \textit{Goal-bound termination}: The episode is terminated when the agent successfully achieves the goal \( g \), defined as the catheter tip reaching within \SI{8}{\milli\meter} of the target location.
\end{itemize}

\subsection{CathSim Validation}\label{sec:cathsim-validation}

\begin{figure}[ht]
	\centering
	\includegraphics[width=.7\linewidth]{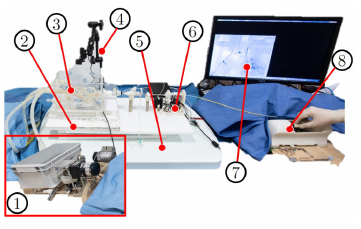}
	\mycaption{Experimental setup of CathBot}{\textit{1)} Pulsatile and continuous flow pumps, \textit{2)} Force Sensor, \textit{3)} Vascular Phantom, \textit{4)} Webcam, \textit{5)} NDI Aurora field generator, \textit{6)} Catheter manipulator (\ie, robotic follower), \textit{7)} Simulated X-ray Screening, and \textit{8)} Master device. Adapted from \textcite{chi2020collaborative} and \textcite{kundrat2021mr}.}
	\label{ch4fig:cathbot-experiment_setup}
\end{figure}

\paragraph{CathSim vs. Real Robot Comparison.} To evaluate the accuracy of the simulator, force measurements from \textit{CathSim} were compared to those obtained in real-world experiments. In the physical experiments~\autocite{kundrat2021mr}, an ATI Mini40 load cell was used to measure interaction forces between the catheter and a Type-I silicone phantom, which was identical to the model used in the simulation. This force-based comparison was selected due to the lack of standardized quantitative metrics for evaluating endovascular simulators~\autocite{Rafii-Tari2017}. Figure~\ref{ch4fig:cathbot-experiment_setup} illustrates the experimental setup.

\paragraph{Statistical Analysis.} To validate the similarity between the simulator and real-world systems, a statistical analysis was performed comparing the empirical force distributions from \textit{CathSim} to those observed in the experiments by \textcite{kundrat2021mr}. Based on the real-world measurements, a cumulative distribution (Fig.~\ref{ch4fig:force-distributions}) was generated by sampling from a Gaussian distribution, which, due to its symmetry, also includes negative values. Although these negative forces are not physically meaningful, they arise naturally from the statistical modelling process and do not impact the validity of the comparison.

The Mann-Whitney U test was applied to determine whether the two distributions were statistically distinguishable. The test yielded a statistic of \( U = 76076 \) and a p-value of \( p \approx 0.445 \), indicating that the observed differences were not statistically significant and were likely attributable to random variation. These results suggest that the force distributions from \textit{CathSim} and the real-world system originate from the same population, confirming that \textit{CathSim} accurately replicates the force interactions experienced in the physical robotic system.

\paragraph{User Study.} To further validate the authenticity and effectiveness of \textit{CathSim} as an endovascular simulator, a user study was conducted with \( 10 \) participants, all of whom were university students with no prior experience in endovascular navigation. The inclusion of novice users was intentional, aiming to assess the accessibility and intuitiveness of the simulator for individuals at the early stages of training. Participants were first shown a fluoroscopic video of an actual endovascular procedure to establish context, after which they interacted with \textit{CathSim} by performing tasks such as cannulating the \gls{bca} and \gls{lcca}.

Upon completion, feedback was collected via a questionnaire in which participants rated \textit{CathSim} across seven criteria using a 5-point Likert scale~\autocite{likert1932technique}:

\begin{enumerate}
	\item \textit{Anatomical Accuracy:} How effectively did the simulator replicate the anatomy and structure of blood vessels?
	\item \textit{Navigational Realism:} How closely did the simulator emulate the visual experience of a real endovascular procedure?
	\item \textit{User Satisfaction:} What was the level of satisfaction regarding the simulator's overall performance and functionality?
	\item \textit{Friction Accuracy:} How accurately did the simulation portray the resistance and friction of the guidewire against vessel walls?
	\item \textit{Interaction Realism:} How realistic were the visual depictions of the guidewire's interactions with the vessel walls?
	\item \textit{Motion Accuracy:} Did the motion of the guidewire align with expectations for a real guidewire's movement?
	\item \textit{Visual Realism:} How visually authentic was the guidewire simulation?
\end{enumerate}

The results, summarized in Table~\ref{ch4tab:user-study}, indicate positive feedback across all criteria. However, participants identified the visual experience of the simulator as an area requiring further improvement. The use of student participants, while enabling assessment of usability for novice users, may limit the generalizability of the findings to experienced clinicians. Nevertheless, this cohort provides valuable insight into the simulator’s learnability and initial user engagement, which are critical for training applications.

\begin{sidebysidefigures}[0.6]
	\begin{leftfigure}
		\centering
		\includegraphics[width=0.9\linewidth]{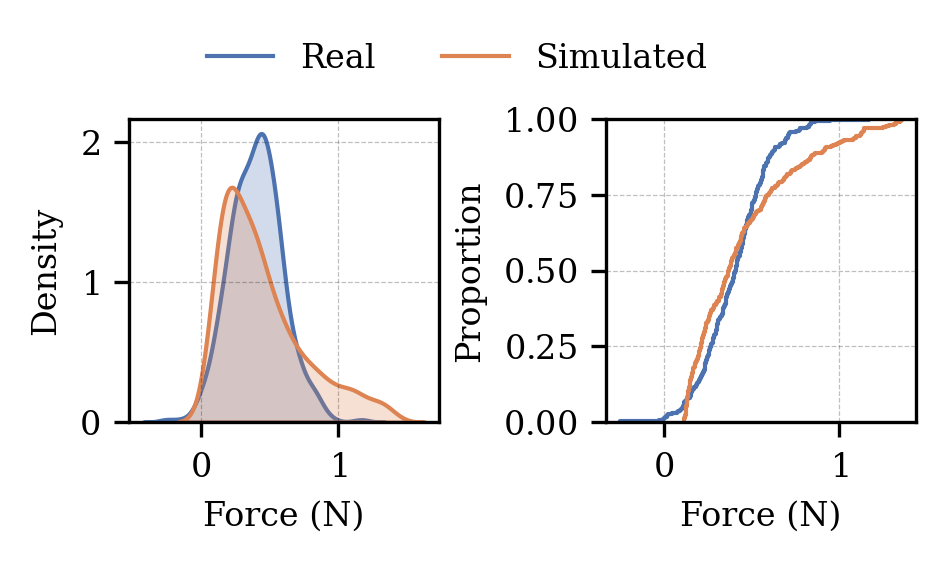}
		\mycaptionof{figure}{Simulated and Real Force Comparison}{Comparison between the simulated force from our CathSim and real force from the real robot. \copyright 2024 IEEE}\label{ch4fig:force-distributions}
	\end{leftfigure}%
	\begin{rightfigure}
		\captionof{table}{User-study results.}
\begin{tabular}{l S}
	\toprule
	\thead{Question}    & {\thead{Average}} \\
	\midrule
	Anatomical Accuracy & 4.57(0.53)        \\
	Navigation Realism  & 3.86(0.69)        \\
	User Satisfaction   & 4.43(0.53)        \\
	Friction Accuracy   & 4.00(0.82)        \\
	Interaction Realism & 3.75(0.96)        \\
	Motion Accuracy     & 4.25(0.50)        \\
	Visual Realism      & 3.67(1.15)        \\
	\bottomrule
\end{tabular}
\label{ch4tab:user-study}
	\end{rightfigure}
\end{sidebysidefigures}

\subsection{Reinforcement Learning Results}

\subsubsection{Setup}

Autonomous catheterization was performed on two principal arteries within the aortic arch: the \acrfull{bca} and the \acrfull{lcca}. For both configurations, the catheter tip was positioned at a predefined starting location within the ascending aorta, and training was terminated once full insertion into the target artery was achieved. The procedure was identical for both Type-I and Type-II aortic arches. Figure~\ref{ch3fig:rl_experiment_setup} illustrates the experimental setup.

The model was trained for a total of \(6 \times 10^5\) time steps. Each episode was initialized with a random catheter displacement of \(1\)~mm, followed by navigation through the ascending aorta to the designated target. Episodes were terminated either upon successful insertion into the target artery or after exceeding a limit of \(2{,}000\) steps. During training, data on contact points and exerted forces were collected at every time step, with forces computed according to Eq.~\ref{eq:force_computation}, enabling the generation of force heatmaps overlaid on RGB virtual images.

The model’s performance was evaluated over \(30\) episodes. The maximum and mean forces exerted during \(n\) samples were calculated as follows:

\begin{equation}\label{eq:force_computation}
	f_{\rm max} = \max\limits_{t \in n} f_t, \quad \quad
	f_{\rm mean} = \frac{1}{n}\sum\limits_{t=1}^n f_t
\end{equation}

The training time varied depending on the algorithm. \gls{ppo} completed training in approximately \(1\) hour, whereas \gls{sac} required \(8\) hours, making \gls{ppo} approximately five times faster. Despite the difference in training time, the simulator maintained a performance rate of \(60\) FPS in the image-based environment and \(80\) FPS in the internal-based environment.

\subsection{Force Analysis}

\begin{figure}[ht]
	\centering
	\subfloat[Type-I - BCA]{\includegraphics[width=0.2\linewidth]{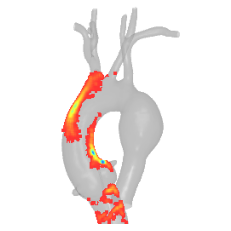}}\hfill
	\subfloat[Type-II - BCA]{\includegraphics[width=0.2\linewidth]{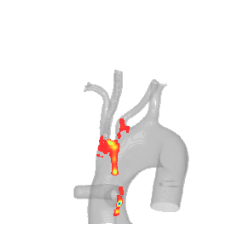}}\hfill
	\subfloat[Type-I - LCCA]{\includegraphics[width=0.2\linewidth]{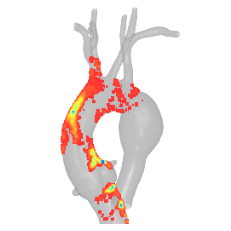}}\hfill
	\subfloat[Type-II - LCCA]{\includegraphics[width=0.2\linewidth]{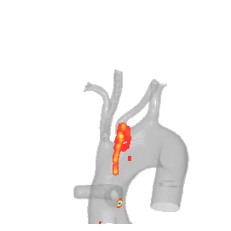}}\hfill
	\subfloat{\includegraphics[width=0.12\linewidth]{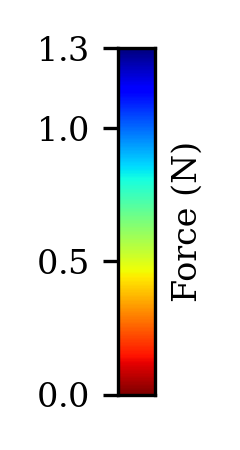}}
	\mycaption{Force Heatmaps}{The heatmaps show the interaction between the catheter and the aortic walls during navigation. (a) and (b) correspond to the \gls{bca} target for the Type-I and Type-II Aortic Arches, respectively, while (c) and (d) correspond to the \gls{lcca} target for the Type-I and Type-II Aortic Arches. Greater forces are observed in the Type-I Aortic Arch, where complex maneuvers are required to reach the targets, compared to the Type-II Aortic Arch. \copyright 2024 IEEE }
	\label{ch3fig:heatmap}
\end{figure}

Figure~\ref{ch3fig:heatmap} illustrates the force heatmaps derived from catheter interactions with the aortic walls during navigation. Subfigures (a) and (b) represent navigation toward the \gls{bca} target for the Type-I and Type-II aortic arches, respectively, while (c) and (d) correspond to navigation toward the \gls{lcca} target for the same anatomical variations. These heatmaps offer insights into the dynamics of catheter-aorta interactions, revealing variations in force intensity and distribution as a function of aortic arch morphology.

\begin{itemize}
	\item \textit{Type-I Aortic Arch}: More complex manoeuvres were required during navigation, resulting in increased forces exerted on the aortic walls, particularly at bends.
	\item \textit{Type-II Aortic Arch}: Navigation involved less complex trajectories, leading to lower force interactions between the catheter and vessel walls.
\end{itemize}

\begin{figure}[ht]
	\centering
	\includegraphics[width=.9\linewidth]{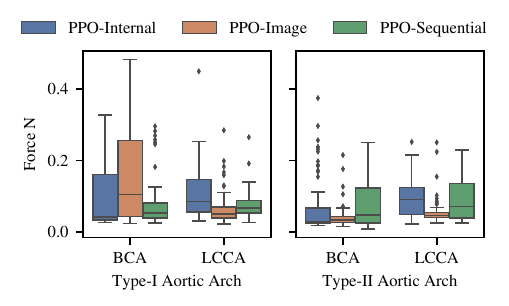}
	\mycaption{Force Distribution}{Maximum forces exerted by the catheter on the aortic walls during \gls{ppo} training evaluation. The left plot shows results for the Type-I Aortic Arch, characterized by greater interactions, while the right plot shows results for the Type-II Aortic Arch, where navigation is more direct. \copyright 2024 IEEE.}
	\label{ch3fig:boxplot}
\end{figure}

Figure~\ref{ch3fig:boxplot} presents an analysis of the force interactions by depicting the maximum forces exerted during evaluation episodes for \gls{ppo}. The relationship between anatomical configuration and the catheter’s interaction dynamics is revealed as follows:

\begin{itemize}
	\item In the \textit{Type-I Aortic Arch} (left), navigation required increased reliance on vessel walls for intricate manoeuvres, resulting in higher exerted forces.
	\item In the \textit{Type-II Aortic Arch} (right), navigation followed more direct paths, leading to overall lower interaction forces.
\end{itemize}

Most force values across both arch types were observed within the \(0.0\) to \(0.2\)~N range. However, outliers exceeding \(0.4\)~N were present. This force range is reasonable for vascular interventions, as forces below \(0.2\)~N are typically sufficient for catheter manipulation without risking vessel damage, while occasional higher forces up to \(0.4\)~N may occur during challenging navigational segments or in areas of increased vessel tortuosity~\cite{li2024modeling}. Notably, the \gls{bca} within the Type-I Aortic Arch exhibited a wider interquartile range compared to the \gls{lcca}, indicating greater variability in force interactions during \gls{bca} targeting.

This analysis underscores the influence of anatomical variations on catheterization dynamics. The Type-I Aortic Arch presents greater challenges, requiring higher forces for complex navigation manoeuvres when compared to the more straightforward pathways in the Type-II Aortic Arch.

\clearpage
\begin{landscape}
	\hspace*{\fill}
	\vspace*{\fill}
	\begin{table}[htbp]
		\centering
		\caption{Results Summary}\label{ch3tab:cathsim_initial_results}
		\begin{originaltabular}{
l 
l 
S[table-format=3(2)] 
S[table-format=3(2)] 
S[table-format=.3(.3)] 
S[table-format=.3(.3)] 
S[table-format=.3(.3)] 
S[table-format=.3(.3)] 
S[table-format=3] 
S[table-format=3]
}
			\toprule
			 &  & \multicolumn{2}{c}{\thead{Reward}}  & \multicolumn{2}{c}{\thead{Mean Force $(N\downarrow)$}} & \multicolumn{2}{c}{\thead{Max Force $(N\downarrow)$}} & \multicolumn{2}{c}{Success~\%}                                                                                  \\
			\cmidrule{3-10}
			\thead{Aorta}						 & \thead{Observation}                                 & {BCA}             & {LCCA}                                                 & {BCA}                                                 & {LCCA}                         & {BCA}                  & {LCCA}                 & {BCA}         & {LCCA}       \\
			\midrule
			\multirow{6}{*}{Type-I}
			                                             & PPO-Internal                     & -70(28)           & -154(66)                                               & 0.011(0.021)                                          & 0.008(0.015)                   & 0.121(0.095)           & 0.123(0.089)           & 53            & 07           \\
			                                             & SAC-Internal                     & -54(36)           & -158(83)                                               & 0.007(0.014)                                          & 0.006(0.015)                   & 0.059(0.047)           & 0.095(0.039)           & 77            & 07           \\
			                                             & PPO-Image                        & -65(52)           & \bfseries -140(96)                                     & \bfseries 0.005(0.007)                                & 0.007(0.016)                   & \bfseries 0.048(0.020) & 0.076(0.059)           & 83            & \bfseries 37 \\
			                                             & SAC-Image                        & -391(157)         & -196(5)                                                & 0.024(0.035)                                          & \bfseries 0.003(0.003)         & 0.258(0.094)           & \bfseries 0.059(0.028) & 00            & 00           \\
			                                             & PPO-Sequential                   & \bfseries -57(36) & -336(55)                                               & 0.006(0.009)                                          & 0.010(0.015)                   & 0.045(0.022)           & 0.094(0.036)           & \bfseries  97  & 00           \\
			                                             & SAC-Sequential                   & -200(20)          & -227(12)                                               & 0.003(0.003)                                          & 0.003(0.003)                   & 0.043(0.009)           & 0.061(0.022)           & 00            & 00           \\
			\midrule                                                                                                                                                                                                            
			\multirow{6}{*}{Type-II}
			                                             & PPO-Internal                     & -52(48)           & -22(30)                                                & 0.016(0.023)                                          & 0.011(0.015)                   & 0.118(0.103)           & 0.068(0.040)           & 50            & 87           \\
			                                             & SAC-Internal                     & \bfseries 0(0)    & -67(30)                                                & 0.014(0.010)                                          & 0.010(0.016)                   & \bfseries 0.030(0.007) & 0.121(0.044)           & \bfseries  100 & 27           \\
			                                             & PPO-Image                        & -20(47)           & \bfseries -8(10)                                       & 0.017(0.025)                                          & 0.015(0.022)                   & 0.046(0.047)           & 0.066(0.054)           & 87            & \bfseries 97 \\
			                                             & SAC-Image                        & -68(91)           & -54(7)                                                 & \bfseries 0.004(0.005)                                & \bfseries 0.005(0.005)         & 0.042(0.013)           & \bfseries  0.050(0.008) & 73            & 03           \\
			                                             & PPO-Sequential                   & -47(49)           & -80(92)                                                & 0.011(0.017)                                          & 0.011(0.018)                   & 0.068(0.059)           & 0.066(0.050)           & 57            & 67           \\
			                                             & SAC-Sequential                   & -99(18)           & -85(15)                                                & 0.005(0.007)                                          & 0.006(0.008)                   & 0.055(0.017)           & 0.100(0.023)           & 03            & 00           \\
			\bottomrule
		\end{originaltabular}
	\end{table}
	\hspace*{\fill}
	\vspace*{\fill}
\end{landscape}
\clearpage

\subsection{Quantitative Results}\label{ch3sec:quantitative_results}

Table~\ref{ch3tab:cathsim_initial_results} summarizes the performance of all evaluated methods in terms of force, reward, and success rate. In experiments conducted on the Type-I Aortic Arch, \gls{ppo} with sequential observations yielded the highest reward when cannulating the \gls{bca} target, although a significant decline in performance was observed when targeting the \gls{lcca}. In contrast, the use of singular image observations produced more consistent rewards, with cannulation of the \gls{lcca} achieving the highest success rate (\(37\%\)) and the \gls{bca} cannulation success rate (\(83\%\)) closely matching that of the sequential observation setup.

The observed performance disparity between \gls{bca} and \gls{lcca} cannulation can be attributed to anatomical differences within the aortic arch. As shown in Fig.~\ref{ch3fig:rl_experiment_setup}, the \gls{bca} is positioned closer to the initial navigation direction, making it more accessible for both human operators and reinforcement learning agents. In contrast, the \gls{lcca} is located further along the navigation path and is surrounded by additional vascular branches, thereby increasing task complexity.

Cannulation within the Type-II Aortic Arch was found to be comparatively more straightforward, as reflected by higher success rates across all evaluated methods. The reduced reliance on vessel walls during navigation enabled smoother catheterization. This trend is further illustrated in Fig.~\ref{ch3fig:heatmap}, where lower forces were exerted on the aortic walls in the Type-II configuration compared to the Type-I arch.

Overall, the results indicate that \gls{ppo} outperformed \gls{sac} in autonomous cannulation tasks, particularly when image-based observations were employed.

\subsection{Rewards Analysis}

\begin{figure}[ht]
	\centering
	\includegraphics[width=.8\linewidth]{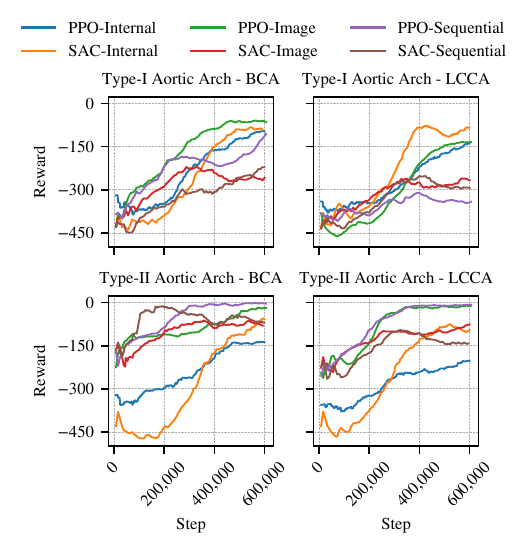}
	\mycaption{Training Rewards Plots}{Training rewards for autonomous agents using an on-policy algorithm (\gls{ppo}) and an off-policy algorithm (\gls{sac}) across two aortic arch types (Type-I and Type-II) and two targets (\gls{bca} and \gls{lcca}). Each curve represents the mean reward averaged over the last 100 steps, with experiments initialized using a fixed random seed of 42. Performance trends vary by anatomical setting: in Type-I arches, image observations yield the strongest results (particularly with \gls{ppo}), while in Type-II arches, sequential observations with \gls{ppo} achieve the highest rewards. Internal observations benefit \gls{sac} in Type-I settings but underperform in Type-II. \copyright 2024 IEEE}
	\label{ch3fig:rewards}
\end{figure}

At each time step, the mean reward over the last \(100\) steps was computed, with results presented in Fig.~\ref{ch3fig:rewards}. Both \gls{ppo} and \gls{sac} exhibited stable learning behaviour with gradual improvements and eventual signs of convergence. The relative performance, however, varied considerably depending on the aortic arch type, target, and observation modality. 

In the Type-I aortic arch, \gls{ppo} with image observations achieved the highest reward for the \gls{bca} target, followed closely by \gls{sac} with internal observations. For the \gls{lcca} target, \gls{sac} with internal observations achieved the best results, slightly outperforming \gls{ppo} with image observations. Across both targets in Type-I arches, sequential observations (particularly with \gls{ppo}) consistently ranked lowest, suggesting that they fail to provide the necessary representational richness for reliable navigation.

In contrast, the Type-II aortic arch revealed a different trend. For both \gls{bca} and \gls{lcca}, \gls{ppo} with sequential observations achieved the highest reward, surpassing all other settings. \gls{ppo} with image observations remained competitive, consistently ranking second. By comparison, \gls{ppo} with internal observations consistently performed worse across both targets in Type-II arches, indicating that internal-only states are insufficient to capture the greater anatomical complexity of these geometries. 

Taken together, these results highlight that the most effective algorithm–observation pair depends on the vascular configuration. Image-based observations are advantageous in Type-I arches, particularly when paired with \gls{ppo}, while sequential observations become critical for Type-II arches, where they enable \gls{ppo} to achieve the best performance. Internal observations favour \gls{sac} in Type-I settings, but degrade performance when paired with \gls{ppo} in Type-II anatomies. These findings emphasize the importance of tailoring both the learning algorithm and the observation space to the anatomical context in order to maximize performance.

\subsection{CathSim Speed Evaluation}

\begin{sidebysidefigures}[.5]
	\begin{leftfigure}
		\adjustimage{left}{assets/episode_time.png}
		\captionof{figure}{CathSim training speed.}
		\label{app1fig:fps}
	\end{leftfigure}%
	\begin{rightfigure}
		\vspace{0.35cm}
		\captionof{table}{Comparative training times}\label{app1tab:training-time}
\begin{tabular}{l r r}
	\toprule
	\multirowcell{2}{\thead{Algorithm}} & \multicolumn{2}{c}{\thead{Training Time~(\(\mathbf{\unit{\hour}}\))}}                   \\
	\cmidrule{2-3}
	                                    & \thead{BCA}                                                           & \thead{LCCA}    \\
	\midrule
	Image                               & $3.00 \pm 0.11$                                                       & $2.54 \pm 0.17$ \\
	Image+Mask                          & $4.20 \pm 0.05$                                                       & $4.60 \pm 1.29$ \\
	Internal                            & $2.38 \pm 0.15$                                                       & $2.20 \pm 0.18$ \\
	Internal+Image                      & $3.15 \pm 0.28$                                                       & $3.54 \pm 0.29$ \\
	\textbf{ENN}                        & $4.61 \pm 0.22$                                                       & $4.83 \pm 0.41$ \\
	\bottomrule
\end{tabular}

	\end{rightfigure}
\end{sidebysidefigures}

As illustrated in Fig.~\ref{app1fig:fps}, we provide a comparison of frames per second (FPS) for the various algorithms we employed during model training. It is evident that utilizing solely the internal state space, comprised of joint positions and velocities, facilitates expedited training processes. In contrast, integrating all modalities into the training process results in its deceleration. The most significant computational demand arises from the dual convolutional neural networks utilized in both the image and mask representations. However, despite this load, the algorithms exhibit respectable computational speed, even during the training phase. Our simulator supports approximately $40$ to $80$ frames per second performance for all implemented algorithms, underscoring the computational speed of our simulation environment. Moreover, we provide the training times in terms of hours for the different modalities in Table~\ref{app1tab:training-time}.

\section{Discussion and Conclusions}

CathSim has been introduced as an open-source simulation environment designed to function as a comprehensive benchmarking platform for autonomous endovascular navigation. Through the development and evaluation of diverse algorithms for autonomous cannulation, CathSim eliminates the reliance on physical robotic systems, thereby mitigating the risks and costs typically associated with experimental hardware. The simulator is intended to serve a dual purpose: to provide researchers with a robust environment for algorithm development and evaluation, and to offer healthcare professionals a risk-free platform for skill acquisition and procedural refinement.

Given the rapid advancements in medical robotics, the need for controlled and flexible simulation environments has become increasingly apparent. This need is effectively addressed by CathSim, which enables the testing and optimization of novel procedures and techniques prior to clinical deployment. Furthermore, the open-source nature of CathSim encourages collaborative innovation and facilitates knowledge sharing across both scientific and medical communities.

Planned future enhancements --- including the integration of deformable anatomical models, improvements in imaging fidelity, and the development of more realistic real-time interaction dynamics --- are expected to further strengthen CathSim as a pivotal tool bridging the gap between simulation and clinical practice. By addressing these challenges, CathSim holds the potential to significantly advance the field of autonomous endovascular navigation and to contribute meaningfully to medical robotics, surgical training, and healthcare education. The simulation environment developed in this chapter serves as the foundation for the Expert Navigation Network presented in the next chapter, which leverages CathSim's multimodal data to enhance navigation precision.

\onehalfspacing
\chapter{Expert Network for Autonomous Navigation}
\chaptermark{ENN}
\glsresetall

\begin{cabstract}
	Endovascular robots have been actively developed in both academic and industrial contexts. However, progress toward autonomous catheterization has often been hindered by the absence of robust navigation strategies capable of generalizing across diverse anatomical and procedural variations. Furthermore, the acquisition of large-scale datasets for training machine learning algorithms in endovascular navigation remains largely infeasible due to the complexity of medical procedures and the scarcity of expert demonstrations. In this chapter, an Expert Navigation Network (ENN) is introduced as a multimodal framework designed to address these challenges. ENN emphasizes real-time adaptability and leverages multimodal data sources to enhance decision-making and navigation performance. Validation is performed using a combination of simulated and real-world robotic setups, demonstrating superior performance compared to traditional navigation approaches in complex endovascular tasks. The experimental results indicate that ENN holds strong potential for advancing research in autonomous endovascular navigation, serving as a bridge between simulation and real-world application.
\end{cabstract}

This chapter presents the work from the following publication:

\fullcite{jianu2024autonomous}

\section{Autonomous Navigation with Expert Navigation Network}

Drawing inspiration from the domains of autonomous driving and human-robot interaction~\autocite{kiran2021deep,kim2021towards}, an expert navigation network is proposed to support downstream tasks in autonomous endovascular interventions. Using \textit{CathSim}~\autocite{jianu2024cathsim}, a substantial dataset of labelled training samples is generated, enabling the model to learn from a broad range of simulated scenarios that reflect diverse procedural variations. This exposure to varied procedural contexts enhances the model's capacity to generalize navigation strategies to real-world applications~\autocite{zhao2020sim}.

In addition, data that is typically unavailable in real-world systems, such as force measurements~\autocite{okamura2009haptic} and shape representations~\autocite{shi2016shape}, are incorporated to improve the robustness and precision of the expert navigation system~\autocite{puschel2022robot}. The simulation environment offers a comprehensive set of sensing modalities, including force feedback and shape representation, which are absent in clinical endovascular procedures where surgeons rely solely on X-ray imaging. By integrating these advanced modalities, the expert navigation system is designed to deliver both high navigation accuracy and enhanced sample efficiency, thereby optimizing learning performance under minimal data requirements.

\begin{sidebysidefigures}[.50]
	\begin{leftfigure}
		\centering
		\includegraphics[width=\textwidth]{figures/autocath_expert_model.pdf}
		\vspace{-1ex}
		\captionof{figure}{The Expert Navigation Network.}
		\label{ch4subfig:autocath_expert_model}
	\end{leftfigure}%
	\begin{rightfigure}
		\centering
		\includegraphics[width=\textwidth]{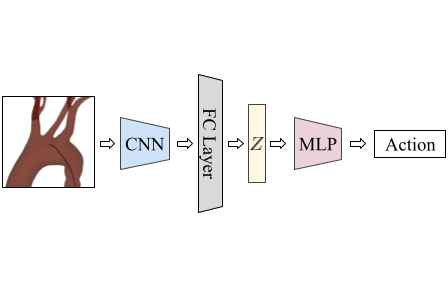}
		\vspace{-0.05cm}
		\captionof{figure}{Downstream IL Network.}
		\label{ch4subfig:autocath_behavioural_clonning_model}
	\end{rightfigure}
\end{sidebysidefigures}

\subsection{Expert Navigation Network}

The \gls{expert navigation network} is implemented as a \textit{multimodal} learning architecture trained within \textit{CathSim}, designed to integrate multiple input types to improve navigation accuracy.

\paragraph{Segmentation Input.} The first input modality consists of semantic segmentation of the guidewire, offering pixel-level information about its position and shape within the environment. This modality enables precise perception of the guidewire’s configuration, thereby facilitating safe and controlled navigation through vascular pathways.

\paragraph{Kinematic Input.} The second input modality comprises the guidewire’s joint positions and velocities, which represent its kinematic and dynamic behaviour~\autocite{tassa2018deepmind}. These parameters contribute to smoother and more efficient navigation compared to previously proposed approaches~\autocite{rafii2012assessment,song2022learning}.

\paragraph{Visual Input.} The third modality incorporates images captured from a top-view camera, offering a visual perspective of the navigation environment. This input provides essential contextual information, enabling the identification of vessel structures and the detection of obstacles that may impact navigation~\autocite{cho2021image}.

\paragraph{Multimodal Feature Fusion.} A \gls{cnn} is employed to extract high-level visual features from both the segmentation maps and top-view images. In parallel, a \gls{mlp} processes the joint positions and velocities to encode kinematic information related to the guidewire. The resulting feature maps are flattened to reduce dimensionality and subsequently concatenated to form a unified representation of the multimodal data. This combined representation is then passed through an \gls{mlp}, which maps the input to a compact feature vector \( Z \) that encapsulates the sensory modalities required for trajectory learning.

\paragraph{Policy Learning.} By integrating these diverse modalities, the expert navigation network learns the complex mapping between multimodal sensory inputs and the guidewire’s desired trajectory. The resulting feature vector \( Z \) is used to train a \gls{sac} policy \( \pi \)~\autocite{haarnoja2018soft}, which serves as the decision-making component within the broader \gls{reinforcement learning} framework. An overview of the complete architecture, including modality integration and the training pipeline, is depicted in Fig.~\ref{ch4subfig:autocath_expert_model}.

\paragraph{Experimental Setup.} The experiments were conducted on a system equipped with an NVIDIA RTX 2060 GPU ($33\unit{\mega\hertz}$) and Ubuntu 22.04 LTS as the operating system. The system also featured an AMD Ryzen 7 5800X 8-core processor with 16 threads and $64\unit{\giga\byte}$ of RAM. All experiments were implemented using PyTorch, and the \gls{sac} algorithm was executed via the Stable-Baselines3 library~\autocite{stable-baselines3}. Training was carried out for \( 6 \times 10^5 \) time steps across $5$ random seeds, with training durations ranging from $2$ to $5$~hours depending on computational load and task complexity.

Each episode was terminated upon satisfying one of two conditions: a time-bound limit (termination after a fixed number of steps), or a goal-bound condition (successful achievement of the goal \( g \in \mathrm{G} \)).

\begin{table}[ht]
	\centering
	\mycaption{The network architectures for ENN}{The CNN is utilized to extract the image-based features (\textit{i.e.}, grayscale image and binary mask) whilst the MLP is used to extract the internal features (\textit{i.e.}, joint positions and joint velocities)}\label{ch4tab:network-architectures-cnn-mlp}
	\begin{tabular}{c c c S[table-format=6,scientific-notation=false] c c c c }
		\toprule
		\thead{Network}       & \thead{Layer (type)} & \thead{Output Shape} & {\thead{Param \#}} & {\thead{Nonlinearity}} & {\thead{Kernel}} & {\thead{Stride}} & {\thead{Padding}} \\
		\midrule
		\multirowcell{7}{CNN} & Input                & (1, 80, 80)          & 0                  & ---                  & ---  &---   & ---                   \\
		                      & Conv2D               & (32, 19, 19)         & 2080               & ReLU                 & 8   & 4 &  0  \\
		                      & Conv2D               & (64, 8, 8)           & 32832              & ReLU                 & 4   & 2 &  0  \\
		                      & Conv2D               & (64, 6, 6)           & 36928              & ReLU                 & 3   & 1 &  0  \\
		                      & Flatten              & (2304)               & 0                  & ---                 & ---  &---   & ---                   \\
		                      & Linear               & (256)                & 590080             & ReLU                 & ---  &---   & ---                   \\
		                      & Linear               & (128)                & 32896              & ReLU                 & ---  &---   & ---                   \\
		\cmidrule{1-8}
		\multirowcell{3}{MLP} & Input                & (1, 336)             & 0                  & ---                 & --- &---   & ---                    \\
		                      & Linear               & (256)                & 86272              & ReLU                & --- &---   & ---                    \\
		                      & Linear               & (128)                & 32896              & ReLU                & --- &---   & ---                    \\
		\bottomrule
	\end{tabular}
\end{table}

\begin{table}[ht]
	\centering
	\captionof{table}{SAC hyperparameters}\label{ch4tab:sac}
	\begin{tabular}{l | r}
		\toprule
		\thead{Hyperparameter}           & \thead{Value}       \\
		\midrule
		Learning Rate                    & \(3\times 10^{-4}\) \\
		Buffer Size                      & \(10^6\)            \\
		Batch Size                       & 256                 \\
		Smoothing Coefficient (\(\tau\)) & 0.005               \\
		Discount (\(\gamma\))            & 0.99                \\
		Train Frequency                  & 1                   \\
		Gradient Steps                   & 1                   \\
		Entropy Coefficient              & 1                   \\
		Target Update Interval           & 1                   \\
		Target Entropy                   & -2                  \\
		\bottomrule
	\end{tabular}
\end{table}

\paragraph{Networks.} Multiple feature extractors are utilized within the \gls{expert navigation network} to isolate and process key multimodal inputs, enabling efficient representation learning. A \gls{cnn}~\autocite{mnih2015human} is employed to extract image-based features, producing two latent feature representations, \( J_I \) and \( J_S \), corresponding to the top-view camera image and the guidewire segmentation map, respectively. The \gls{cnn} comprises three convolutional layers, each followed by a \( \operatorname{ReLU} \) activation function~\autocite{agarap2019deep}, and a final flattening operation to prepare the features for downstream processing. Concretely, we use a Nature-CNN configuration with kernels \((8\times 8,\ 4\times 4,\ 3\times 3)\), strides \((4,\ 2,\ 1)\), and channel widths \(32\!-\!64\!-\!64\), followed by a fully connected projection to a 256-dimensional embedding for each image stream, yielding \(J_I, J_S \in \mathbb{R}^{256}\).

Joint positions \( Q \) and joint velocities \( V \) are concatenated to form a joint feature vector \( J_J = Q \parallel V \) with a dimensionality of \(336\), which is subsequently processed by a \gls{mlp}. This dimensionality arises from concatenating \( n_Q = 168 \) joint position variables and \( n_V = 168 \) joint velocity variables as defined in the MuJoCo model. The extracted features \( J_I \), \( J_S \), and \( J_J \) are concatenated to form a unified feature vector \( J = J_I \parallel J_S \parallel J_J \), which serves as input to the policy network \( \pi(a_t, \theta) \) responsible for decision-making and trajectory generation. Detailed specifications of the \gls{cnn} and \gls{mlp} architectures, including layer parameters, are provided in Table~\ref{ch4tab:network-architectures-cnn-mlp}.

\paragraph{SAC.} The primary \gls{reinforcement learning} algorithm used is \gls{sac}~\autocite{haarnoja2018soft}, selected for its ability to balance exploration and exploitation in stochastic environments. \gls{sac} is a model-free algorithm that concurrently learns a stochastic policy and a value function. Its objective function combines the policy’s expected return with an entropy term, thereby promoting exploration while avoiding premature convergence to suboptimal policies.

The algorithm maintains three networks: a state-value function \( V \), parameterized by \( \psi \), which estimates the value of a given state; a soft \( Q \)-function \( Q \), parameterized by \( \theta \), which evaluates the value of state-action pairs; and a policy network \( \pi \), parameterized by \( \phi \), which defines the action distribution. The policy network consists of linear layers structured to effectively map the fused feature vector \( J \) to the action space. Parameter configurations used for training with \gls{sac} are listed in Table~\ref{ch4tab:sac}.

\subsection{Downstream Tasks}

The effectiveness of the \gls{expert navigation network} and the \textit{CathSim} simulator is demonstrated through downstream tasks such as \gls{imitation learning} and force prediction. These tasks are of practical relevance, as they provide critical support and insights for surgeons during real-world endovascular procedures.

\paragraph{Imitation Learning.} The \gls{expert navigation network} is applied in conjunction with \gls{behavioural cloning}, a supervised form of \gls{imitation learning}~\autocite{hussein2017imitation}, to train a navigation policy. This approach highlights the capability of the simulation environment to extract meaningful representations that can be leveraged in downstream learning tasks.

Initially, expert trajectories are generated by executing the expert policy \( \pi_{\rm exp} \) within the \textit{CathSim} simulator. These trajectories serve as labelled training data, specifying the optimal actions for each encountered state during the navigation process. To emulate conditions consistent with real-world clinical scenarios, the image modality is used as the sole input for training.

Subsequently, the \gls{behavioural cloning} algorithm is trained to replicate the expert’s actions based on these input observations. This is accomplished by optimizing the policy parameters \( \theta \) to minimize the discrepancy between expert and predicted actions using the following loss function:
\begin{equation}
	\mathcal{L}(\theta) = -\mathbb{E}_{\pi_\theta}[ \log \pi_\theta(a|s)] - \beta H(\pi_\theta(a|s)) + \lambda ||\theta||_2^2,
\end{equation}
where \( -\mathbb{E}_{\pi_\theta}[ \log \pi_\theta(a|s)] \) denotes the negative log-likelihood of the expert actions given the observed states. The term \( - \beta H(\pi_\theta(a|s)) \) introduces an entropy penalty, encouraging exploration and diversity in action selection, weighted by \( \beta \). The final term, \( \lambda ||\theta||_2^2 \), applies \( L_2 \) regularization to mitigate overfitting, weighted by \( \lambda \).

To support the learning process, the feature space \( Z \), extracted from the expert policy, is used to train the \gls{behavioural cloning}. Concretely, the \gls{cnn} comprises three convolutional blocks (Conv2D–ReLU–max pooling); the resulting feature maps are flattened and passed through a fully connected projection to form the latent variable \( Z \). A subsequent policy head (fully connected) maps \( Z \) to action scores, and a value head (fully connected) estimates the state value. This feature representation encapsulates key attributes of expert navigation strategies~\autocite{hou2019learning}, providing a compact and expressive input space. A subsequent mapping is then learned from the feature space \( Z \) to the action space, enabling the behavioural cloning model to replicate expert decision-making while requiring fewer input modalities than the original policy. The architecture for the \gls{imitation learning} task is illustrated in Fig.~\ref{ch4subfig:autocath_behavioural_clonning_model}.
\paragraph{Force Prediction.} Force prediction is a critical task in endovascular intervention, as surgeons rely on force feedback to avoid damage to the endothelial walls of blood vessels. Several force prediction methods have been explored, including sensor-based techniques~\autocite{yokoyama2008novel} and image-based approaches~\autocite{song2022learning}.

To assess the force prediction capabilities of the \gls{expert navigation network}, a supervised learning-based regression model is proposed. The architecture integrates a \gls{cnn} for visual feature extraction with an \gls{mlp} head for predicting the contact forces. The model is trained using the following composite loss function:

\begin{equation}
	\mathcal{L} = \mathcal{L}_Z + \mathcal{L}_f = \sum_{i=1}^{D}(Z_i-\hat{Z}_i)^2 + \sum_{i=1}^{D}(f-\hat{f})^2\:,
\end{equation}

where \( \hat{Z} \) denotes the feature vector predicted by the network, \( \hat{f} \) represents the predicted force corresponding to the expert transition \( \pi_{\rm exp} \), and \( D \) is the total number of samples in the dataset. This formulation penalizes discrepancies between both the predicted and ground truth feature vectors \( (Z_i, \hat{Z}_i) \) and the predicted and actual forces \( (f, \hat{f}) \), thereby ensuring accurate modelling of both feature consistency and physical interaction dynamics.

\section{Experiments}

\begin{figure}[ht]
	\begin{centering}
		\includegraphics[width=.8\linewidth]{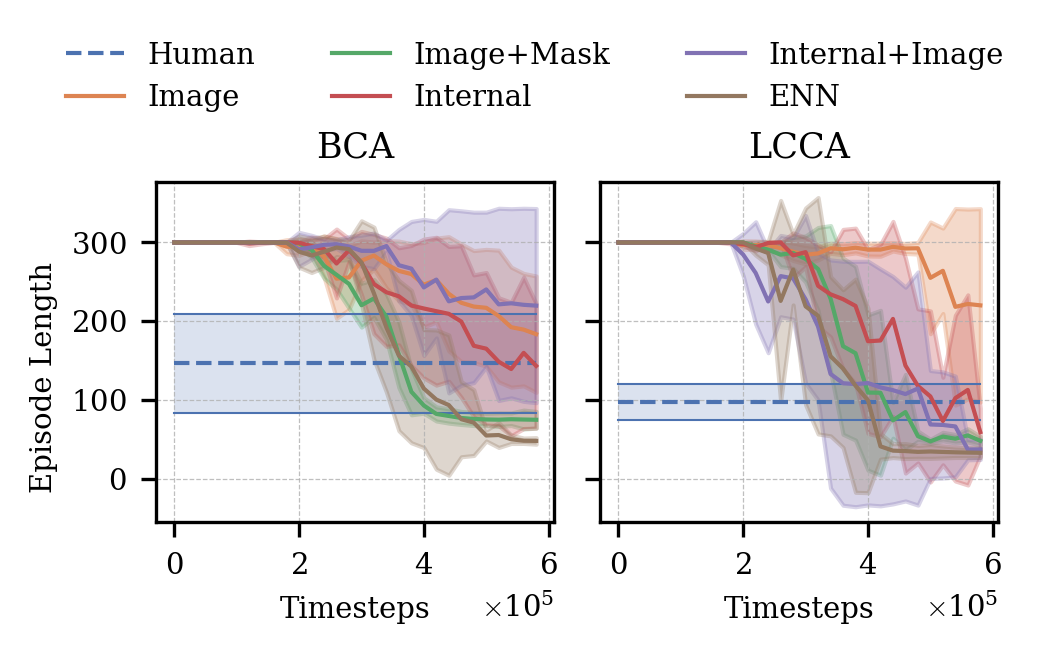}
		\mycaption{Episode Length Distributions}{Episode lengths when utilizing different input modalities.}\label{cathsim-fig:episode-length}
	\end{centering}
\end{figure}

\begin{sidebysidefigures}
	\begin{leftfigure}
		\centering
		\includegraphics[width=.65\linewidth]{assets/setup.pdf}
		\captionof{figure}{Experiment Setup}\label{cathsim-fig:setup}
	\end{leftfigure}%
	\begin{rightfigure}
		\captionof{table}{Force prediction results.}\label{ch4tab:force-prediction}
\centering
\begin{tabular}{l S[table-format=1.3]}
	\toprule
	\thead{Algorithm} & {\thead{MSE~(\unit{\newton})$~\downarrow$}} \\
	\midrule
	Baseline          & 5.0021                                      \\
	FPN (\(1\)K)      & 0.5047                                      \\
	FPN (\(10\)K)     & 0.1828                                      \\
	FPN (\(100\)K)    & 0.0898                                      \\
	\bottomrule
\end{tabular}

	\end{rightfigure}
\end{sidebysidefigures}

\subsection{Expert Trajectory Analysis}

Experiments were conducted to validate the effectiveness of expert trajectories using various input modalities, including image, internal state, and segmentation mask representations. In addition, a professional endovascular surgeon was recruited to manually control the \textit{CathSim} environment, thereby generating a ``Human'' trajectory for comparative analysis.

Catheterization performance was evaluated using a set of quantitative metrics, as described in Subsection~\ref{ch4subsubsec:app_evaluation_metric}: Force (\unit{\newton}), Path Length (\unit{cm}), Episode Length (steps), Safety (\%), Success (\%), and SPL (\%). The experiments were carried out for two navigation targets: the \gls{bca} and the \gls{lcca}. The initial catheter placements and target locations are visualized in Fig.~\ref{cathsim-fig:setup}.

Across all training configurations, \textit{CathSim} maintained operational speeds ranging from \(40\) to \(80\) frames per second, thereby supporting compatibility with real-time applications.

\begin{table}[ht]
	\centering
	\mycaption{Ablation study on input modalities}{ENN (I+V+M) consistently outperforms other input combinations across all metrics. Human expert results are included as reference.}
	\label{ch4tab:main}

	\begin{threeparttable}
	\begin{tabular}{l c S[table-format=1.2(.2)] S[table-format=2.2(2.2)] S[table-format=3.2(3.2)] S[table-format=2(2)] S[table-format=3(2)] S[table-format=3]}
		\toprule
		\thead{Target} & \thead{Input} & {\thead{Force \\ N~$\downarrow$}} & {\thead{Path Length \\ cm~$\downarrow$}} & {\thead{Episode Length \\ s~$\downarrow$}} & {\thead{Safety \\ \%~$\uparrow$}} & {\thead{Success \\ \%~$\uparrow$}} & {\thead{SPL \\ \%~$\uparrow$}} \\
		\midrule
		\multirow{7}{*}{BCA} 
		               & H        & \bfseries 1.02(0.22) & 28.82(11.80)          & 146.30(62.83)         & \bfseries 83(04) & 100(00) & 62           \\
		\cmidrule{2-8}
		               & V        & 3.61(0.61)           & 25.28(15.21)          & 162.55(106.85)        & 16(10)           & 65(48)  & 74           \\
		               & V+M      & 3.36(0.41)           & 18.55(2.91)           & 77.67(21.83)          & 25(07)           & 100(00) & 86           \\
		               & I        & 3.33(0.46)           & 20.53(4.96)           & 87.25(50.56)          & 26(09)           & 97(18)  & 80           \\
		               & I+V      & 2.53(0.57)           & 21.65(4.35)           & 221.03(113.30)        & 39(15)           & 33(47)  & 76           \\
		               & \textbf{I+V+M (ENN)} & 2.33(0.18)           & \bfseries 15.78(0.17) & \bfseries 36.88(2.40) & 45(04)           & 100(00) & \bfseries 99 \\
		\midrule
		\multirow{7}{*}{LCCA} 
		               & H        & \bfseries 1.28(0.30) & 20.70(3.38)           & 97.36(23.01)          & \bfseries 77(06) & 100(00) & 78           \\
		\cmidrule{2-8}
		               & V        & 4.02(0.69)           & 24.46(5.66)           & 220.30(114.17)        & 14(14)           & 33(47)  & 69           \\
		               & V+M      & 3.00(0.29)           & 16.32(2.80)           & 48.90(12.73)          & 33(06)           & 100(00) & 96           \\
		               & I        & 2.69(0.80)           & 22.47(9.49)           & 104.37(97.29)         & 39(17)           & 83(37)  & 79           \\
		               & I+V      & 2.47(0.48)           & 14.87(0.79)           & 37.80(10.50)          & 42(08)           & 100(00) & 100          \\
		               & \textbf{I+V+M (ENN)} & 2.26(0.33)           & \bfseries 14.85(0.79) & \bfseries 33.77(5.33) & 45(05)           & 100(00) & 100          \\
		\bottomrule
	\end{tabular}

	\vspace{0.3em}
	\begin{minipage}{0.9\linewidth}
		\footnotesize
		\textbf{Input Legend:} H = Human, I = Internal, V = Vision (Image), M = Segmentation Mask.
	\end{minipage}
	\end{threeparttable}
\end{table}

\begin{figure}[ht]
	\centering
	\includegraphics[width=.9\linewidth]{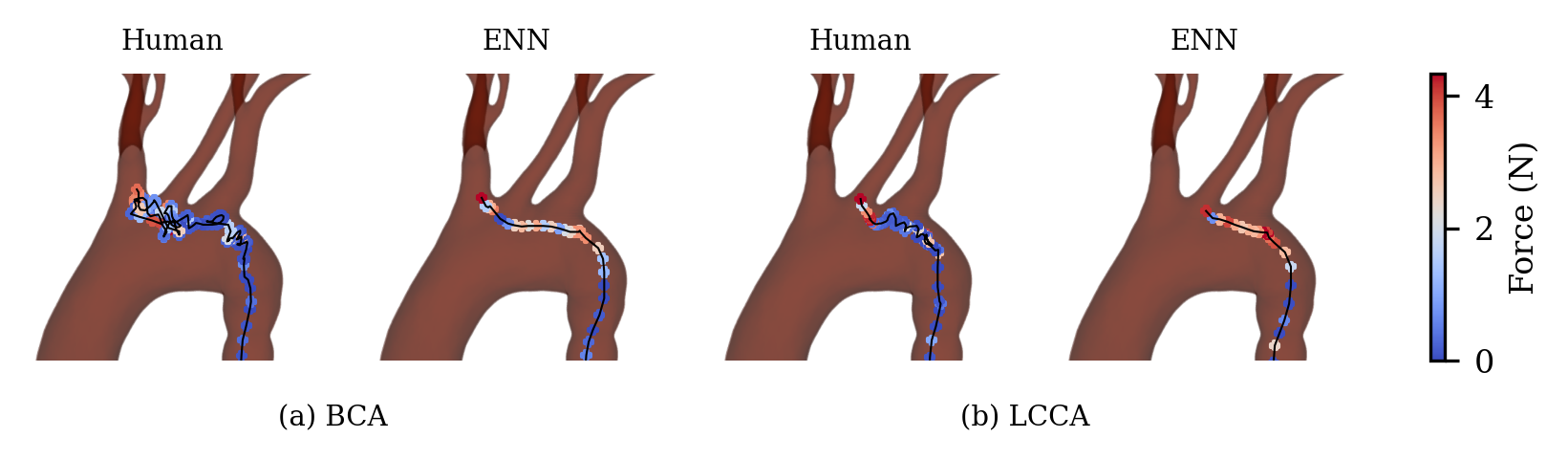}
	\mycaption{Navigation Paths Comparison}{Examples of navigation path from an endovascular surgeon and our \gls{expert navigation network}.}\label{ch4fig:path-visualization}
\end{figure}

\paragraph{Quantitative Results.} Table~\ref{ch4tab:main} presents the ablation study results across various input configurations for both the \gls{bca} and \gls{lcca} targets. The \gls{expert navigation network}, which integrates internal features, vision, and segmentation mask (I+V+M), consistently achieves the best performance in terms of minimal force exertion, shortest path length, and lowest episode duration. While the human expert (H) demonstrates the highest safety score, \gls{expert navigation network} outperforms all ablated configurations across the remaining metrics. These findings highlight the importance of combining multiple input modalities to improve catheter navigation performance. However, it is important to note that \textit{the human surgeon continues to perform the task with greater safety}, which remains a critical consideration in real-world clinical applications.

\paragraph{Expert Trajectory vs. Human Skill.} To evaluate the performance of various iterations of the model during training, the mean episode length was computed and compared with that of the human operator. As illustrated in Fig.~\ref{cathsim-fig:episode-length}, successful navigation of the \gls{bca} environment was achieved after \(6 \times 10^5\) time steps, with approximately half of the model checkpoints outperforming the human benchmark. Moreover, the consistency of navigation across both target vessels demonstrates that the \gls{expert navigation network} achieves robust performance with low inter-seed variance.

Although the \gls{expert navigation network} surpasses the human operator in terms of trajectory efficiency and completion time, it benefits from access to multiple modalities—including image, internal force, and segmentation mask—whereas the human operator relies solely on image-based inputs for task completion.

\paragraph{Navigation Path.} A comparison of the navigation paths produced by the \gls{expert navigation network} and the human operator is shown in Fig.~\ref{ch4fig:path-visualization}. The surgeon’s trajectory follows a more meandering path compared to the expert network’s direct and efficient route. In the case of \gls{bca} navigation (Fig.~\ref{ch4fig:path-visualization}), the human trajectory displays increased irregularity, reflecting the greater anatomical complexity of targeting the \gls{bca} relative to the \gls{lcca}. This additional challenge is likely attributed to the \gls{bca}’s deeper anatomical location within the thoracic region. Despite this, the human operator exerted less force than the \gls{expert navigation network}, illustrating a higher degree of finesse and tactile sensitivity during catheter manipulation.

\subsubsection{Evaluation Metrics}\label{ch4subsubsec:app_evaluation_metric}

\paragraph{Force.} The force exerted by surgical instruments during simulation is a critical metric for evaluating performance and interaction dynamics with the aorta. To measure this accurately, a manual cannulation method was employed within the simulator. At each simulation time step, collision points between the guidewire and the aortic wall were recorded. These data provided key insights into the tridimensional force components acting on the system, including the normal force (\(f_z\)) and the frictional forces (\(f_x\) and \(f_y\)).

The total force magnitude at a given time step \(t\), denoted as \(f_t\), was computed using the Euclidean norm of the force vector:

\begin{equation}
	f_t = \sqrt{ f_{x,t}^2 + f_{y,t}^2 + f_{z,t}^2}
\end{equation}

This calculation offers a comprehensive assessment of the collective impact of the force components, enhancing the understanding of guidewire behaviour and interaction throughout the simulation. Furthermore, it facilitates direct comparison with the experimental force measurements reported by~\textcite{kundrat2021mr}, which were used to validate the simulation model in Sec.~\ref{sec:cathsim-validation}.

\paragraph{Path Length.} The path length was computed by summing the Euclidean distances between consecutive positions of the guidewire tip. At each time step, the guidewire head position \(h_t\) was recorded. The Euclidean distance between positions at time \(t\) and \(t+1\) was calculated as \(d(h_{t}, h_{t+1}) = \sqrt{(h_{t+1} - h_{t})^2}\). The total path length for an episode of length \(n\) was given by:

\begin{equation}
	\operatorname{Path Length} = \sum\limits_{i=1}^n||h_{t+1} - h_{t}||
\end{equation}

\paragraph{SPL.} To evaluate navigation optimality, the \gls{spl} (Success weighted by Path Length) metric, as proposed by~\textcite{anderson2018evaluation}, was applied. This metric compares the actual path length \(p_i\) to the shortest observed path \(l_i\), treating the latter as the optimal reference. The metric was computed using:

\begin{equation}
	\operatorname{SPL}=\frac{1}{N} \sum\limits_{i=1}^N S_i\frac{l_i}{\max(p_i,l_i)},
\end{equation}

where \(S_i\) is a binary success indicator, and \(N\) is the number of episodes. Given that the expert navigation policy consistently outperforms the human operator, the shortest successful path was used as the optimal baseline.

\paragraph{Safety.} Safety was defined based on the frequency of force values exceeding \(2\unit{\newton}\), a threshold informed by real-world findings in~\autocite{kundrat2021mr}. A binary variable \(a \in \{0,1\}\) was introduced, where \(a = 1\) if \(f_t \geq 2\unit{\newton}\). The safety score was computed as:

\begin{equation}
	\operatorname{Safety} = 1 - \frac{1}{N} \sum_{i=1}^N a_i,
\end{equation}

where \(N\) denotes the number of steps per episode. A score of \(0\%\) indicates that excessive force was applied at every time step, while a score of \(100\%\) implies that all forces remained within the safe threshold.

\paragraph{Episode Length.} Episode length was defined as the total number of steps taken to complete a task. This metric offers a measure of efficiency; shorter episodes generally reflect more efficient navigation, provided task quality is not compromised.

\paragraph{Success.} An episode was marked as successful if the agent reached the target location within a fixed limit of \(300\) time steps. This binary metric reflects the agent’s ability to complete tasks under time constraints, emulating the urgency often encountered in clinical scenarios.

\subsection{Downstream Task Results}

\subsubsection{Imitation Learning}

A comparison was performed between the baseline algorithm (Image-only), the expert policy (i.e., \gls{expert navigation network}), and the \gls{behavioural cloning} algorithm enhanced by expert-generated trajectories. The results, presented in Table~\ref{ch4tab:bc-results}, demonstrate that the integration of expert trajectories markedly improves performance under constrained observation conditions.

Relative to the Image-only baseline, the use of expert trajectories in the \gls{behavioural cloning} algorithm led to substantial improvements across multiple evaluation metrics. For both \gls{bca} and \gls{lcca} targets, the enhanced model exhibited lower force application, shorter path and episode lengths, increased safety, and improved success rate and \gls{spl} scores. These findings emphasize the importance of expert-guided learning in simulation and its potential to facilitate sim-to-real transfer in future applications.

\begin{table}[ht]
	\centering
	\mycaption{Imitation Learning Results}{The Imitation Learning results with and without using the Expert Navigation Network (ENN).}\label{ch4tab:bc-results}
	\scriptsize
	\begin{tabular}{l l S[table-format=1.2(.2)] S[table-format=2.2(2.2)] S[table-format=3.2(3.2)] S[table-format=2(2)] S[table-format=3] S[table-format=3]}
		\toprule
		{\thead{Target}}      & {\thead{Algorithm}} & {\thead{Force                                                                                                    \\ (N)~$\downarrow$}} & {\thead{Path Length \\ cm~$\downarrow$}} & {\thead{Episode Length \\ cm~$\downarrow$}} & {\thead{Safety \\ \%~$\uparrow$}} & {\thead{Success \\ \%~$\uparrow$}} & {\thead{SPL \\ \%~$\uparrow$}} \\
		\midrule
		\multirow{3}{*}{BCA}  & ENN                 & 2.33(0.18)           & \bfseries 15.78(0.17) & \bfseries 36.88(2.40) & 45(04)           & 100(00) & \bfseries 99 \\
		                      & Image w\/o. ENN     & 3.61(0.61)           & 25.28(15.21)          & 162.55(106.85)        & 16(10)           & 65(48)  & 74           \\
		                      & Image w. ENN        & \bfseries 2.23(0.10) & 16.06(0.33)           & 43.40(1.50)           & \bfseries 49(03) & 100(00) & 98           \\  \cmidrule{1-8}
		\multirow{3}{*}{LCCA} & ENN                 & \bfseries 2.26(0.33) & 14.85(0.79)           & 33.77(5.33)           & \bfseries 45(05) & 100(00) & 100          \\
		                      & Image w\/o ENN      & 4.02(0.69)           & 24.46(5.66)           & 220.30(114.17)        & 14(14)           & 33(47)  & 69           \\
		                      & Image w. ENN        & 2.51(0.21)           & \bfseries 14.71(0.20) & \bfseries 33.10(2.07) & 43(04)           & 100(00) & 100          \\
		\bottomrule
	\end{tabular}
\end{table}

\paragraph{Network Architecture.} The \gls{behavioural cloning} network architecture consists of a structured sequence of convolutional and fully connected layers. The input, a tensor of shape \( (1, 80, 80) \), is a single-channel greyscale crop (80\(\times\)80\,px) of the fluoroscopic frame, resized from the original image and intensity-normalized to \([0,1]\). It is processed through three Conv2D layers, each followed by a \gls{relu} activation function. The resulting feature maps are then flattened into a single vector, which is passed through a fully connected Linear layer with \gls{relu} activation, and finally through an output Linear layer without activation to produce the raw action scores. This architecture enables effective transformation of visual input into control decisions. Architectural details are provided in Table~\ref{ch4tab:bc-network}.

\begin{table}[ht]
	\mycaption{BC Architecture}{BC Architecture}\label{ch4tab:bc-network}
	\centering
	\begin{tabular}{c c S[table-format=9,scientific-notation=false] c c c c}
		\toprule
		\thead{Layer (type)} &  \thead{Output Shape} & {\thead{Param \#}} & \thead{Nonlinearity} & \thead{Kernel} & \thead{Stride} & \thead{Padding} \\
		\midrule
		Input                & (1, 80, 80)          & 0                  & -                    & ---   & --- & --- \\
	Conv2D                 & (32, 19, 19)         & 2080               & ReLU                 & 8     & 4   & 0   \\
	Conv2D                 & (64, 8, 8)           & 32832              & ReLU                 & 4     & 2   & 0   \\
	Conv2D                 & (64, 6, 6)           & 36928              & ReLU                 & 3     & 1   & 0   \\
	Flatten                & (2304)               & 0                  & -                    & ---   & --- & --- \\
	Linear                 & (256)                & 590080             & ReLU                 & ---   & --- & --- \\
	Linear                 & (128)                & 32896              & ReLU                 & ---   & --- & --- \\
	Linear                 & (2)                  & 258                & None                 & ---   & --- & --- \\
		\bottomrule
\end{tabular}
\end{table}

\begin{table}[ht]
	\centering
	\mycaption{BC Hyperparameters}{BC Hyperparameters}\label{ch4tab:bc-hyperparameters}
	\scriptsize
	\begin{tabular}{c | c}
		\toprule
		\thead{Hyperparameter} & \thead{Default Value} \\
		\midrule
		Batch Size             & \(32\)                \\
		Optimizer              & Adam                  \\
		Learning Rate          & \(1 \times 10^{-3}\)  \\
		Entropy Coefficient    & \(1 \times 10^{-3}\)  \\
		Epochs                 & 800                   \\
	\bottomrule
\end{tabular}
\end{table}


\paragraph{Training Methodology.} Training was conducted using the Adam optimizer~\autocite{kingma2014adam}, with a batch size of \(32\), a learning rate of \(1 \times 10^{-3}\), and an entropy coefficient of the same magnitude. These hyperparameters, summarized in Table~\ref{ch4tab:bc-hyperparameters}, controlled both the optimization step size and the degree of exploratory behaviour. The network was trained for \(800\) epochs, with each epoch encompassing a complete pass through the dataset, allowing for sufficient convergence of the model parameters.

\subsubsection{Force Prediction}

Table~\ref{ch4tab:force-prediction} summarizes the results of fine-tuning the \gls{force prediction network} using varying quantities of trajectory samples generated by the \gls{expert navigation network}. Performance was evaluated based on the \gls{mean squared error} between the predicted and actual force values.

The baseline \gls{mean squared error} was measured at \(5.0021\unit{\newton}\). Following fine-tuning with \(1 \times 10^3\) samples, the \gls{mean squared error} decreased to \(0.5047\unit{\newton}\), illustrating the value of expert-generated trajectories. This trend continued as additional samples were introduced: \(1 \times 10^4\) samples reduced the \gls{mean squared error} to \(0.1828\unit{\newton}\), and \(1 \times 10^5\) samples led to a further reduction to \(0.0898\unit{\newton}\). These results indicate that the expert network is capable of generating high-quality data that is instrumental in training robust predictive models. Such volume and quality of training data would be difficult, if not infeasible, to obtain from human operators in real-world settings.

\section{Conclusion and Discussion}

The work introduces the \gls{expert navigation network} and investigates its application to downstream tasks in autonomous catheterization. Although no new learning algorithms are proposed, the approach demonstrates how the capabilities of the \gls{expert navigation network} can be leveraged to improve performance in tasks such as \gls{imitation learning} and force prediction. Through the integration of multiple input modalities and the generation of expert trajectories, the \gls{expert navigation network} exhibits strong potential for addressing the complex challenges associated with endovascular navigation.

Drawing inspiration from advancements in related domains such as autonomous driving~\autocite{dosovitskiy17}, this research is intended to encourage greater engagement from the \gls{machine learning} community with the emerging yet critical domain of autonomous catheterization. The results highlight the promise of machine learning frameworks in enhancing patient safety, precision, and procedural efficiency in endovascular interventions. The multimodal approach demonstrated in this chapter sets the stage for addressing data scarcity challenges in endovascular imaging, which will be the focus of the next chapter through the introduction of the Guide3D dataset.

\onehalfspacing
\chapter{Stereo X-Ray-Based 3D Reconstruction}
\chaptermark{Stereo X-Ray-Based 3D Reconstruction}
\glsresetall

\begin{cabstract}
	Endovascular navigation is highly dependent on fluoroscopic imaging due to the limited availability of sensory feedback. However, traditional methods for 3D reconstruction often rely on specialized hardware or require prolonged exposure to ionizing radiation. To address these limitations, a novel deep learning framework is proposed for 3D guidewire reconstruction, which utilizes a spherical coordinate representation for efficient modelling and a \acrfull{gru}-based architecture for accurate shape prediction. Extensive evaluations demonstrate strong predictive performance, including a \acrfull{maxed} of $6.88 \pm 5.23$ and a \acrfull{mete} of $3.28 \pm 2.59$, even under challenging anatomical conditions. To support reproducibility and accelerate progress in the field, a new benchmark dataset, \textit{Guide3D}, is introduced. This dataset comprises publicly available bi-planar X-ray images with annotated fluoroscopic video sequences for guidewire shape prediction. The proposed approach enhances both the accuracy and efficiency of guidewire 3D reconstruction and contributes to the broader research community by providing an open-access resource for future innovation.
\end{cabstract}

This chapter presents the work from the following publications:

\begin{itemize}
	\item \fullcite{jianu2024guide3d}
	\item \fullcite{jianu2024deepwire}
	\item \fullcite{jianu20233d}
\end{itemize}

\section{Introduction}

Minimally invasive surgery has transformed endovascular intervention by offering less intrusive alternatives that promote accelerated recovery~\autocite{puschel2022robot}. These advancements rely significantly on precise instrument navigation, which is typically achieved through imaging technologies, particularly 2D visualization methods such as \textit{monoplanar fluoroscopy}. This imaging modality is widely adopted due to its minimal disruption to surgical workflows and its cost-effectiveness~\autocite{rafii2014current}. However, its inherent reliance on two-dimensional projections introduces critical limitations, most notably the absence of depth perception. This deficit complicates the spatial localization of surgical instruments and increases the likelihood of unintentional contact with fragile arterial structures, potentially resulting in arterial damage, clot formation, or decreased procedural accuracy~\autocite{hoffmann2013reconstruction, hoffmann2015electrophysiology}. Furthermore, 2D imaging techniques are often constrained by limited resolution and an inability to provide real-time 3D guidance, thereby complicating the execution of complex interventions. Overcoming these imaging constraints is imperative for improving patient safety, minimizing procedural risks, and enhancing the success rate of minimally invasive techniques.

To address the lack of depth perception, multi-view imaging systems—particularly biplanar scanners—have been introduced in endovascular procedures. These systems enhance spatial localization by integrating multiple angular perspectives~\autocite{burgner2011toward,wagner20164d,hoffmann2015electrophysiology,ambrosini2017fully}, often employing epipolar geometry-based methods to reconstruct accurate 3D representations~\autocite{baur2016automatic}. Despite their advantages, two key challenges limit the widespread adoption and utility of these systems~\autocite{ramadani2022survey}: \textit{i)} the difficulty of accurate image segmentation due to low contrast, overlapping anatomical structures, and motion artefacts, which undermine the reliability of shape reconstruction algorithms; and \textit{ii)} the dependence on specialized biplanar scanners, which are cost-prohibitive and not readily available in many clinical environments~\autocite{burgner2011toward}. Addressing these challenges necessitates the development of robust datasets encompassing a wide range of anatomical and imaging scenarios. Such resources would support the training and validation of segmentation and reconstruction algorithms, ultimately enabling more accessible and reliable imaging solutions across diverse clinical contexts.

In this chapter, a series of interconnected contributions to guidewire reconstruction in endovascular intervention are introduced, spanning simulation-based experimentation, real-world data application, and dataset development. First, we apply reconstruction techniques to real-life data using a manually annotated dataset and introduce a spherical coordinate representation enhanced by \glspl{recurrent neural network}. Secondly, the Guide3D dataset is presented as a large-scale, publicly available resource informed by insights from the prior contributions. Guide3D facilitates scalable 3D reconstruction, offering a foundation for advancing guidewire navigation in both research and clinical applications.

\section{Related Works}

\paragraph{Endovascular Datasets.} Datasets serve a pivotal role in advancing endovascular navigation by providing the foundational resources for algorithm development, evaluation, and benchmarking. Existing datasets have been derived from multiple imaging modalities, including mono X-ray~\autocite{barbu2007hierarchical,ambrosini2017fully,yi2020automatic,nguyen2020end}, 3D echocardiography~\autocite{wu2014fast}, and 3D MRI~\autocite{mastmeyer2017model}. These datasets comprise both real and synthetic imagery; however, their limited availability for critical tasks such as tool segmentation and 3D reconstruction continues to pose a significant bottleneck~\autocite{barbu2007hierarchical,ambrosini2017fully,mastmeyer2017model,nguyen2020end,danilov2023use}. Mono X-ray datasets remain dominant within the context of endovascular interventions (see Table~\ref{ch5tab:dataset-comparison}), yet they often lack the spatial fidelity required for high-precision 3D reconstruction. This shortfall underscores the need for comprehensive and standardized datasets specifically designed for guidewire and catheter-related applications.

\paragraph{Catheter and Guidewire Segmentation.} The segmentation of endovascular instruments, such as guidewires and catheters, is a prerequisite for accurate navigation and procedural success. However, progress in this domain is tightly coupled with the availability and granularity of annotated datasets. Due to the scarcity of real-world annotated data, synthetic and semisynthetic datasets have been increasingly adopted, leveraging modalities such as 2D X-ray and 3D MRI~\autocite{barbu2007hierarchical,ambrosini2017fully,mastmeyer2017model}. Synthetic datasets have been shown to improve training efficiency and model generalizability~\autocite{nguyen2020end,danilov2023use}. Deep learning-based approaches, particularly those employing U-Net architectures~\autocite{ronneberger2015u}, have significantly advanced segmentation accuracy, enabling fully automated pipelines. Furthermore, optical flow-based unsupervised methods have contributed to streamlined tracking of instruments during interventions. Nevertheless, the continued absence of a publicly available, standardized dataset for catheter and guidewire segmentation remains a major limitation, hindering both reproducibility and clinical translation of state-of-the-art methods.

\clearpage
\begin{landscape}
	\begin{table}[htbp]
		\centering
		\caption{Endovascular intervention datasets comparison.}\label{ch5tab:dataset-comparison}
		\begin{tabular}{l l l r l l l c}
			\toprule
			{\thead{Dataset}}                        & {\thead{Data Collection}} & {\thead{Data                                                                                                       \\ Type}}  & {\thead{\#Frames}} & {\thead{Data \\ Source}} & {\thead{Annotation}} & {\thead{Public}} & {\thead{Task}} \\
			\midrule
			\textcite{barbu2007hierarchical}         & Mono X-ray                & Video        & 535                & Real                  & Manual    & No  & \multirowcell{10}{Segmentation}      \\
			\cmidrule{1-7}
			\textcite{wu2014fast}                    & 3D Echo                   & Video        & 800                & Real                  & Manual    & No                                         \\
			\cmidrule{1-7}
			\textcite{ambrosini2017fully}            & Mono X-ray                & Image        & 948                & Real                  & Manual    & No                                         \\
			\cmidrule{1-7}
			\textcite{mastmeyer2017model}            & 3D MRI                    & Image        & 101                & Real                  & Manual    & No                                         \\
			\cmidrule{1-7}
			\textcite{yi2020automatic}               & Mono X-ray                & Image        & 2,540              & Synthesis             & Automatic & No                                         \\
			\cmidrule{1-7}
			\textcite{nguyen2020end}                 & Mono X-ray                & Image        & 25,271             & Phantom               & Semi-Auto & No                                         \\
			\cmidrule{1-7}
			\textcite{danilov2023use}                & 3D Ultrasound             & Video        & 225                & Synthetic             & Manual    & No                                         \\

			\cmidrule{1-8}

			\textcite{wagner20164d}                  & Mono X-ray                & Video        & ---                & \makecell[l]{Phantom} & Automatic & No  & \multirowcell{10}{3D Reconstruction} \\
			\cmidrule{1-7}
			\textcite{hoffmann2015electrophysiology} & Mono X-ray                & Image        & 176                & \makecell[l]{Phantom} & Semi-Auto & No                                         \\
			\cmidrule{1-7}
			\textcite{delmas2015three}               & Mono X-ray                & Image        & \makecell[r]{2,289                                                                                  \\ 5 \\ 63} & \makecell[l]{Simulated \\ Phantom \\ Clinical} & Automatic & No \\
			\cmidrule{1-7}
			\textcite{baur2016automatic}             & Bi-planar X-ray           & Image        & 70                 & Canine                & Manual    & No                                         \\
			\cmidrule{1-7}
			\textcite{brost2010catheter}             & Mono X-ray                & Image        & 938                & Clinical              & Semi-Auto & No                                         \\
			\cmidrule{1-7}
			\textcite{ma2010real}                    & Mono X-ray, CT            & Image        & 1,048              & Clinical              & Manual    & No                                         \\
			\cmidrule{1-7}
			\textcite{hoffmann2012semi}              & Bi-planar X-Ray           & Video        & 33                 & Clinical              & Semi-Auto & No                                         \\
			\cmidrule{1-8}
			\bfseries Guide3D (ours)                 & Bi-planar X-ray           & Video        & 8,746              & Phantom               & Manual    & Yes & \makecell{ Segmentation              \\ 3D Reconstruction} \\
			\bottomrule
		\end{tabular}
	\end{table}
\end{landscape}
\clearpage

\paragraph{3D Reconstruction in Endovascular Procedures.} Accurate 3D reconstruction plays a critical role in endovascular interventions by facilitating advanced visualization, precise instrument tracking, and improved clinical outcomes. Historically, biplane fluoroscopic imaging systems have served as the cornerstone for 3D shape reconstruction, employing triangulation and epipolar geometry principles. Early work by \textcite{burgner2011toward} introduced projective-invariant triangulation for shape estimation, which was later refined through elastic grid registration and graph-search algorithms by \textcite{wagner20164d}, \textcite{hoffmann2015electrophysiology}, \textcite{hoffmann2013reconstruction}, and \textcite{hoffmann2012semi}. The clinical relevance of 3D vascular modelling has also been demonstrated in both biplanar and single-view contexts by \textcite{delmas2015three} and \textcite{petkovic2014real}.

Despite their precision, biplane systems see limited clinical use due to high costs, increased radiation, and workflow disruptions. This has driven interest in monoplane imaging as a more accessible alternative. Methods like those of \textcite{lobaton2013continuous} and \textcite{vandini20133d} combine fluoroscopic views with kinematic priors for shape reconstruction, but often require C-arm repositioning, interrupting procedures. To mitigate this, \textcite{vandini2015vision} proposed an automated robotic approach fusing kinematics with fluoroscopy, avoiding C-arm adjustments. Other works, such as \textcite{otake2014piecewise} and \textcite{papalazarou20123d}, explore piecewise-rigid 2D/3D registration and non-rigid SfM. Yet, monoplane systems still rely on multiple closely-spaced views, raising computational demands and radiation exposure.

\paragraph{Challenges and Emerging Approaches.} Despite continued advancements in biplane and monoplane reconstruction methods, several persistent challenges remain, including elevated system costs, workflow disruptions, and high computational complexity. These limitations underscore the promise of deep learning techniques, which have demonstrated the ability to infer guidewire and catheter shapes directly from monoplane images with high accuracy. Unlike traditional methods, deep learning approaches eliminate the need for dual-view setups and reduce reliance on manual intervention. However, such methods demand access to large, diverse datasets to ensure robust training and generalizability.

To address these gaps, there is a pressing need for the development of specialized datasets that support both 3D reconstruction and tool segmentation at scale. These datasets must account for anatomical and imaging variability encountered in real-world clinical environments and provide a reliable foundation for benchmarking algorithmic progress. Establishing such resources is essential for advancing the field toward safer and broadly accessible endovascular procedures.

\section{The Guide3D Dataset}

\begin{figure}[h]
	\centering
	\begin{minipage}{0.70\textwidth}
		\subfloat[]{\includegraphics[width=\linewidth]{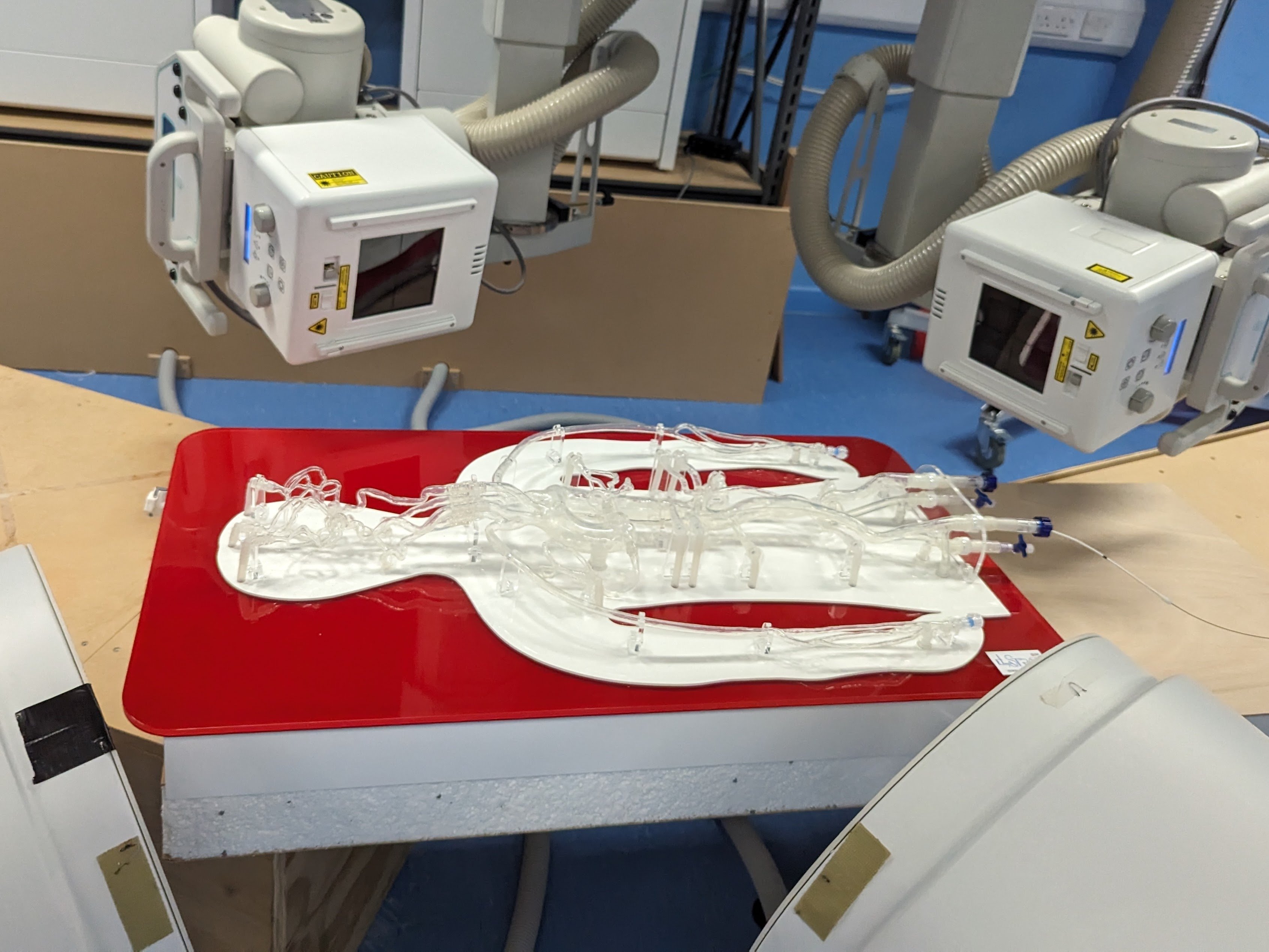}\label{guide3d-fig:setup}} \\
	\end{minipage}\hfill%
	\begin{minipage}{0.24\textwidth}
		\subfloat[]{\includegraphics[width=.99\linewidth]{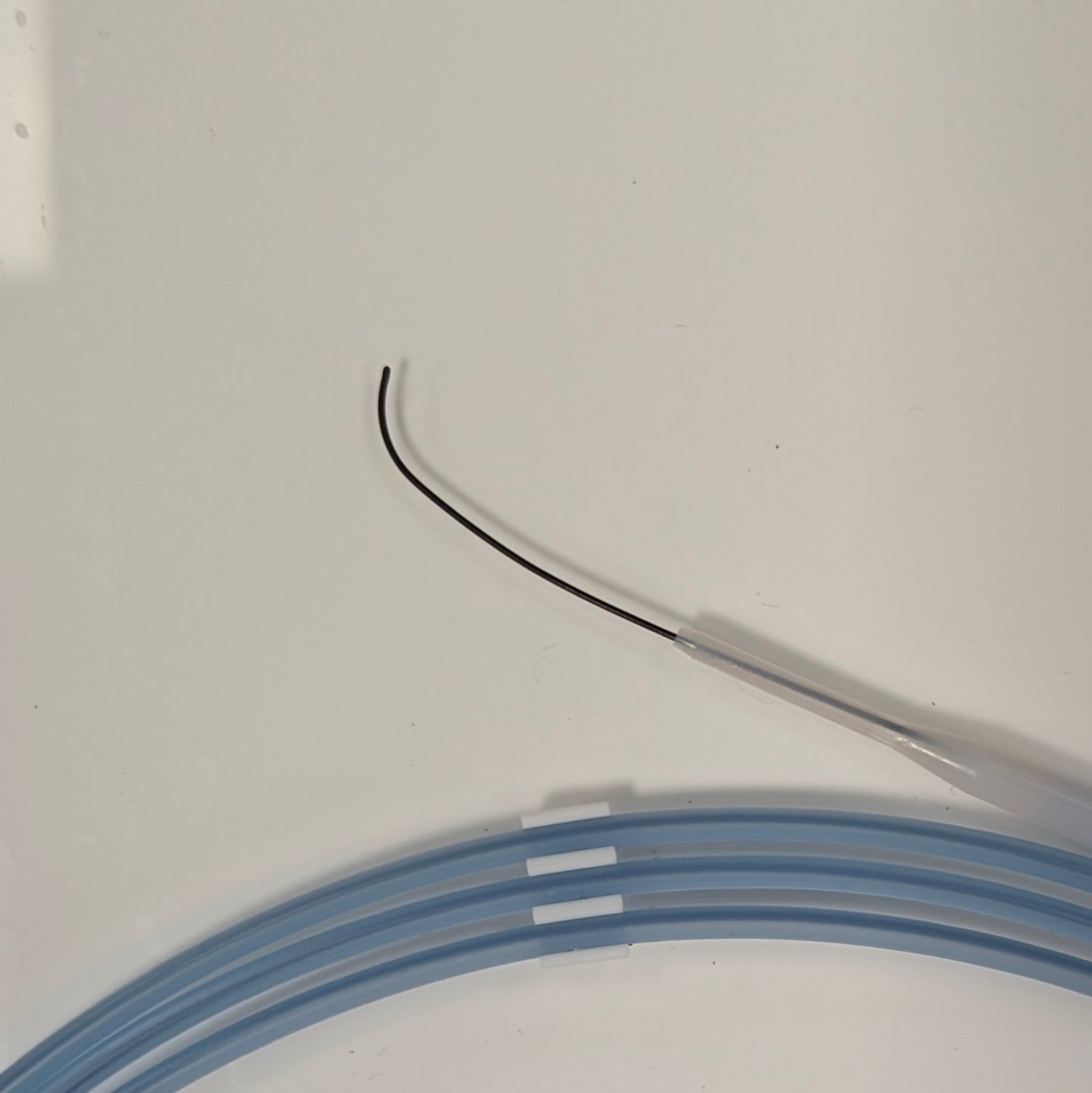}\label{guide3d-fig:guidewire-nitrex}}  \\
		\subfloat[]{\includegraphics[width=.99\linewidth]{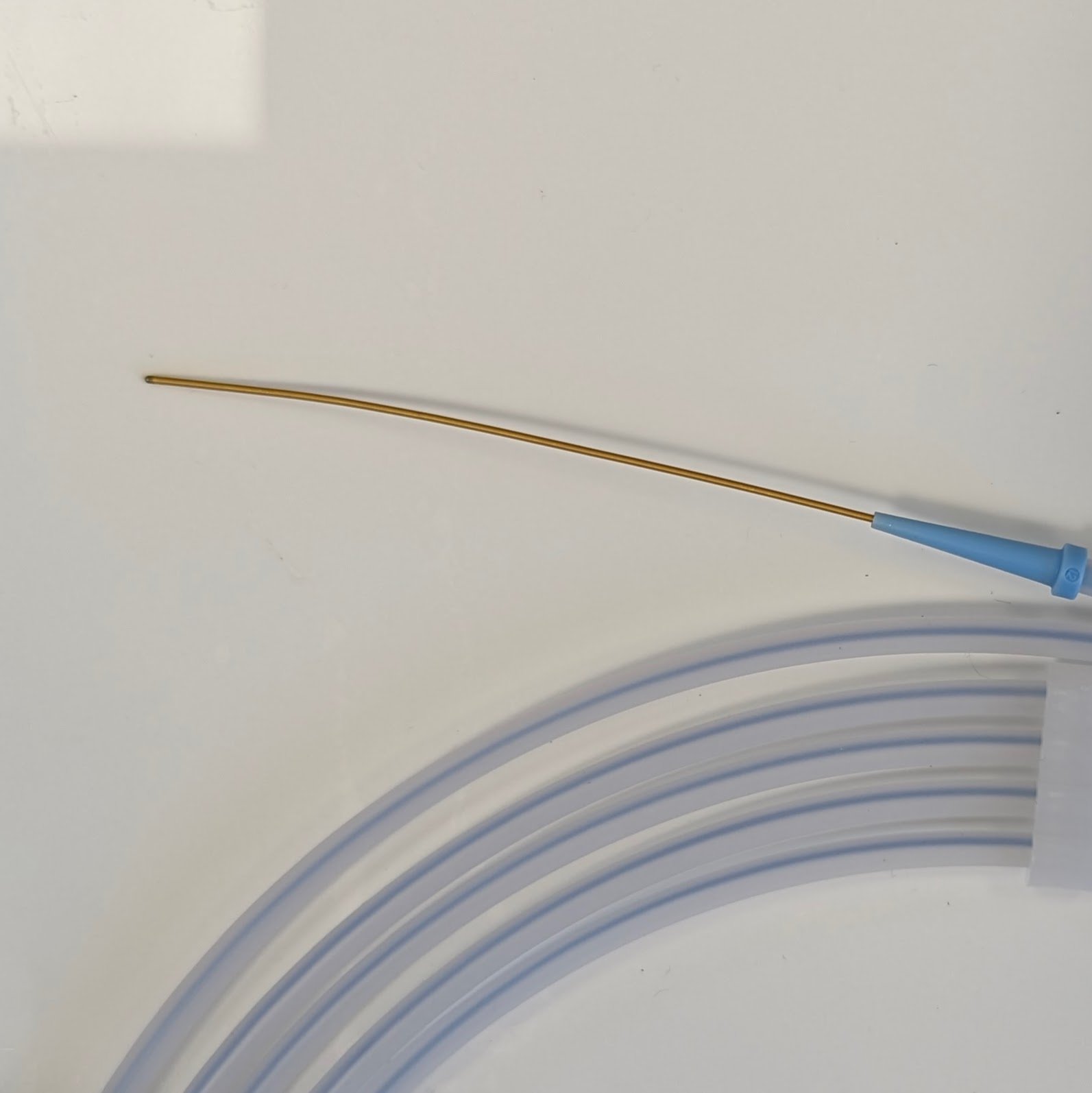}\label{guide3d-fig:guidewire-radifocus}}
	\end{minipage}
	\mycaption{Dataset Composition Materials}{\textit{a)} Overall setup \& endovascular phantom, \textit{b)} Radifocus (angled) guidewire. and \textit{c)} Nitrex (straight) guidewire.}
\end{figure}

\subsection{Data Collection Setup}\label{ch5-subsec:materials}

\paragraph{X-ray System.} The setup utilized a Bi-planar X-ray system (Fig.~\ref{guide3d-fig:setup}) with \(60\unit{\kilo\watt}\) Epsilon X-ray Generators from EMD Technologies Ltd. and \(16\)-inch Image Intensifier Tubes by Thales, featuring dual focal spot Variant X-ray tubes for high-definition imaging. The setup included Ralco Automatic Collimators for precise alignment and exposure, with calibration achieved through acrylic mirrors and geometric alignment grids.

\paragraph{Anatomical Models.} The study utilized the half-body vascular phantom model from Elastrat Sarl Ltd., Switzerland (Fig.~\ref{guide3d-fig:setup}). This model is enclosed in a transparent box and integrated into a closed water circuit to simulate blood flow. Constructed from soft silicone and incorporating compact continuous flow pumps with a slippery liquid, it replicates human blood flow dynamics. The design is based on detailed postmortem vascular casts, ensuring anatomical accuracy reflective of human vasculature as documented in seminal works~\autocite{martin1998vitro,gailloud1999vitro}. The utilization of this anatomically accurate model facilitated the realistic simulation of vascular scenarios.

\paragraph{Surgical Tools.} The dataset was enhanced by navigating complex vascular structures using two distinct types of guidewires that are widely used in real-world endovascular surgery. The first, the Radifocus™ Guide Wire M Stiff Type (Terumo Ltd.) (Fig.~\ref{guide3d-fig:guidewire-radifocus}), is manufactured from nitinol and coated in polyurethane-tungsten. It features a diameter of \(0.89\unit{\milli\meter}\), a length of \(260\unit{\centi\meter}\), and a \(3\unit{\centi\meter}\) angled tip. This guidewire is commonly used for seeking, dissecting, and crossing lesions. The second, the Nitrex Guidewire (Nitrex Metal Inc.) (Fig.~\ref{guide3d-fig:guidewire-nitrex}), is also composed of nitinol and offers a gold-tungsten straight tip. It measures \(0.89\unit{\milli\meter}\) in diameter and approximately \(300\unit{\centi\meter}\) in length, with a \(15\unit{\centi\meter}\) straight tip, making it suitable for navigating tortuous vascular paths.

\begin{table}[t]
	\caption{Dataset Composition Overview.}\label{guide3d-tab:dataset-stats}
	\centering
	\setlength{\tabcolsep}{12pt}
	\scriptsize
	\begin{tabular}{c S[table-format=4] S[table-format=4] S[table-format=4]}
		\toprule
		\thead{Sample Type} & {\thead{Radifocus™ Guide Wire               \\ (Angled)}} & {\thead{Nitrex Guidewire \\ (Straight)}} & {\thead{Total}} \\
		\midrule
		w fluid             & 3664                          & 484  & 4148 \\
		w\/o fluid          & 2472                          & 2126 & 4598 \\
		\cmidrule{1-4}
		Total               & 6136                          & 2610 & 8746 \\
		\bottomrule
	\end{tabular}
\end{table}

\subsection{Data Acquisition, Labelling, and Statistics}

Utilizing the materials described in Subsection~\ref{ch5-subsec:materials}, a dataset of 8,746 high-resolution samples ($\num{1024} \times \num{1024}$ pixels) was compiled. This dataset includes \num{4373} paired instances with and without a simulated blood flow medium. Specifically, it comprises \num{6136} samples from the Radifocus Guidewire and \num{2610} from the Nitrex guidewire, establishing a robust foundation for automated guidewire tracking in bi-planar scanner images. Manual annotation was performed using the Computer Vision Annotation Tool (CVAT)~\cite{cvat2023}, where polylines were created to accurately track the dynamic path of the guidewire. The decision to represent the guidewire as a polyline was due to the inherent structure of a guidewire, where certain parts would inevitably overlap, making a segmentation mask an unsuitable representation. In contrast, a polyline can effectively represent a looping guidewire, providing better accuracy. As detailed in Table~\ref{guide3d-tab:dataset-stats}, the dataset includes 3,664 instances of angled guidewires with fluid and \num{484} without, while straight guidewires are represented by \num{2472} instances with fluid and 2,126 without. This distribution illustrates a variety of procedural contexts. It is noted that all \num{8746} images in the dataset are equipped with \textit{manual segmentation ground truth}, hence supporting the development of algorithms that need segmentation maps as the reference.

\subsection{Camera Calibration and Distortion Correction}~\label{guide3d-subsec:calibration}

\begin{figure}[t]
	\centering
	\subfloat[Undistortion]{\includegraphics[width=.48\textwidth]{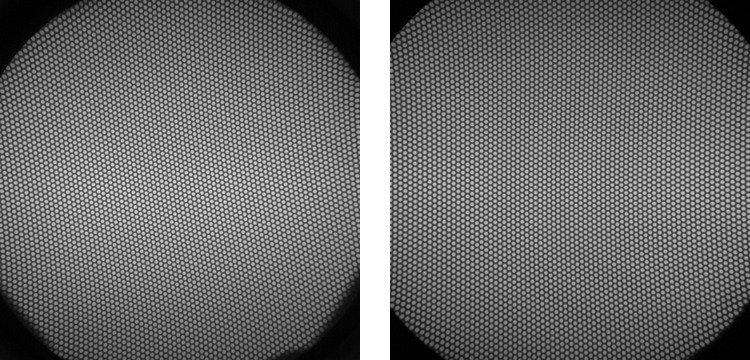}}\hfill
	\subfloat[Marker-based calibration]{\includegraphics[width=.48\textwidth]{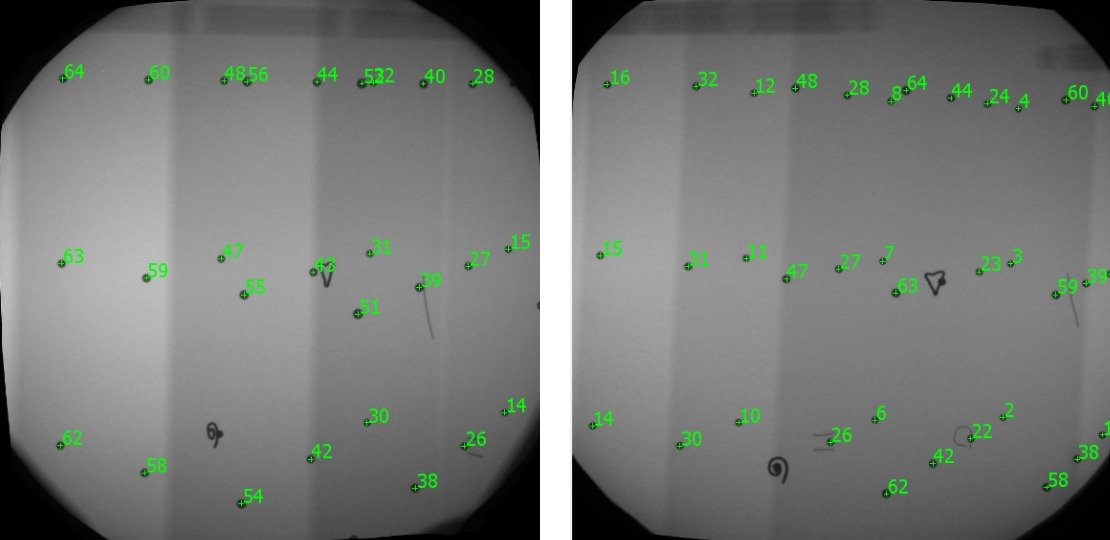}}
	\mycaption{Undistortion and Calibration Procedure}{\textit{a)} Application of the hexagonal undistortion grid to correct lens distortion, and \textit{b)} Identification of calibration markers on the perforated steel sheet for projection matrix estimation.}\label{guide3d-fig:calibration}
\end{figure}

The camera parameters were extracted by following the traditional undistortion and calibration method. Initially, undistortion was achieved using a Local Weighted Mean (LWM) algorithm, applying a perforated steel sheet with a hexagonal pattern (from McMaster-Carr Ltd., part number 9255T645) as the framing reference, and employing a blob detection algorithm for precise identification of distortion points~\autocite{brainerd2010x}. This setup established correspondences between distorted and undistorted positions, enabling accurate distortion correction as per \textcite{verdonck1999variations}. A subsequent calibration step was undertaken utilizing a semi-automatic approach for marker identification and a Random Sampling Consensus (RANSAC) method to ensure robustness in computing the projection matrix and deriving intrinsic and extrinsic camera parameters. The calibration process was refined through Direct Linear Transformation (DLT) and non-linear optimization, considering multiple poses of the calibration object to optimize the entire camera setup~\autocite{zhang2000flexible}. Figure~\ref{guide3d-fig:calibration} illustrates two key stages of the calibration procedure. Panel (a) shows the application of the hexagonal undistortion grid used to correct lens distortion. Panel (b) depicts the subsequent step of identifying calibration markers on the reference sheet, which enables the computation of projection matrices.

\subsection{Guidewire Reconstruction}

Given polyline representations of a curve in both planes, the reconstruction process is initiated by parameterizing these curves using B-Spline interpolation. This is achieved by expressing each curve as a function of the cumulative distance along the curve. Specifically, let \( \mathbf{C}_A(\mathbf{u}_A) \) and \( \mathbf{C}_B(\mathbf{u}_B) \) be the parameterized B-Spline curves in their respective planes, where \( \mathbf{u}_A \) and \( \mathbf{u}_B \) are the normalized arc-length parameters. Next, the corresponding \( \mathbf{u}_B \) for a given \( \mathbf{u}_A \) is found using epipolar geometry. Once the corresponding points \( \mathbf{C}_A(\mathbf{u}_A^i) \) and \( \mathbf{C}_B(\mathbf{u}_B^i) \) are identified, the 3D coordinates \( \mathbf{P}^i \) of these points are computed by triangulation. This results in a set of 3D points \( \{\mathbf{P}^i\}_{i=1}^{M} \), where \( M \) is the total number of sampled points, effectively reconstructing the original curve in 3D space.

\subsubsection{Retrieving the Fundamental Matrix \( \mathbf{F} \).}

The relationship between the points in Image A (\( \mathbf{I}_A \)) and Image B (\( \mathbf{I}_B \)) is described by the fundamental matrix \( \mathbf{F} \), which satisfies the condition \( \mathbf{x}_B^T \mathbf{F} \mathbf{x}_A = 0 \) for corresponding points \( \mathbf{x}_A \) in \( \mathbf{I}_A \) and \( \mathbf{x}_B \) in \( \mathbf{I}_B \). Given the calibration steps undertaken in Subsection \ref{guide3d-subsec:calibration}, the projection matrices \( \mathbf{P}_A \) and \( \mathbf{P}_B \) are now available. From these projection matrices, the fundamental matrix can be derived as follows:

\begin{equation}
	\mathbf{F} = [\mathbf{e}_B]_\times \mathbf{P}_B \mathbf{P}_A^+ ,
\end{equation}
where \( \mathbf{e}_B \) is the epipole in Image B, defined as \( \mathbf{e}_B = \mathbf{P}_B \mathbf{C}_A \), with \( \mathbf{C}_A \) being the camera centre of \( \mathbf{P}_A \). The notation \( [\mathbf{e}_B]_\times \) represents the skew-symmetric matrix of the epipole \( \mathbf{e}_B \), which is given by:

\begin{equation}
	[\mathbf{e}_B]_\times = \begin{bmatrix}
		0       & -e_{B3} & e_{B2}  \\
		e_{B3}  & 0       & -e_{B1} \\
		-e_{B2} & e_{B1}  & 0
	\end{bmatrix}
\end{equation}

Here, \( \mathbf{e}_B = (e_{B1}, e_{B2}, e_{B3})^T \). The matrix \( \mathbf{P}_A^+ \) is the pseudoinverse of the projection matrix \( \mathbf{P}_A \). The fundamental matrix \( \mathbf{F} \) encapsulates the epipolar geometry between the two views, ensuring that corresponding points \( \mathbf{x}_A \) and \( \mathbf{x}_B \) lie on their respective epipolar lines.

\subsubsection{Epipolar Matching of Curve Representations}

The matching phase begins with uniformly sampling points along the curve \(\mathbf{C}_A(u_A)\) at intervals \(\Delta u_A\). For each sampled point \(x_A = \mathbf{C}_A(u_A)\), the epiline \(l_B = F x_A\) is projected into Image B. The intersection of the epiline \(l_B\) with the curve \(\mathbf{C}_B(u_B)\) is then determined, thereby obtaining the corresponding parameter \(u_B\) for each \(u_A\).

\begin{figure}[t]
	\centering
	\includegraphics[width=\linewidth]{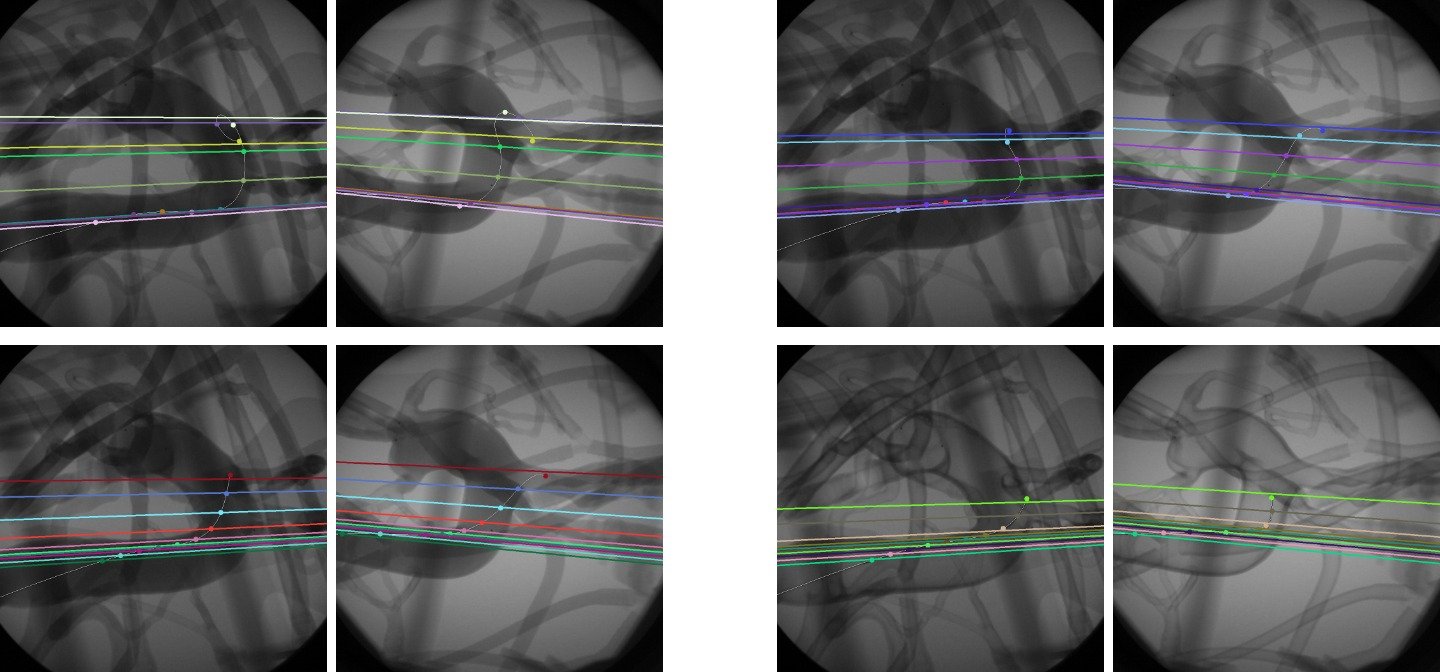}
	\mycaption{Point Matching Process}{ Sampled points from image \(I_A\) (\(C_A(u_A)\)) and their corresponding epilines \(l_A\) on image \(I_B\) with their matches \(C_B(u_B)\). The epilines for \(C_B(u_B)\) are then computed and displayed on the image \(I_A\).}\label{guide3d-fig:matching}
\end{figure}

Due to errors in the projection matrices \(P_A\) and \(P_B\), there are instances where the epiline \(l_B\) does not intersect with any part of the curve \(\mathbf{C}_B\). To address this, similar to the approach in \autocite{altingovde20223d}, a monotonic function \(f_A(u_A) \rightarrow u_B\) is fitted using a Piecewise Cubic Hermite Interpolating Polynomial (PCHIP), thus interpolating the missing intersections. The matching can be visualized in Fig.~\ref{guide3d-fig:matching}.

\subsubsection{Triangulation for 3D Point Recovery}

Given the projections from both the first and second cameras, a system of linear equations can be constructed as:

\begin{equation}
	\begin{aligned}
		\mathbf{x}_1 \times (\mathbf{P}_1 \mathbf{X}) = 0 \\
		\mathbf{x}_2 \times (\mathbf{P}_2 \mathbf{X}) = 0
	\end{aligned}
\end{equation}

Therefore, a system of linear equations of the form \( \mathbf{A} \mathbf{X} = 0 \) is formed, wherein \( \mathbf{A} \) is a \( 4 \times 4 \) matrix composed of rows from the projection matrices \( \mathbf{P}_1 \) and \( \mathbf{P}_2 \) and the image coordinates \( \mathbf{x}_A \) and \( \mathbf{x}_B \). By employing the Singular Value Decomposition (SVD) \autocite{klema1980singular} on \( \mathbf{A} \), the desired \( \mathbf{X} \) minimizing \( ||\mathbf{A} \mathbf{X}|| \) subject to \( ||\mathbf{X}|| = 1 \) is determined. The solution corresponds to the eigenvector associated with \( \mathbf{A} \)'s smallest singular value, thereby deriving the homogeneous coordinates for the triangulated 3D point. Finally, all reprojections were verified to ensure geometric plausibility. Reconstructions with an average reprojection error greater than \(\num{25}\) pixels (approximately \(2.5\%\) of the \(1024 \times 1024\) image resolution) were considered inaccurate and discarded.

\subsection{Utility of Gude3D Dataset for the Research Community}

Guide3D advances endovascular imaging by offering a bi-planar fluoroscopic dataset for \textit{segmentation} and \textit{3D reconstruction}, serving as an open-source benchmark. It supports precise algorithm comparisons for segmentation \autocite{ambrosini2017fully,subramanian2019automated,nguyen2020end,zhou2020real} and facilitates method development in 3D reconstruction with its use of bi-planar imagery \autocite{ambrosini20153d,baur2016automatic,brost2010catheter,ma2010real}. With video data, Guide3D enables \textit{video}-based methods, leveraging temporal dimensions for dynamic analysis, which enriches the segmentation and reconstruction capabilities whilst also adhering to the nature of the procedure. This versatility underscores Guide3D's crucial role in advancing endovascular imaging~\autocite{ramadani2022survey}.

\section{Guidewire Shape Prediction}\label{ch5-sec:3D-shape-prediction}

Utilizing the Guide3D dataset, this section builds a benchmark for the shape prediction task. Accurate shape prediction of the guidewire is essential for successful navigation and intervention. In particular, a novel shape prediction network is presented, aimed at predicting the guidewire shape from a sequence of monoplanar images. The motivation behind this approach lies in the potential of deep learning to learn spatio-temporal correlations from a static camera with a dynamic scene \autocite{sitzmann2019deepvoxels}. Unlike the conventional reconstruction methods requiring biplanar images \autocite{burgner2011toward}, the network leverages a sequence of images to provide temporal information, enabling the network to learn a mapping from a single image $\mathbf{I}_A$ to the 3D guidewire curve $\mathbf{C}(\mathbf{u})$. By adopting deep learning techniques, the aim is to simplify the shape prediction process while maintaining high accuracy. This method holds promise for enhancing endovascular navigation by providing real-time, accurate predictions of the guidewire shape, ultimately improving procedural outcomes and reducing reliance on specialized equipment.

\paragraph{Task formulation.}
Given a sequence of single-plane images $\{\mathbf{I}_{A,t}\}_{t=1}^{T}$, the goal is to predict a 3D curve $\mathbf{C}(u)$ parameterized by arclength. For supervision, the curve is uniformly discretized at step $\Delta u$ into $M$ points $\{u_j\}_{j=1}^{M}$. The network outputs (i) a tip position $\mathbf{p}\in\mathbb{R}^3$, (ii) per-point angular offsets $\{(\Delta\phi_j,\Delta\theta_j)\}_{j=1}^{M}$ under a fixed radius $r$, and (iii) a stop distribution $\mathbf{S}\in[0,1]^M$ over the $M$ positions indicating the terminal index. At inference, the 3D curve is reconstructed by integrating the angular offsets from the tip and truncating at $\arg\max_j S_j$.

\subsection{Spherical Coordinates Representation}

\begin{figure}[t]
	\centering
	\subfloat[Spherical Representation]{%
		\makebox[0.30\textwidth]{%
			\includegraphics[width=0.25\textwidth]{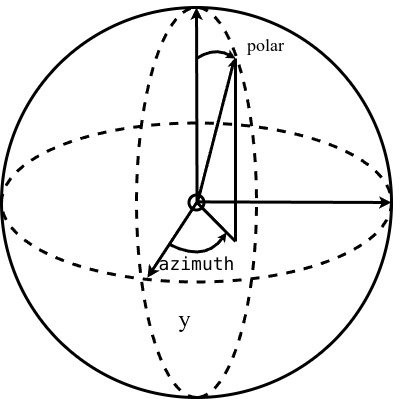}%
		}%
		\label{ch5-fig:spherical-representation}%
	}\hfill
	\subfloat[Network Architecture]{%
		\includegraphics[width=0.68\textwidth]{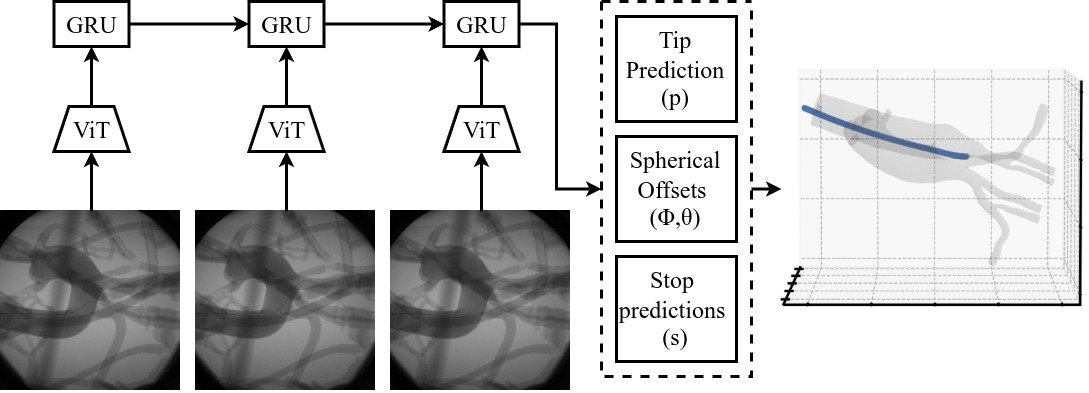}%
		\label{ch5-fig:network-architecture}%
	}
	\mycaption{Network Key Components}{The figure illustrates essential components of the proposed model. \textit{a)} Spherical coordinates $(r, \theta, \phi)$ used for predicting the guidewire shape. \textit{b)} The model predicts the 3D shape of a guidewire from image sequences $\mathbf{I}_t$. A \gls{vit} extracts spatial features $\mathbf{z}_t$, which a \gls{gru} processes to capture temporal dependencies, producing hidden states $\mathbf{h}_t$. The final hidden state drives three prediction heads: the Tip Prediction Head for 3D tip position $\mathbf{p} \in \mathbb{R}^3$, the Spherical Offset Prediction Head for coordinate offsets $(\Delta\phi, \Delta\theta)$, and the Stop Prediction Head for terminal point probability $\mathbf{S}$.}
\end{figure}

Predicting 3D points directly can be challenging due to the high degree of freedom. To mitigate this, spherical coordinates are used, which offer significant advantages over Cartesian coordinates for guidewire shape prediction. Spherical coordinates, as represented in Fig.~\ref{ch5-fig:spherical-representation}, defined by the radius $ r $, polar angle $ \theta $, and azimuthal angle $ \phi $, provide a more natural representation for the position and orientation of points along the guidewire, which is typically elongated and curved. Mathematically, a point in spherical coordinates $(r, \theta, \phi)$ can be converted to Cartesian coordinates $(x, y, z)$ using the transformations $x = r \sin \theta \cos \phi$, $y = r \sin \theta \sin \phi$, and $z = r \cos \theta$. This conversion simplifies the modelling of angular displacements and rotations, as spherical coordinates directly encode directional information.

Predicting angular displacements $(\Delta\theta, \Delta\phi)$ relative to a known radius $r$ aligns with the physical constraints of the guidewire, facilitating more accurate and interpretable shape predictions. By predicting an initial point (tip position) and representing subsequent points as offsets in $\Delta \phi$ and $\Delta \theta$ while keeping $r$ fixed, this method simplifies shape comparison and reduces the parameter space. This approach enhances the model's ability to capture the guidewire's spatial configuration and improves overall prediction performance.

\subsection{Network Architecture}

The proposed model (shown in Fig.~\ref{ch5-fig:network-architecture}) addresses the problem of predicting the 3D shape of a guidewire from a sequence of images. Each image sequence captures the guidewire from different time steps ($\mathbf{I}_{A,t}$), and the goal is to infer the continuous 3D shape $\mathbf{C}_t(\mathbf{u}_t)$. This many-to-one prediction task is akin to generating a variable-length sequence from variable-length input sequences, a technique commonly utilized in fields such as machine translation \autocite{sutskever2014sequence} and video analysis.

To achieve this, the input pipeline consists of a sequence of images depicting the guidewire. A \gls{vit} \autocite{dosovitskiy2020image}, pre-trained on ImageNet, is employed to extract high-dimensional spatial feature representations from these images. The output embeddings of the \gls{vit} are subsequently passed through a \gls{mlp}, producing the feature maps $\mathbf{z}_t \in \mathbb{R}^{512}$. These feature maps are then fed into a \gls{gru} to capture the temporal dependencies across the image sequence. The \gls{gru} processes the feature maps $\mathbf{z}_t$ from consecutive time steps, producing a sequence of hidden states $\mathbf{h}_t$. Formally, the \gls{gru} operation at time step $t$ is defined as:
\begin{equation}
	\mathbf{h}_t = \text{GRU}(\mathbf{z}_t, \mathbf{h}_{t-1}),
\end{equation}
where $\mathbf{h}_t \in \mathbb{R}^{H}$ denotes the final hidden state dimension. \textit{Module details:} the ViT-to-$\mathbf{z}_t$ projection is a lightweight MLP (single hidden layer, GELU, dropout $p{=}0.1$) to 512 dimensions; the GRU uses a single recurrent layer with hidden size $H$ (512 in our implementation) and dropout $p{=}0.1$ between time steps.

The final hidden state $\mathbf{h}_t$ from the \gls{gru} is used by three distinct prediction heads. Each head is implemented as a lightweight \gls{mlp} with a single hidden layer (width 256), GELU activation, and dropout $p=0.1$:

\begin{itemize}
	\item \textbf{Tip Prediction Head:} an \gls{mlp} maps $\mathbf{h}_t$ to a 3D Cartesian anchoring point.
	\item \textbf{Spherical Offset Prediction Head:} an \gls{mlp} outputs angular offsets $(\Delta\phi, \Delta\theta)$ for a discretized curve.
	\item \textbf{Stop Prediction Head:} an \gls{mlp} outputs logits over the same discretization; a softmax is applied to obtain the terminal-point probability vector $\mathbf{S}$.
\end{itemize}

\noindent\textit{Softmax length ($M$).} The number of positions $M$ is set by a uniform arclength discretization with step $\Delta u$ and fixed to the maximum sequence length observed across the training set under this sampling. Shorter sequences are right-padded to length $M$ and masked in the loss; the target stop label is a one-hot vector of length $M$ indicating the terminal index.
\subsection{Loss Function}

The custom loss function for training the model combines multiple components to handle the point-wise tip error, variable guidewire length (stop criteria), and tip position predictions. The overall loss function $\mathcal{L}_{\text{total}}$ is defined as:

\begin{equation}\label{eq:loss_fn}
	\begin{split}
		\mathcal{L}_\text{total} = \frac{1}{N} \sum_{i=1}^N \bigg(
		 & \lambda_\text{tip} \left\| \hat{\mathbf{p}}_i - \mathbf{p}_i \right\|^2 +                                                                         \\
		 & \lambda_\text{offset} \big( (\hat{\boldsymbol{\phi}}_i - \boldsymbol{\phi}_i)^2 + (\hat{\boldsymbol{\theta}}_i - \boldsymbol{\theta}_i)^2 \big) + \\
		 & \lambda_\text{stop} \big( -\mathbf{s}_i \log(\hat{\mathbf{s}}_i) - (1 - \mathbf{s}_i) \log(1 - \hat{\mathbf{s}}_i) \big) \bigg)
	\end{split},
\end{equation}

where $N$ is the number of samples, and $\lambda_{\text{tip}}$, $\lambda_{\text{offset}}$, and $\lambda_{\text{stop}}$ are weights that balance the contributions of each loss component. The tip prediction loss ($\mathcal{L}_{\text{tip}}$) uses \gls{mean squared error} to ensure accurate 3D tip coordinates. The spherical offset loss ($\mathcal{L}_{\text{offset}}$) also uses \gls{mean squared error} to align predicted and ground truth angular offsets, capturing the guidewire's shape. The stop prediction loss ($\mathcal{L}_{\text{stop}}$) employs \gls{binary cross-entropy} to accurately predict the guidewire's endpoint.

\paragraph{Stop criterion (variable length).}
Each guidewire curve is uniformly sampled along arclength into $M$ positions (step $\Delta u$). For sample $i$, a one-hot vector $\mathbf{s}_i \in \{0,1\}^{M}$ marks the terminal index $j_i^\star$. The network outputs $\hat{\mathbf{s}}_i \in [0,1]^M$ over these positions. We fix $M$ to the maximum length observed in the training set under this discretization; shorter sequences are right-padded and a binary mask excludes padded indices from the loss. At inference, the terminal index is $\hat{j}_i=\arg\max_j \hat{s}_{i,j}$, and the predicted curve is truncated at $\hat{j}_i$.

\subsection{Training Details} The model was trained end-to-end using the loss from Eq.~\ref{eq:loss_fn}. The NAdam \autocite{dozat2016incorporating} optimizer was used with an initial learning rate of $1 \times 10^{-4}$. Additionally, a learning rate scheduler was employed to adjust the learning rate dynamically based on the validation loss. Specifically, the ReduceLROnPlateau scheduler was configured to reduce the learning rate by a factor of $0.1$ if the validation loss did not improve for \num{10} epochs. The model was trained for 400 epochs, with early stopping based on the validation loss to further prevent overfitting. In all experiments, data were partitioned into train/validation/test sets in an 80/10/10 ratio at the acquisition-sequence level to avoid temporal leakage, with all frames from the same sequence --- and their bi-planar counterparts (cameras A/B) and with/without-flow pairs --- kept within the same split (a fixed random seed ensured reproducibility).

\subsection{Shape Prediction Network}

The task is formulated as a supervised regression problem: given a 2D image $\mathbf{I} \in \mathbb{R}^{80 \times 80}$ of the guidewire, the network predicts a set of 3D coordinates $\hat{\mathbf{y}} = \{\hat{y}_1, \ldots, \hat{y}_n\}$ corresponding to the reconstructed guidewire shape. Each sample thus maps from a single-channel image input to a variable-length sequence of 3D spatial positions in $\mathbb{R}^3$ that represent the underlying geometry of the guidewire.

After deriving the reconstruction of the guidewire shape, \gls{3d-fgrn} is trained to approximate the instrument shapes from 2D images. \gls{3d-fgrn} is composed of two main parts: a convolutional feature extraction module and a linear classification module. The convolutional module consists of three convolutional layers with kernel sizes of $3 \times 3$ and filter depths of 32, 64, and 128, respectively. Each convolutional layer is followed by a \gls{relu} \autocite{agarap2019deep} activation function, which introduces non-linear transformations essential for intricate pattern delineation. Max-pooling layers~\autocite{nagi2011max} with a window size of $2 \times 2$ are interspersed after each activation, aiming to reduce spatial dimensionality and enhance the representational potency of the extracted feature maps.

The linear classification module, following the convolutional block, consists of two fully connected layers of sizes 256 and 128, each followed by a \gls{relu} activation and a dropout layer \autocite{srivastava2014dropout} with dropout rate $p = 0.3$ as a regularization technique. The final output layer maps to the predicted 3D positions of the guidewire points.

The training dataset is composed of image observations of shape $80 \times 80$ paired with the derived 3D positions, $p$, of the guidewire bodies, enabling the model to learn the mapping from 2D images to the reconstructed 3D shapes. The \gls{cnn} is optimized using the NAdam \autocite{dozat2016incorporating} optimizer, with a refined loss function that combines the Huber Loss and a regularization term to ensure both the accuracy of the predictions and the preservation of the geometric structure of the guidewire.

The Huber Loss, $\mathcal{L}_{\text{Huber}}$, is defined as:

\begin{equation}
	\mathcal{L}_{\text{Huber}}(y, \hat{y}) = \begin{cases}
		\frac{1}{2}(y - \hat{y})^{2}             & \quad \text{if } \left | y - \hat{y} \right | < 1 \\
		\left| y - \hat{y} \right| - \frac{1}{2} & \quad \text{otherwise}
	\end{cases}
\end{equation}

Here, $y$ represents the ground truth, the derived 3D positions of the guidewire bodies, and $\hat{y}$ represents the predicted positions by the model.

To maintain the geometric integrity of the guidewire, a regularization term, $\mathcal{L}_{\text{reg}}$, is introduced, which penalizes the deviation of the interbody spacing from $s = 0.002\,\unit{m}$. This spacing value was chosen to match the simulation configuration, ensuring consistency between the physical model and its simulated counterpart and allowing for direct comparison of deformation behaviour.

\begin{equation}
	\mathcal{L}_{\text{reg}} = \frac{1}{n-1} \sum_{i=1}^{n-1} \left| \lVert \hat{y}_{i+1} - \hat{y}_i \rVert_2 - s \right|
\end{equation}

The final loss function, $\mathcal{L}$, is a weighted sum of the Huber Loss and the regularization term, ensuring both accurate predictions and preservation of the guidewire’s geometric structure:

\begin{equation}
	\mathcal{L}= \alpha \mathcal{L}_{\text{Huber}}(y, \hat{y}) + \beta \mathcal{L}_{\text{reg}}
\end{equation}

Similarly to the 3D reconstruction, a univariate spline interpolation \autocite{dierckx1995curve} is applied to the 3D points, thus getting a smooth representation of the guidewire.

\section{Experiments}

The proposed dataset, Guide3D, is evaluated through a structured experimental analysis, focusing on three key aspects. First, the dataset's validity is assessed by analysing reprojection errors and their distribution to evaluate its accuracy and reliability (Subsection~\ref{ch5-subsec:dataset-validation}). Next, the applicability of Guide3D in the context of a 3D reconstruction task is investigated, examining its effectiveness in accurately modelling 3D shapes (Subsection~\ref{ch5-subsec:shape-prediction-result}). Finally, multiple segmentation algorithms are benchmarked to evaluate their performance on Guide3D, providing a comprehensive understanding of the dataset’s utility for segmentation tasks (Section~\ref{ch5-subsec:segmentation-results}).

\subsection{Dataset Validation}\label{ch5-subsec:dataset-validation}

\begin{figure}[ht]
	\centering
	\subfloat{\includegraphics[width=.5\linewidth]{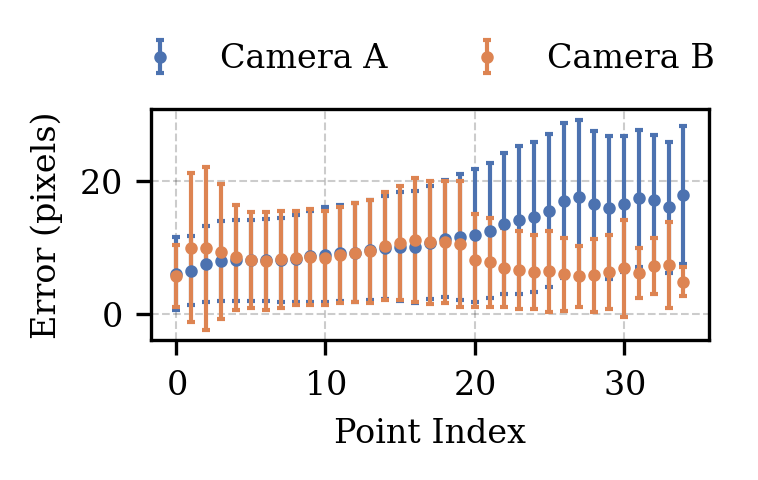}}\hfill
	\raisebox{0.3cm}{\subfloat{\includegraphics[width=.4\linewidth]{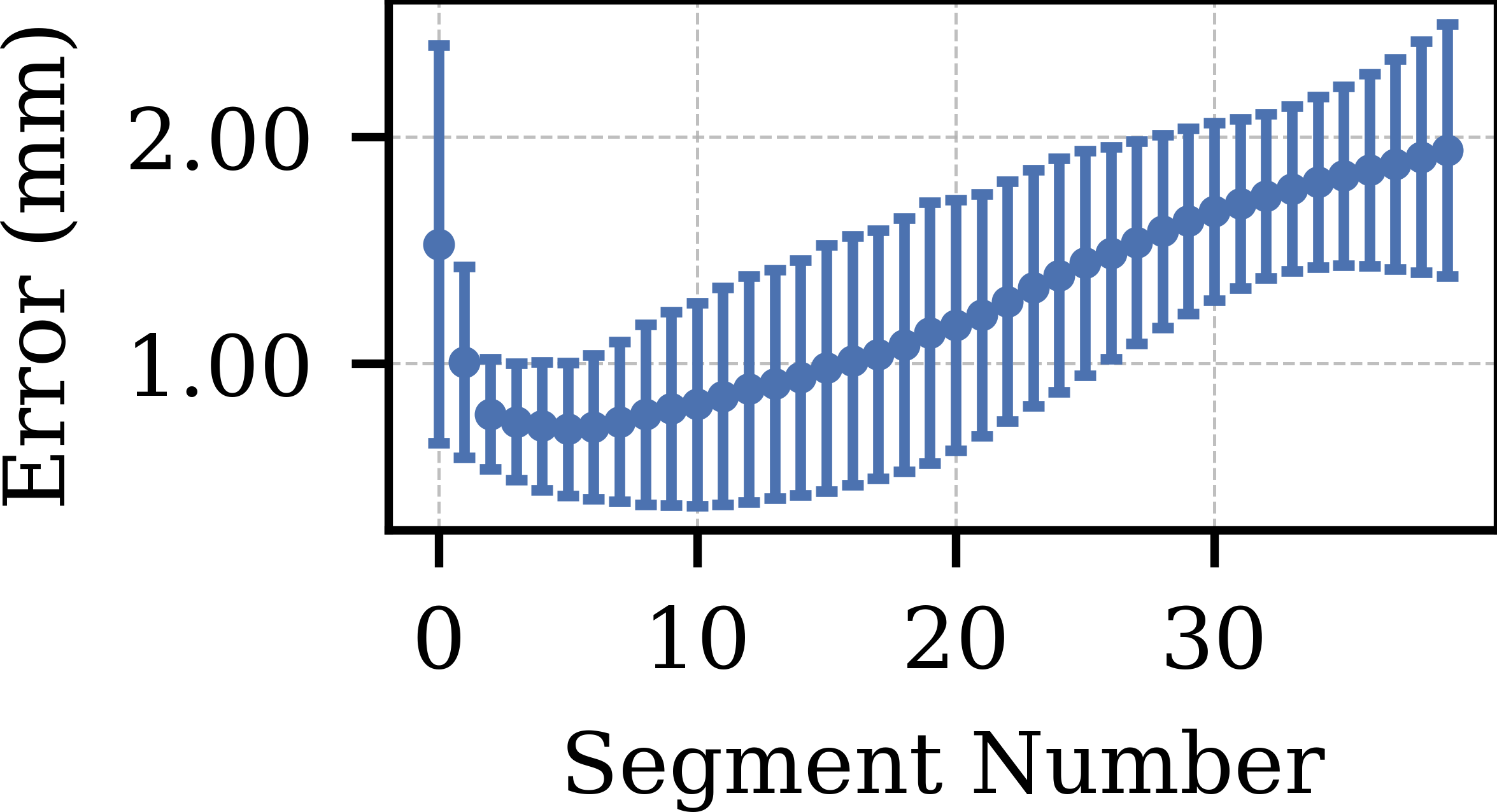}}}
	\mycaption{Guidewire Reconstruction Error Analysis}{(Left) Reprojection error distribution along the guidewire, where the x-axis denotes the point index obtained from a uniform discretization of the guidewire in 2D image space. (Right) Reconstruction error per segment, where the x-axis indicates the corresponding segment index in 3D space.}
	\label{ch5-fig:reprojection-error}
\end{figure}

The analysis revealed a non-uniform distribution of reprojection errors across the dataset, with the highest variability and errors concentrated at the proximal end of the guidewire reconstructions. As shown in Figure~\ref{ch5-fig:reprojection-error}, the reprojection error patterns differ significantly between Camera $A$ and Camera $B$. For Camera $A$, mean errors increase progressively from approximately \(6\)px to a peak of \(20\)px, with standard deviations rising from \(5\)px to \(11\)px, indicating both growing inaccuracies and increasing variability over time. Notably, significant fluctuations around indices \(25\) to \(27\) correspond to periods of particularly high error. In contrast, Camera $B$ demonstrates an initial peak in mean errors at \(9\)px at index 1, followed by fluctuations that diminish towards the dataset's end. The standard deviations for Camera $B$ start at \(11\)px and decrease over time, reflecting high initial variability that stabilizes later. These patterns align with the inherent flexibility of the guidewire, which is capable of forming complex shapes such as loops.

Additionally, a validation procedure was conducted using CathSim~\autocite{jianu2024cathsim}, incorporating the aortic arch model detailed in Subsection~\ref{ch5-subsec:materials} and a guidewire with comparable diameter and mechanical properties. For sampling, the \gls{sac} algorithm was employed, leveraging segmented guidewire and kinematic data to generate realistic validation samples. Evaluation metrics revealed a \gls{maxed} of \(\num{2.880} \pm \num{0.640} \unit{\milli\metre}\), a \gls{mete} of \(\num{1.527} \pm \num{0.877} \unit{\milli\metre}\), and a \gls{mers} of \(\num{0.001} \pm \num{0.000}\unit{\milli\metre}\). These results confirm the precision and reliability of the proposed methodology.

\subsection{Guidewire Prediction Results}\label{ch5-subsec:shape-prediction-result}

\begin{sidebysidefigures}
	\begin{leftfigure}
		\captionof{table}[Shape Comparison]{Shape Comparison (mm).}\label{ch5-tab:shape-accuracy}
\centering
\scriptsize
\begin{tabular}{
		l
		S[table-format=2.2(1.2)]
		S[table-format=1.2(1.2)]
	}
	\toprule
	                     & {\thead{Cartesian}} & {\thead{Spherical}}  \\
	\midrule
	MaxED~\(\downarrow\) & 10.00(4.64)         & \bfseries 6.88(5.23) \\
	METE~\(\downarrow\)  & 6.93(3.94)          & \bfseries 3.28(2.59) \\
	MERS~\(\downarrow\)  & 5.33(2.73)          & \bfseries 4.54(3.67) \\
	\bottomrule
\end{tabular}

	\end{leftfigure}%
	\begin{rightfigure}
		\captionof{table}{Segmentation Results.}\label{ch5-tab:segmentation-results}
\centering
\scriptsize
\begin{tabular}{
		l
		S[table-format=2.2]
		S[table-format=2.2]
		S[table-format=2.2]
	}
	\toprule
	                                       & {\thead{DiceM}} & {\thead{mIoU}}  & {\thead{Jaccard}} \\
	\midrule
	UNet \autocite{ronneberger2015u}       & 92.25           & 36.60           & 86.57             \\
	TransUnet \autocite{chen2021transunet} & \bfseries 95.06 & \bfseries 41.20 & \bfseries 91.10   \\
	SwinUnet \autocite{cao2022swin}        & 93.73           & 38.58           & 88.55             \\
	\bottomrule
\end{tabular}



	\end{rightfigure}
\end{sidebysidefigures}

The capability of the network introduced in Section~\ref{ch5-sec:3D-shape-prediction} is demonstrated, highlighting the importance of the proposed dataset. The network's predictions are evaluated through the following steps: \textit{1)} an analysis of the predicted versus reconstructed curves using piecewise metrics, and \textit{2)} an examination of the reprojection error.

\subsubsection{Shape Prediction Errors}

\begin{figure}[ht]
	\centering
	\hfill
	\subfloat[]{\includegraphics[width=.4\columnwidth]{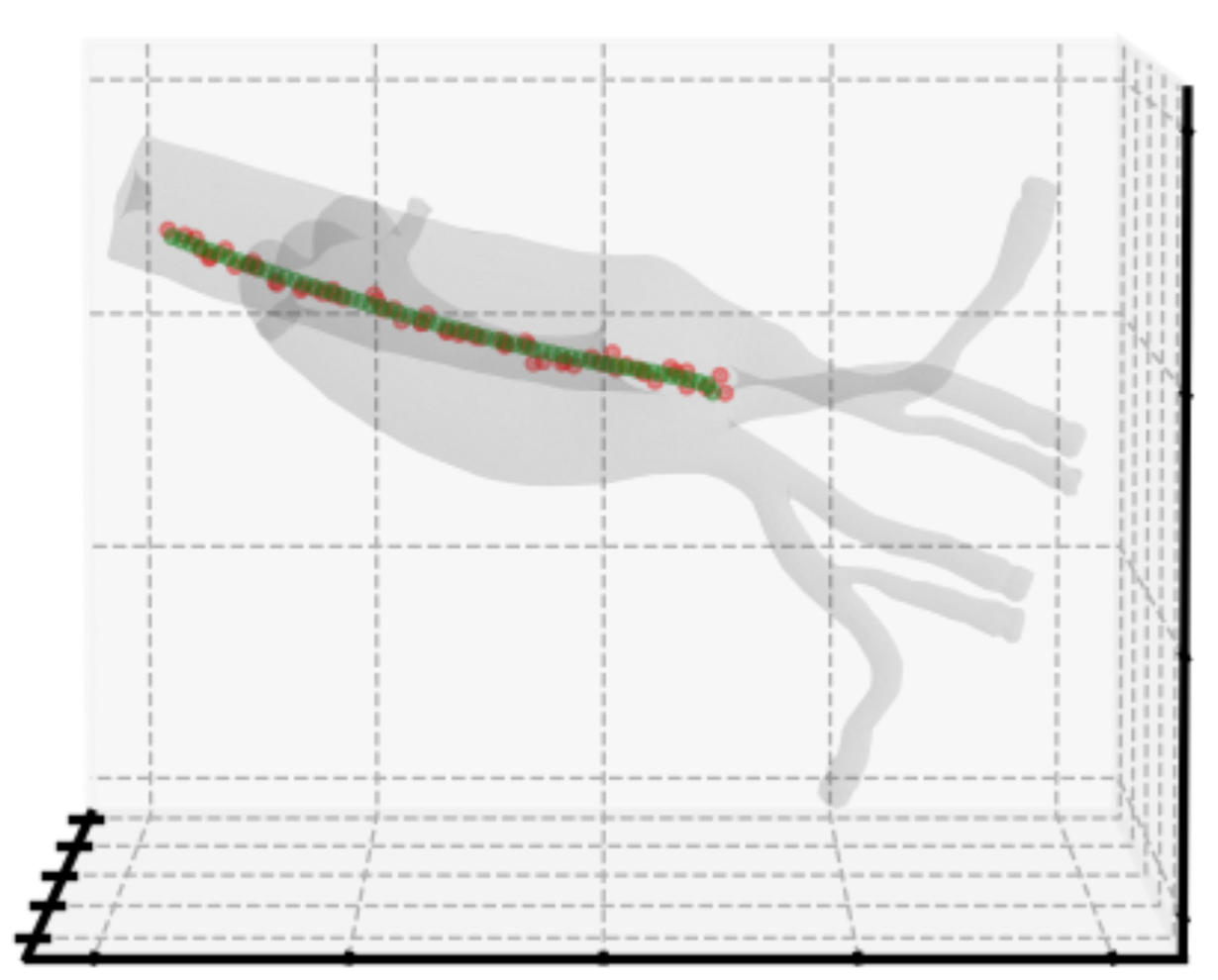}}\hfill
	\subfloat[]{\includegraphics[width=.4\columnwidth]{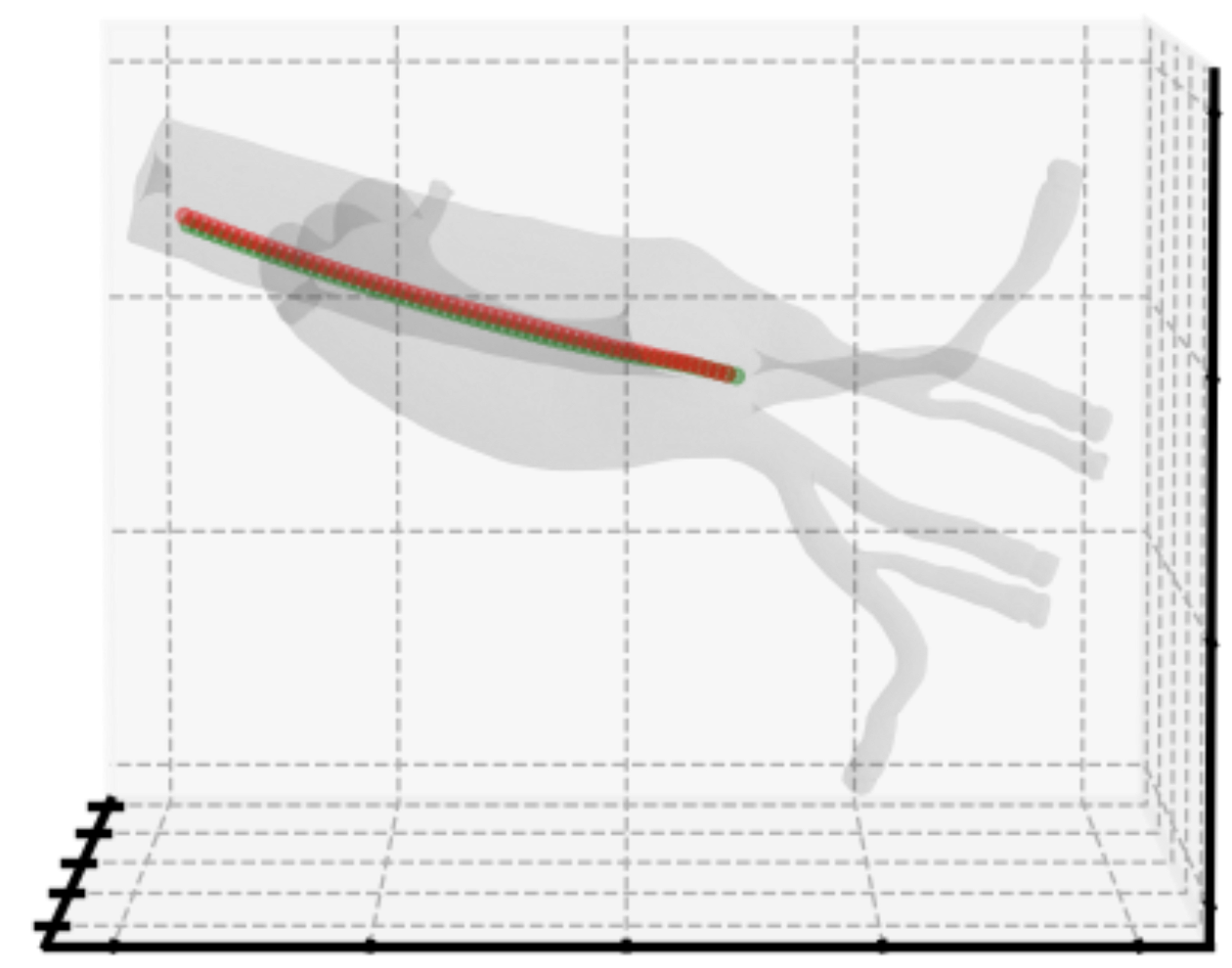}}\hfill
	\mycaption{Guidewire Shape Prediction Comparison}{Comparison between the predictions of the method not employing spherical coordinates (left) and the method employing spherical coordinates (right). It is evident that the method using spherical coordinates results in a more uniform and accurate shape reconstruction.}
	\label{ch5fig:simple-vs-spherical}
\end{figure}

Table~\ref{ch5-tab:shape-accuracy} presents a comparison of various metrics for shape prediction accuracy. As outlined in Subsection~\ref{ch5-subsec:dataset-validation}, the shape differences are quantified using the following metrics: \textit{1)} \acrfull{maxed}, \textit{2)} \acrfull{mete}, and \textit{3)} \acrfull{mers}. For consistency, the shape of the guidewire, represented as a 3D curve \( \mathbf{C}(u) \), is sampled at equidistant \( \Delta u \) intervals along the arclength parameter \( u \). Consequently, these metrics reflect the pointwise discrepancies between the predicted and reconstructed shapes along the curve's arclength.

The results demonstrate that the Spherical representation consistently outperforms the Cartesian representation across all metrics. Specifically, the \gls{maxed} error is significantly lower for the Spherical representation (\( 6.88 \pm 5.23 \unit{\milli\meter}\)) compared to the Cartesian representation (\( 10.00 \pm 4.64 \unit{\milli\meter}\)). Similarly, the \gls{mete} is reduced in the Spherical representation (\( 3.28 \pm 2.59 \unit{\milli\meter}\)) compared to the Cartesian representation (\( 6.93 \pm 3.94\unit{\milli\meter}\)). For the \gls{mers}, the Spherical representation achieves a smaller error (\( 4.54 \pm 3.67 \)\unit{\milli\meter}) than the Cartesian representation (\( 5.33 \pm 2.73\unit{\milli\meter}\)). Finally, the Fréchet distance further underscores the superiority of the Spherical representation, with a lower error (\( 6.70 \pm 5.16\unit{\milli\meter}\)) compared to the Cartesian representation (\( 8.95 \pm 4.37\unit{\milli\meter}\)). This can also be visually observed in Fig.~\ref{ch5fig:simple-vs-spherical}.

\subsubsection{Shape Comparison Visualization}

\begin{figure}[t]
	\centering
	\includegraphics[width=.75\textwidth]{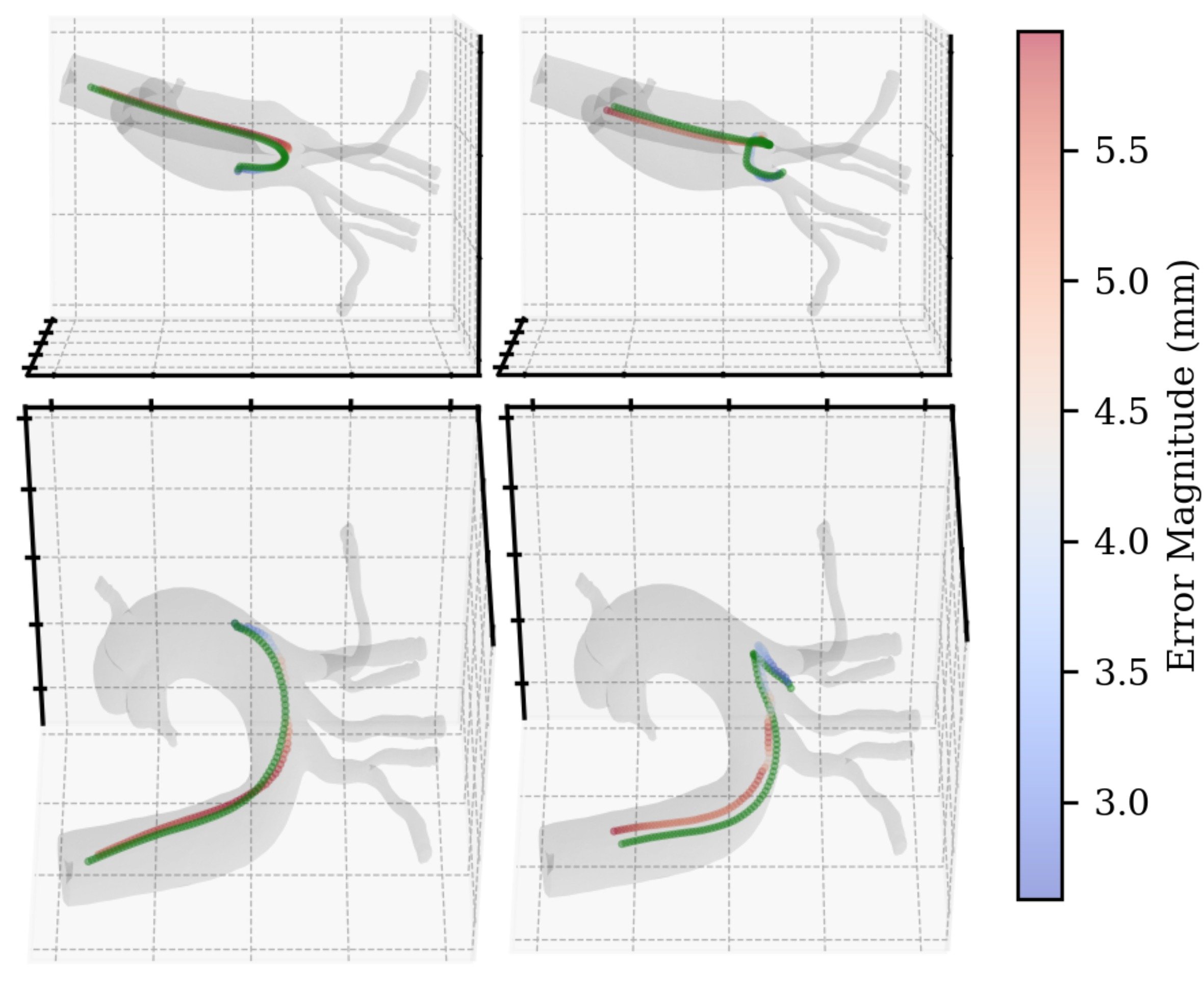}
	\mycaption{Guidewire Shape Predictions}{The figure shows the predicted 3D guidewire shape overlaid with a colormap indicating confidence or positional information. The results demonstrate the network's ability to accurately reconstruct guidewire geometry, even in challenging configurations.}
	\label{ch5fig:3D}
\end{figure}

\begin{figure}[t]
	\centering
	\includegraphics[width=.72\textwidth]{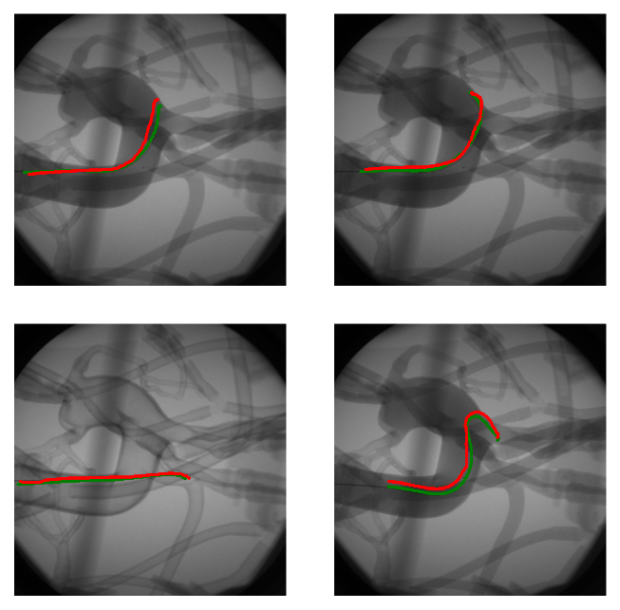}
	\mycaption{Reprojection Visualization}{The predicted 3D guidewire is reprojected into the image plane for visual comparison. The accurate overlap between predicted and actual guidewire projections highlights the robustness of the reconstruction network, even under angular distortions.}
	\label{ch5fig:reprojection_visualization}
\end{figure}

Figure~\ref{ch5fig:3D} presents two 3D plots from different angles, comparing the ground truth guidewire shape to the predicted shape generated by the network. The network demonstrates strong capability in accurately predicting the guidewire’s configuration, even in challenging scenarios involving loops and self-obstructions in the images. The predicted shape closely aligns with the ground truth, highlighting the effectiveness of the network in handling complex guidewire geometries.

Notably, the proximal end exhibits a greater deviation compared to the distal end, with discrepancies ranging from \( 2 \unit{mm} \) at the distal end to \( 5 \unit{mm} \) at the proximal end. These results underscore the network’s robustness in predicting the guidewire's shape, leveraging only consecutive single-plane images. To validate the predictions further, the 3D points are reprojected onto the original images, as illustrated in Figure~\ref{ch5fig:reprojection_visualization}, demonstrating consistent alignment with the input data.

\subsection{Segmentation Results}\label{ch5-subsec:segmentation-results}

\begin{figure}[ht]
	\centering
	\subfloat{\includegraphics[width=0.4\linewidth]{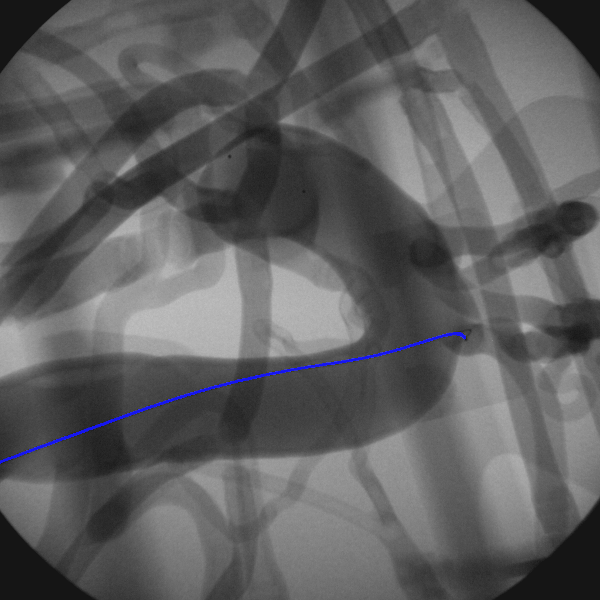}}
	\hspace{1em}
	\subfloat{\includegraphics[width=0.4\linewidth]{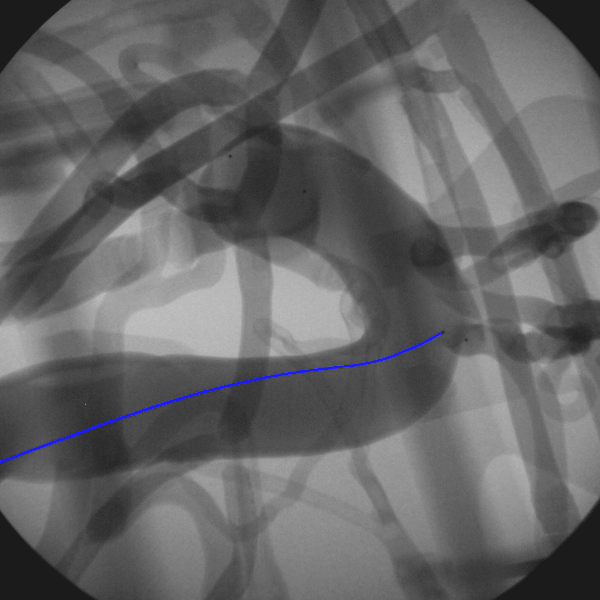}}
	\mycaption{Segmentation Examples}{Images illustrating the segmentation of the guidewire from fluoroscopic images using SwinUnet~\cite{cao2022swin} architecture.}
	\label{fig:segmentation-examples}
\end{figure}

The potential of Guide3D to advance guidewire segmentation research is evaluated by assessing the performance of three state-of-the-art network architectures: \textit{1)} UNet \autocite{ronneberger2015u}, \textit{2)} TransUnet \autocite{chen2021transunet}, and \textit{3)} SwinUnet \autocite{cao2022swin}. The models were trained under the following configurations: UNet with a learning rate of \( 1 \times 10^{-5} \) for 135 epochs, TransUnet incorporating ResNet50 and Vision Transformer (ViT-B-16) with a learning rate of 0.01 for 199 epochs, and SwinUnet based on the Swin Transformer architecture with a learning rate of 0.01 for 299 epochs.

Performance metrics included the Dice Coefficient (DiceM), mean \gls{mIoU}, and Jaccard index, as summarized in Table~\ref{ch5-tab:segmentation-results}. Among the models, TransUnet achieved the highest performance, with a \gls{dice coefficient} of 95.06, an \gls{mIoU} of 41.20, and a Jaccard index of 91.10. SwinUnet followed with a \gls{dice coefficient} of 93.73, an \gls{mIoU} of 38.58, and a Jaccard index of 88.55, while UNet achieved a \gls{dice coefficient} of 92.25, an \gls{mIoU} of 36.60, and a Jaccard index of 86.57. Fig.~\ref{fig:segmentation-examples} presents example outputs from SwinUnet, illustrating its ability to segment the guidewire in fluoroscopic images.

These findings benchmark the dataset's performance and highlight the efficacy of modern transformer-based architectures, particularly TransUnet, for guidewire segmentation tasks. However, challenges such as loops and occlusions in the guidewire remain, suggesting that incorporating polyline prediction methods could significantly enhance the dataset's utility for complex scenarios.

\section{Discussion and Conclusion}

In this study, \textbf{Guide3D} is introduced, a publicly available bi-planar endovascular navigation dataset designed for the segmentation and 3D reconstruction of flexible, curved endovascular tools, addressing a significant gap in medical imaging research. While extensive experiments demonstrate the dataset's value, limitations are acknowledged, such as the absence of clinical real human data due to regulatory challenges and the focus on synthetic and experimental scenarios that may not fully capture the variability of real-world clinical environments. Nevertheless, the standardized platform accommodates both video and image-based approaches, providing a versatile resource to facilitate the translation of these technologies into clinical settings. By including complexities like the guidewire's flexibility and the presence of loops and occlusions, the aim is to push the boundaries of current methodologies, although further validation with clinical data is necessary to ensure robustness and generalizability. The objective is to bridge the disparity between research developments and clinical application by establishing a standardized framework for evaluating various methodologies.

The Guide3D dataset represents a significant contribution to the field by providing researchers with a high-quality, annotated resource that captures the complex dynamics of endovascular tools. With over 8,700 annotated images across varied flow conditions, it enables systematic evaluation and comparison of different reconstruction approaches. This comprehensive dataset not only facilitates algorithm development but also establishes benchmarks for performance metrics in the domain. The dataset introduced in this chapter provides the foundation for developing advanced geometric modelling techniques, which will be explored in the next chapter through the SplineFormer model for guidewire shape prediction.

%
\onehalfspacing
\chapter{SplineFormer}
\chaptermark{SplineFormer}
\glsresetall

\begin{cabstract}
	Endovascular navigation constitutes a critical component of minimally invasive procedures, where the precise manipulation of curvilinear instruments, such as guidewires, is essential for successful intervention. One of the primary challenges in this domain involves the accurate prediction of the guidewire’s evolving shape as it traverses the vascular system. This task is complicated by the complex deformations that result from continuous interaction with vessel walls. Traditional segmentation-based approaches often fail to deliver reliable real-time shape estimations, limiting their utility in highly dynamic clinical environments. To overcome these limitations, a novel transformer-based architecture, \textit{SplineFormer}, is proposed for predicting the continuous and smooth shape of the guidewire in an interpretable manner. By leveraging the inherent attention mechanisms of transformers, the model effectively captures intricate patterns of bending and twisting, and represents the guidewire trajectory as a B-spline. This formulation enhances both the accuracy and smoothness of the predicted shape. The proposed SplineFormer model is integrated into an end-to-end robotic navigation framework, where its compact and informative representation contributes directly to autonomous decision-making. Experimental evaluations demonstrate that the SplineFormer enables autonomous endovascular navigation and achieves a success rate of 50\% in cannulating the brachiocephalic artery using a real robotic platform.
\end{cabstract}

This chapter presents the work from the following publications:

\fullcite{jianu2025splineformer}

\section{Experiments}

\begin{figure}[h]
	\centering
	\subfloat[]{\includegraphics[width=0.37\linewidth]{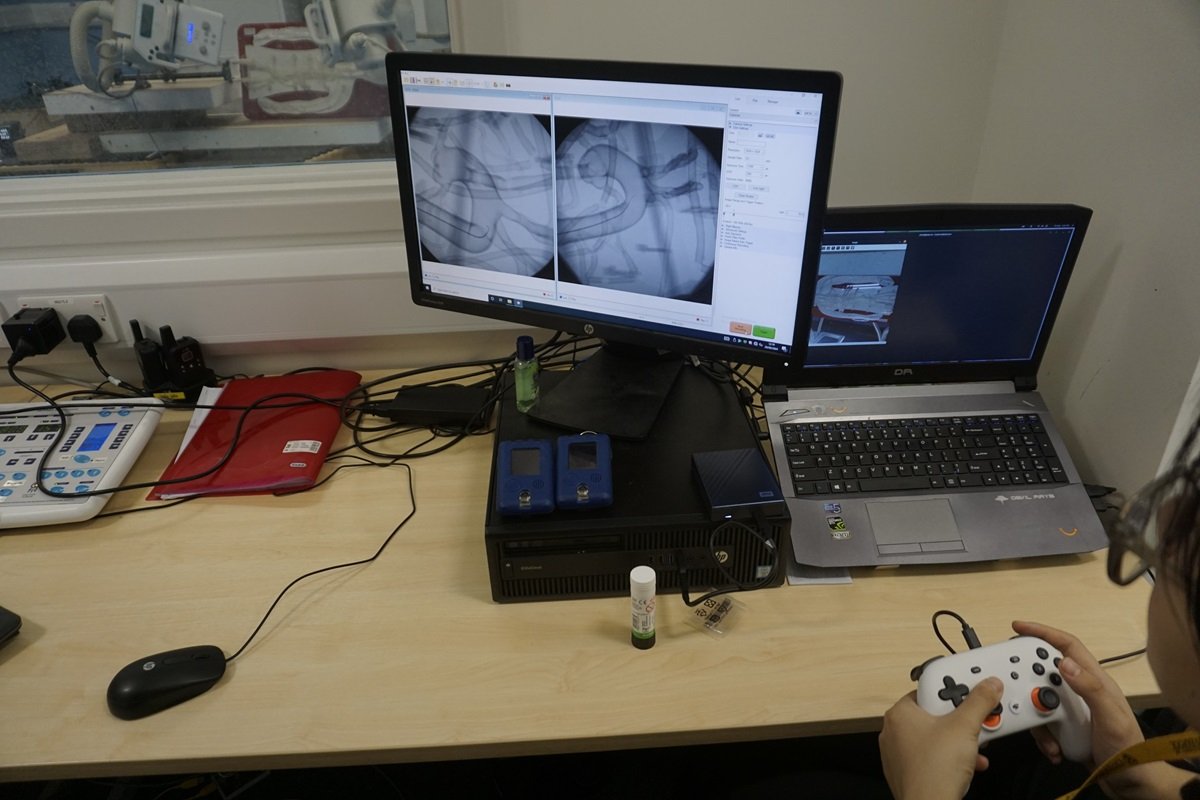}} \hspace{1ex}
	\subfloat[]{\includegraphics[width=0.37\linewidth]{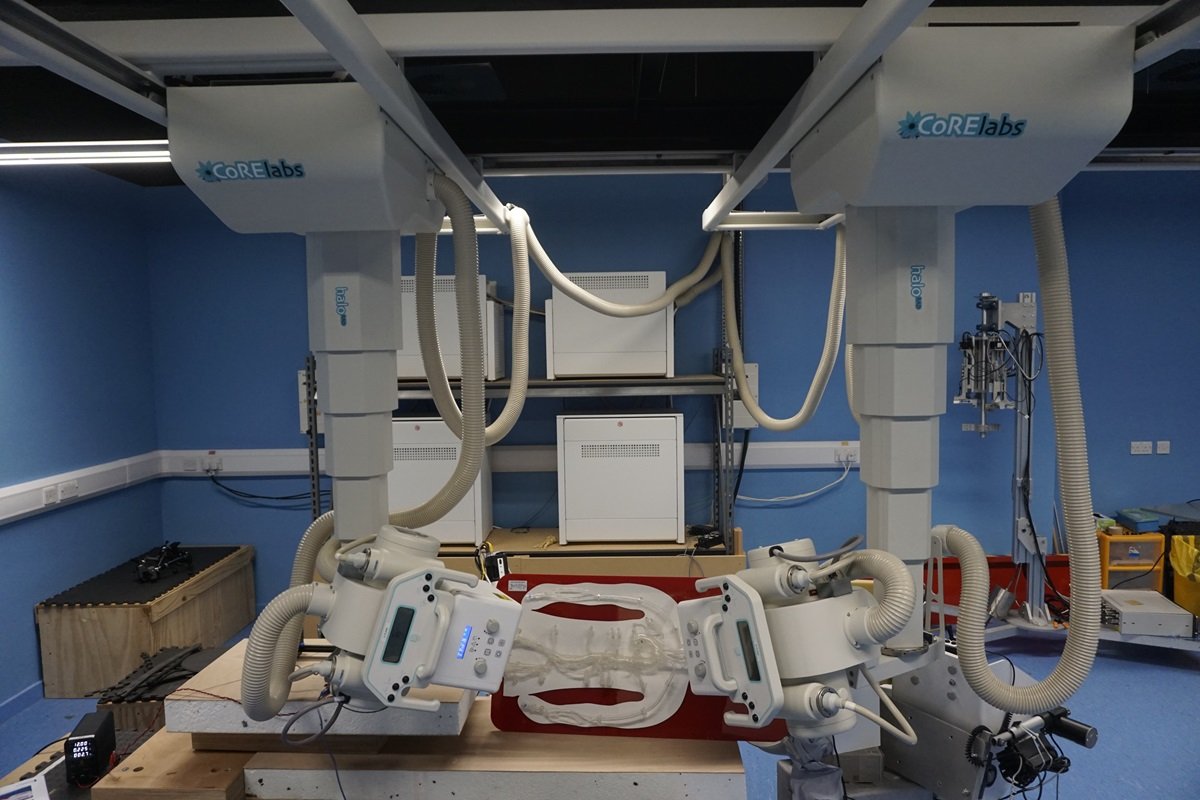}} \hspace{1ex}
	\subfloat[]{\includegraphics[width=0.2\linewidth, height=0.247\linewidth]{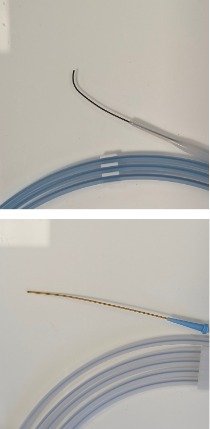}}
	\mycaption{Robot Setup in X-ray Room}{The data acquisition and teleoperation process is highlighted, showcasing the user navigating the guidewire towards the designated target within the vascular phantom. The setup employs a Bi-planar X-ray. To enhance data variability, two different guidewires are used—the Radifocus™ Guide Wire M Stiff Type with an angled tip and the Nitrex Guidewire straight tip.}
	\label{ch6fig:materials}
\end{figure}

To evaluate the effectiveness of SplineFormer, the real endovascular robot was set up and data were collected to train the network. After training, the learned policy was evaluated on state-action pairs within the endovascular robotic system to assess how effectively the predicted B-spline representation supports autonomous guidewire navigation. All experiments were conducted on a computer equipped with an NVIDIA RTX 4080 GPU, \(\qty{128}{\giga\byte}\) RAM, and an Intel Core i9-13900 processor, implemented using PyTorch.

\subsection{Endovascular Robot Setup}\label{ch6sec:robotic-setup}

\paragraph{Robot Setup.} Endovascular robotic systems commonly utilize a leader-follower (master-slave) architecture, where the leader device—operated by a clinician—translates input commands to a follower robot that manipulates the catheter~\cite{Song2023-ns}. These systems typically offer up to six \gls{degrees of freedom}, enabling precise translation and rotation control~\cite{saliba2006novel}. Human-Machine Interfaces (HMIs), such as multi-\gls{degrees of freedom} joysticks or handheld controllers, convert operator movements into electromechanical actions, facilitating accurate catheter navigation~\cite{saliba2006novel,khan2013first}. In the present system, since only a guidewire is used, the robotic setup is streamlined to focus solely on translation and rotation movements. This simplification reduces mechanical complexity, making the design more accessible and easier to replicate compared to multi-\gls{degrees of freedom} systems. The actuation mechanism comprises a NEMA 17 Bipolar stepper motor (\(\SI{59}{\newton\centi\meter}\), \(\SI{2}{\ampere}\)) for linear motion, along with an additional motor for rotational control. System control is managed by an Arduino Uno Rev3 with a CNC shield and two A4988 drivers, powered by a \(\SI{12}{\volt}\) DC power supply. Teleoperation input is provided via a Google Stadia joystick for intuitive manual control.

\paragraph{Data Collection.} The experiments utilize a Bi-planar X-ray system (Fig.~\ref{ch6fig:materials}) equipped with \(\SI{60}{\kilo\watt}\) Epsilon X-ray Generators (EMD Technologies Ltd.) and \(16\)-inch Image Intensifier Tubes (Thales), incorporating dual focal spot Variant X-ray tubes for high-definition imaging. System calibration is achieved using acrylic mirrors and geometric alignment grids. To simulate human vascular anatomy, a half-body vascular phantom model (Elastrat Sarl Ltd., Switzerland) is employed, enclosed in a transparent box and integrated into a closed water circuit to replicate blood flow. The model, constructed from soft silicone and featuring continuous flow pumps, was derived from detailed postmortem vascular casts, ensuring anatomical accuracy consistent with human vasculature~\cite{martin1998vitro,gailloud1999vitro}. A Radifocus™ Guide Wire M Stiff Type (Terumo Ltd.), a \(\SI{0.89}{\milli\meter}\) nitinol wire with a \(\SI{3}{\centi\meter}\) angled tip, and a Nitrex Guidewire with a straight tip were utilized.

\paragraph{Dataset.} A dataset of \num{8746} high-resolution samples (\(1,024 \times 1,024\) pixels) was collected, consisting of \(4,373\) paired instances with and without a simulated blood flow medium. Specifically, the dataset includes \(6,136\) samples from the Radifocus Guidewire and \(2,610\) from the Nitrex Guidewire, establishing a foundation for automated guidewire tracking in bi-planar X-ray images. Manual annotation was performed using the CVAT tool~\cite{cvat2023}, where polylines were meticulously created to capture the dynamic trajectory of the guidewire with high precision. To ensure balanced representation across different guidewire types and imaging conditions, the dataset was partitioned using a stratified sampling method.

\subsection{Autonomous Navigation Results}

\begin{figure}[h]
	\centering
	\includegraphics[width=0.5\linewidth]{assets/setup.pdf}
	\mycaption{Navigation Configuration:}{The guidewire tip is situated in the descending aorta. Two main targets are aimed for navigation, namely the \acrfull{bca}, and the \acrfull{lcca}.}\label{ch6fig:target-setup}
\end{figure}

SplineFormer was evaluated in a \textit{fully autonomous} guidewire navigation task. The objective was to navigate from a predefined position in the descending aorta toward two distinct arterial targets: the \gls{bca} and the \gls{lcca}, as illustrated in Fig.~\ref{ch6fig:target-setup}. For each target, the system performed 20 trials, recording trajectories to construct datasets of state-action pairs. The agent operated using fluoroscopic image observations, with an action space defined by translation within \(\pm 2\unit{mm}\) and rotation within \(\pm 15\unit{\degree}\).

Following training, SplineFormer was deployed on the robotic platform for fully autonomous navigation. The system achieved a 50\% success rate in reaching the \gls{bca}, with a mean completion time of \num{2.5(0.76)}\,min. This represents a substantial improvement over the baseline \gls{bca} method, which achieved only 5.6\% success under identical autonomous conditions. The semi-autonomous GAIL-PPO approach, which incorporates human demonstrations, performed better with success rates of 69.4\% for the \gls{bca} and 72.2\% for the \gls{lcca}, but unlike SplineFormer, it still required human intervention.

\begin{table}[ht]
	\centering
	\caption{Endovascular navigation results.}\label{ch6tab:navigation-result}
	\fontsize{9.6}{11}\selectfont
	\begin{originaltabular}{l l l S[table-format=3.1] S[table-format=3.1] S[table-format=3.1] S[table-format=3.1]}
		\toprule
		&                  &                      & \multicolumn{2}{c}{\thead{BCA}} & \multicolumn{2}{c}{\thead{LCCA}}                                                 \\
		\cmidrule(lr){4-5} \cmidrule(lr){6-7}
		\thead{Method}                                  & \thead{Setup}    & \thead{Explainable?} & {\textit{Success (\%)}}         & {\textit{Time (s)}}              & {\textit{Success (\%)}} & {\textit{Time (s)}} \\
		\midrule
		Expert Teleoperation                            & Fully Manual     & ---                  & 100.0                           & 32.1                             & 100.0                   & 25.0                \\
		GAIL-PPO\autocite{chi2020collaborative}         & Semi-Autonomous  & No                   & 69.4                            & 52.1                             & 72.2                    & 76.5                \\
		Behavior Cloning\autocite{chi2020collaborative} & Fully Autonomous & No                   & 5.6                             & \multicolumn{1}{c}{\(\geq 200\)} & \multicolumn{2}{c}{---}                       \\
		\textbf{Splineformer (Ours)}                    & Fully Autonomous & Yes                  & 50.0                            & 150.0                            & \multicolumn{2}{c}{---}                       \\
		\bottomrule
	\end{originaltabular}
\end{table}

While this method did not surpass GAIL-PPO~\cite{chi2020collaborative}, a semi-autonomous approach, in success rate, it offers the advantage of full autonomy. Moreover, the B-spline representation provides an explainable and structured state space, improving model interpretability. However, neither SplineFormer nor \gls{behavioural cloning} successfully cannulated the \gls{lcca}, highlighting challenges in navigating more complex vascular geometries. The failure in \gls{lcca} navigation can be attributed to its sharper curvature, narrower diameter, and more abrupt bifurcation angle compared to the \gls{bca}. As reflected in the success rates in Table~\ref{ch6tab:navigation-result}, the guidewire was generally able to advance into the aortic arch and perform rotational manoeuvres, but it failed to align with and enter the ostium of the target branch. Instead, it bypassed the branch and continued within the arch, underscoring that the main limitation was branch cannulation rather than forward progression.

\begin{figure}[!ht]
	\centering
	\includegraphics[width=1\linewidth]{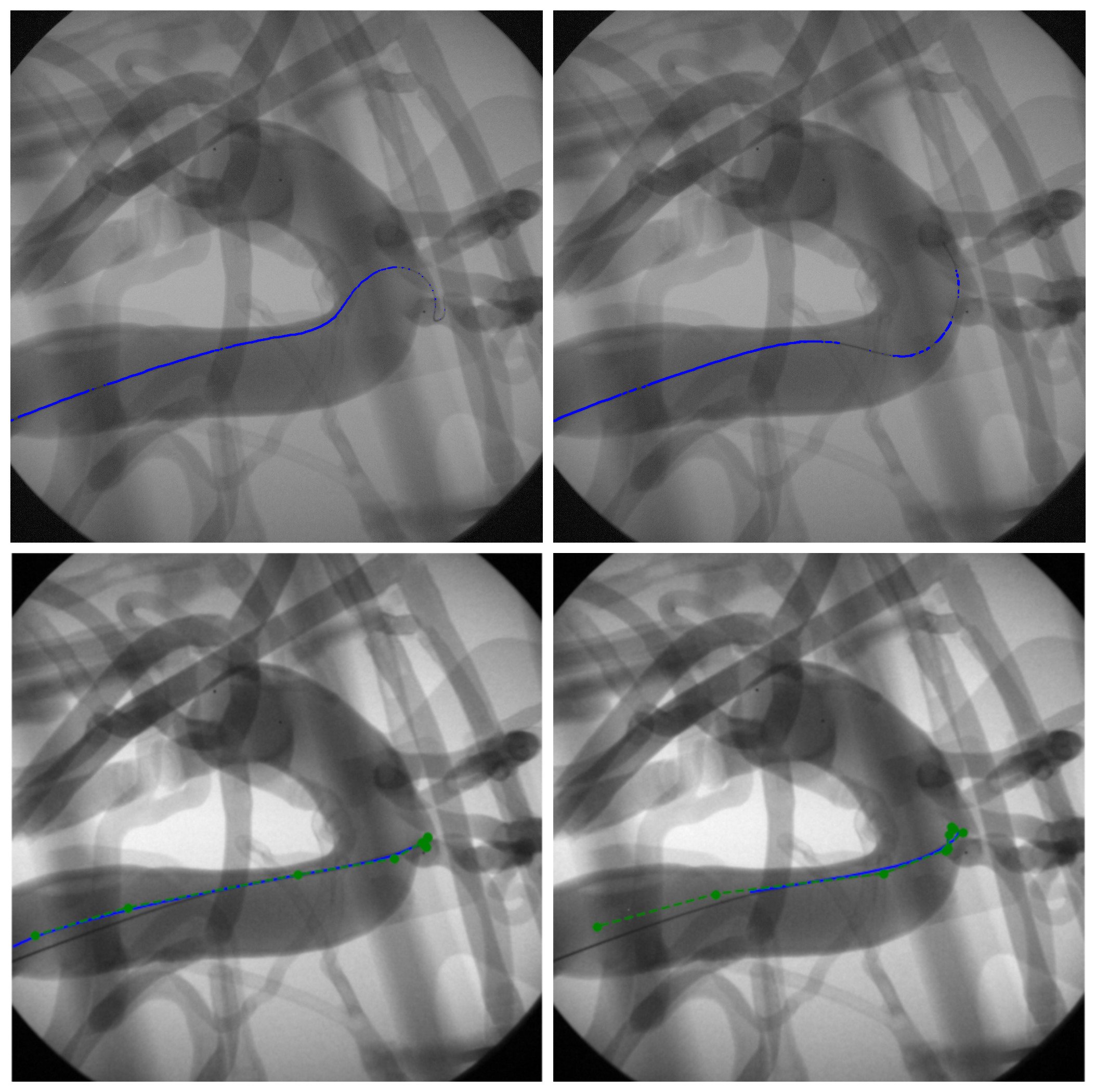}
	\mycaption{SplineFormer vs. Segmentation}{U-Net (\textit{top row}) produces segmentation masks, which often contain discontinuities and require additional post-processing before being used for robotic control. In contrast, our SplineFormer (\textit{bottom row}) directly predicts the guidewire's geometry as a structured shape representation, making it more useful for navigation. }
	\label{ch6fig:spline-vs-segment}
\end{figure}

\break

\paragraph{Navigation Behaviour and Failure Modes.} Throughout the autonomous trials, successful navigation typically exhibited smooth and gradual trajectory adjustments, with the guidewire tip remaining aligned with the vessel centreline. In contrast, failure cases primarily involved two distinct behaviours: \textit{i)} premature turning into an incorrect vessel branch due to ambiguous visual cues near bifurcations, and \textit{ii)} overshooting, where the guidewire tip failed to decelerate or rotate sufficiently to align with and enter the ostium of the target vessel. These failures were particularly prevalent in the \gls{lcca} navigation task, where the vessel geometry presents sharper curvature, narrower diameters, and more abrupt branching angles compared to the \gls{bca}. In most such cases, the guidewire continued advancing along the aortic arch without successfully cannulating the intended branch, indicating that failure was due to spatial misalignment rather than mechanical instability. These observations highlight the need for improved spatial reasoning or the integration of geometric priors in future autonomous navigation models.

\subsection{Qualitative Results}

SplineFormer was trained for 300 epochs on the annotated dataset from Section~\ref{ch6sec:robotic-setup} using the Adam optimizer with an initial learning rate of \(1 \times 10^{-5}\) and the loss function from Eq.~\ref{ch6eq:spline_loss}. As illustrated in Fig.~\ref{ch6fig:spline-vs-segment}, the model effectively predicts the global guidewire shape within a compressed feature space, ensuring a compact and structured geometric representation. A key strength of SplineFormer is its ability to localize key guidewire points precisely. By leveraging a B-spline-based formulation, the model maintains smoothness and structural integrity, ensuring a continuous representation that aligns well with the guidewire’s physical properties.

\subsection{Attention Visualization}

\begin{figure}[ht]
	\centering
	\includegraphics[width=1\linewidth]{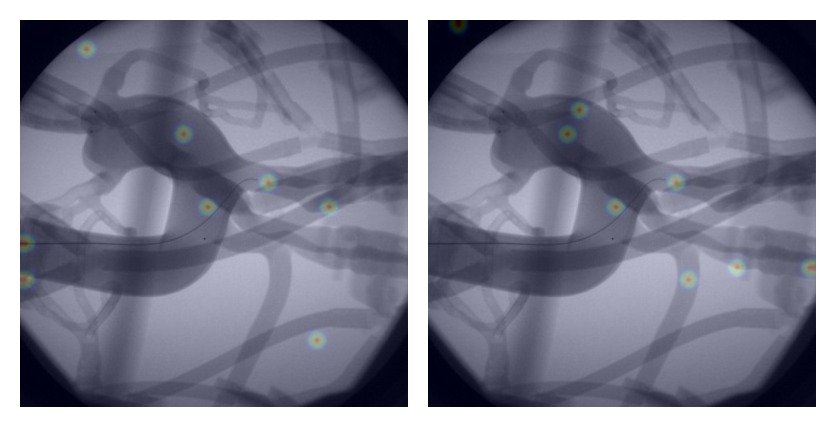}
	\mycaption{Attention visualization}{Attention maps highlight regions where SplineFormer focuses its predictions, including the guidewire tip and key vascular landmarks. This localized attention improves interpretability and facilitates precise autonomous navigation.}
	\label{ch6fig:attention-maps}
\end{figure}

To better understand how the model processes fluoroscopic images, attention maps were generated from SplineFormer’s transformer layers. Using maximal fusion across the final layer, with a discard factor to isolate key features, the resulting visualizations in Fig.~\ref{ch6fig:attention-maps} highlight the critical regions where the model concentrates its predictions. These attention maps reveal a strong focus on the guidewire tip and essential anatomical landmarks, including the central portion of the aortic arch and the \gls{bca} cannulation site.

Unlike traditional segmentation methods, where attention is often dispersed across large areas of the image, SplineFormer exhibits highly localized attention, refining its focus on the most relevant regions for navigation. This precise attention mechanism enhances stability and accuracy in guidewire positioning, improving the reliability of autonomous navigation.

However, not all attention outputs are ideal. In some cases, the attention maps highlight regions that do not correspond directly to anatomical structures or the guidewire. This behaviour is largely attributed to distortions inherent in X-ray imaging, such as noise, non-uniform contrast, and overlapping anatomical features, which may cause the model to attend to visually prominent but semantically irrelevant regions. These artefacts can reduce the reliability of attention interpretation in some frames. Addressing this limitation by incorporating domain-specific priors or preprocessing techniques remains an important direction for future work.

\onehalfspacing
\chapter{Conclusions}
\chaptermark{Conclusion}
\glsresetall

This thesis presents a comprehensive framework to advance autonomy in endovascular navigation, addressing key technical, procedural, and data-centric challenges that hinder current robotic-assisted interventions. Through the development of the CathSim simulation environment, the \gls{expert navigation network}, the Guide3D dataset, and the SplineFormer model, this work introduces novel solutions that integrate high-fidelity simulation, multimodal sensory processing, and geometric learning to enable precise and adaptable catheter navigation. Together, these contributions lay a robust foundation for future autonomous surgical systems that are scalable, safe, and clinically relevant. The outlined future directions—including enhanced simulation realism, multimodal sensing, adaptive learning, and clinical validation—highlight a roadmap toward translating these innovations from controlled settings to real-world interventional medicine, promising to improve patient outcomes and broaden access to minimally invasive procedures.

\section{Summary}

This thesis provides a robust framework to advance autonomy in endovascular navigation, addressing key technical and procedural challenges associated with robotic-assisted interventions in complex vascular environments. By leveraging advancements in simulation-based \gls{reinforcement learning}~\autocite{sutton2018reinforcement, arulkumaran2017deep}, multimodal sensory integration~\autocite{sarker2021machine}, and geometric modelling~\autocite{wei2022coacd}, the proposed solutions overcome existing limitations in heuristic-based systems, enabling precise, adaptable, and scalable solutions for minimally invasive procedures.

The primary contributions are summarized as follows:

\begin{enumerate}
	\item \textbf{CathSim Simulation Framework:}
	      A novel high-fidelity simulation environment was developed to address the bottleneck in reinforcement learning for autonomous catheter navigation. CathSim integrates anatomically accurate vascular models, precise catheter dynamics, and modular simulation tools optimized for \gls{reinforcement learning} algorithms. Leveraging MuJoCo’s rigid body dynamics~\autocite{todorov2014convex}, CathSim ensures realistic frictional forces and collision interactions while maintaining computational efficiency. This framework enables high-throughput sampling and training in simulated environments that closely replicate clinical conditions.

	\item \textbf{Expert Navigation Network (ENN):}
	      A multimodal learning framework, the \acrfull{expert navigation network}, was introduced to improve autonomous catheterization accuracy. \gls{expert navigation network} integrates diverse sensory modalities, including guidewire tip position, joint velocity, fluoroscopic imaging, and force feedback, through a \gls{sac}  policy~\autocite{haarnoja2018soft}. The network’s architecture is optimized for high-dimensional input processing, leveraging convolutional neural networks \gls{cnn} for imaging data and \glspl{mlp} for kinematic and force data. \gls{expert navigation network} demonstrated superior navigation accuracy and robustness, reducing reliance on manual operator intervention~\autocite{arulkumaran2017deep}.

	\item \textbf{Guide3D Dataset and Deep Learning Framework:}
	      The Guide3D dataset was developed as the first high-resolution, open-source dataset for endovascular segmentation and 3D reconstruction. It incorporates over 8,700 annotated biplanar X-ray images with detailed calibration parameters for accurate 3D guidewire geometry estimation. In addition to the dataset, the chapter introduced a \gls{deep learning} framework based on a \gls{gru} architecture and a spherical coordinate representation, enabling accurate and efficient guidewire shape prediction. Together, these contributions address the scarcity of annotated datasets and robust methods for dynamic endovascular imaging, providing a benchmark for the development of \gls{reinforcement learning} and \gls{deep learning}-based algorithms.

	\item \textbf{SplineFormer Model:}
	      A transformer-based B-spline parameterization model, SplineFormer, was introduced for real-time guidewire navigation. SplineFormer predicts guidewire shapes as continuous parametric splines, enabling dimensionally compact, interpretable representations of vascular trajectories. This approach eliminates the reliance on pixel-wise segmentation~\autocite{ronneberger2015u} and enhances robustness against imaging artefacts. Experimental validation demonstrated that SplineFormer achieves high prediction accuracy across diverse vascular configurations, even under noisy or partially occluded imaging conditions~\autocite{vaswani2017attention}.
\end{enumerate}

\section{Future Work}

While this thesis introduces substantial advancements, it also reveals critical paths for further research and development. Future work can be categorized into the following key areas:

\begin{enumerate}
	\item \textbf{Advanced Simulation-to-Reality Transfer (Sim2Real):}
	      The realism of CathSim, while high, cannot fully replicate the complexities of real-world clinical environments, including patient-specific vascular anomalies, biological variability, and imaging artefacts. Future work should focus on minimizing domain discrepancies through techniques such as adversarial domain adaptation, physics-informed neural networks, and \gls{generative adversarial imitation learning}. Additionally, incorporating stochastic variability into CathSim’s vascular models (\eg, dynamic wall compliance or blood flow dynamics) could enhance its utility for simulating real-world scenarios.

	\item \textbf{Expansion of Multimodal Sensing Frameworks:}
	      The current implementation of \gls{expert navigation network} leverages fluoroscopy-based imaging and synthetic force feedback. Future work should expand these sensory modalities by integrating advanced imaging techniques such as \textcolor{primary}{Intravascular Ultrasound (IVUS)}, \textcolor{primary}{Optical Coherence Tomography (OCT)}, and electromagnetic field sensors. Such multimodal fusion would provide richer contextual information for autonomous systems, improving the robustness of catheter navigation in highly tortuous or occluded vessels.

	\item \textbf{Scalable Annotated Data Generation:}
	      Although Guide3D addresses a critical gap in annotated datasets, its scope is limited by the labour-intensive nature of manual annotation and calibration. Future work should focus on developing scalable synthetic data generation pipelines using procedural modelling of vascular anatomy and automated annotation algorithms. Additionally, incorporating domain-specific data augmentation methods (\eg, vascular deformation, occlusions, or contrast variations) could improve algorithm generalization.

	\item \textbf{Adaptive Reinforcement Learning for Dynamic Environments:}
	      The \gls{reinforcement learning} algorithms employed in this thesis rely on pre-defined simulation environments with fixed anatomical models. However, real-world scenarios involve dynamically changing conditions, such as blood flow variations, vessel deformation, or patient movement. Future research should explore meta-reinforcement learning and curriculum learning techniques to enable adaptive policies that generalize across unseen vascular anatomies and procedural conditions. Additionally, hierarchical \gls{reinforcement learning} could be employed to break down complex tasks, such as multi-target cannulation, into sub-goals, improving learning efficiency.

	\item \textbf{High-Fidelity Modelling of Tissue-Catheter Interactions:}
	      Current simulation models assume simplified rigid body dynamics for catheter-vessel interactions. Future developments should incorporate high-fidelity biomechanical models, including tissue elasticity, shear stress, and endothelial damage thresholds. Such enhancements would enable more realistic simulations of catheterization risks, such as vessel perforation or thrombosis, allowing for safer and more accurate training.

	\item \textbf{Real-Time Control Optimization:}
	      The computational demands of SplineFormer and \gls{expert navigation network}, while suitable for simulated environments, may pose challenges in real-time clinical applications. Future research should focus on optimizing these architectures through neural network pruning, quantization, and hardware acceleration (e.g., FPGA or TPU deployment). Moreover, incorporating \gls{model predictive control} frameworks could enable real-time trajectory adjustments based on environmental feedback, improving navigation safety and precision.

	\item \textbf{Expanding Applications to Other Interventional Domains:}
	      The methodologies developed in this thesis have the potential to be extended beyond endovascular navigation. Applications in bronchoscopic navigation, neurosurgical interventions, and gastrointestinal endoscopy could leverage the same principles of \gls{reinforcement learning}, multimodal sensing, and geometric modelling. Future work could explore domain-specific modifications to adapt these frameworks for other minimally invasive procedures.

	\item \textbf{Clinical Validation and Workflow Integration:}
	      The ultimate goal of autonomous systems is safe and effective deployment in clinical settings. Future research should prioritize clinical validation through multicentre trials, focusing on metrics such as procedural success rates, time savings, and safety outcomes. Additionally, integrating these systems into existing surgical workflows requires user-friendly interfaces and intuitive controls that align with clinicians’ expertise and preferences.
\end{enumerate}

\section{Final Reflections}

This thesis represents a significant advancement in the field of autonomous endovascular navigation, combining state-of-the-art simulation, \gls{artificial intelligence}, and robotics to address key challenges in \gls{minimally invasive surgery}. By bridging gaps in simulation fidelity, multimodal data integration, and geometric modelling, this work lays the foundation for the next generation of autonomous surgical systems. Future developments, informed by the directions outlined above, hold the potential to revolutionize interventional medicine, reducing risks, enhancing precision, and expanding accessibility to life-saving procedures.

Beyond the immediate technical contributions, the broader significance of this research lies in its potential to transform endovascular robotics into a domain characterized by greater autonomy, reproducibility, and procedural efficiency. The structured and interpretable approaches developed here address a key clinical need: ensuring that autonomous systems remain transparent and auditable, which is critical for both physician trust and regulatory approval. Regulatory pathways, such as those governed by the FDA in the United States or the EMA in Europe, will require rigorous pre-clinical validation in phantoms and large-animal models, followed by clinical trials that demonstrate safety, reliability, and clinician override mechanisms. Integration into existing workflows must prioritize user-centric design, ensuring that automation enhances rather than disrupts established clinical practice.

Looking ahead, the rapid evolution of foundation models opens promising directions for advancing autonomy in surgical robotics. \gls{large language models} can provide procedural priors and context-aware policy guidance, while \gls{large vision models} can enhance perception under noisy and variable imaging conditions. Recent work on combining \glspl{large language models} with fuzzy reinforcement learning for cooperative endovascular navigation~\cite{yao2025multi} demonstrates how these models can contribute higher-level reasoning while maintaining safety constraints through adaptive uncertainty management. Integrating such models into future iterations of SplineFormer and related frameworks could enable systems that generalize more effectively across patient anatomies, dynamically adapt to intraoperative variability, and remain aligned with clinical safety requirements.

In summary, this thesis provides both a technical foundation and a translational perspective for autonomous endovascular navigation. By addressing regulatory considerations and pointing to the integration of \glspl{large language models} and \glspl{large vision models} as part of future development, it charts a pathway toward clinically viable systems that combine intelligence, interpretability, and safety, key prerequisites for eventual adoption in real-world interventional medicine.



\printbibliography[title={References}, heading=bibintoc]



\end{document}